\documentclass{article}

\usepackage{amssymb}
\usepackage{amsmath}
\usepackage{graphicx}

\usepackage{mathtools}

\usepackage{amsthm}
\theoremstyle{plain} 
\newtheorem{theorem}{Theorem}[section]

\theoremstyle{definition}
\newtheorem{definition}{Definition}[section]
\theoremstyle{remark}

\usepackage{appendix}

\usepackage{titlesec}

\graphicspath{ {figures/} }

\usepackage{resizegather}

\usepackage{parskip}
\makeatletter
\def\thm@space@setup{
  \thm@preskip=\parskip
  \thm@postskip=\thm@preskip 
}
\makeatother

\usepackage{xargs}
\usepackage[pdftex,dvipsnames]{xcolor}

\usepackage[colorinlistoftodos,prependcaption,textsize=tiny]{todonotes}
\newcommandx{\add}[2][1=]{\todo[linecolor=blue,backgroundcolor=blue!25,bordercolor=blue,#1]{#2}}
\newcommandx{\edit}[2][1=]{\todo[linecolor=red,backgroundcolor=red!25,bordercolor=red,#1]{#2}}
\newcommandx{\misc}[2][1=]{\todo[linecolor=Plum,backgroundcolor=Plum!25,bordercolor=Plum,#1]{#2}}
\newcommandx{\info}[2][1=]{\todo[linecolor=OliveGreen,backgroundcolor=OliveGreen!25,bordercolor=OliveGreen,#1]{#2}}
\newcommandx{\comment}[2][1=]{\todo[disable,#1]{#2}}

\showboxdepth=\maxdimen
\showboxbreadth=\maxdimen

\usepackage{hyperref}
\hypersetup{
    colorlinks=true,
    linkcolor=blue,
    filecolor=magenta,      
    urlcolor=cyan,
}

\usepackage{siunitx}

\begin{document}

\begin{titlepage}
\begin{center}

\textsc{
\LARGE MathMods Erasmus Mundus Program\\[4mm]
University of L'Aquila\\[2mm]
University of Hamburg\\[2mm]
University of Barcelona\\[4.5cm]
}
\textsc{\Large Master Thesis}\\[0.7cm]

{ \huge \bfseries Neural network models}\\[1.0cm]

\begin{minipage}{0.4\textwidth}
\begin{flushleft} \large
\emph{Author:}\\
Plamen \textsc{Dimitrov}
\end{flushleft}
\end{minipage}
\begin{minipage}{0.4\textwidth}
\begin{flushright} \large
\emph{Advisors:} \\
Prof.~Leonardo \textsc{Guidoni}
Prof.~Debora \textsc{Amadori \ }
\end{flushright}
\end{minipage}

\vfill
{\large \today}

\end{center}
\end{titlepage}

\vspace*{2.5cm}
\begin{abstract}
\noindent This work presents the current collection of mathematical models related to neural networks and proposes a new family of such with extended structure and dynamics in order to attain a selection of cognitive capabilities. It starts by providing a basic background to the morphology and physiology of the biological and the foundations and advances of the artificial neural networks. The first part then continues with a survey of all current mathematical models and some of their derived properties. In the second part, a new family of models is formulated, compared with the rest, and developed analytically and numerically. Finally, important additional aspects and any limitations to deal with in the future are discussed.
\end{abstract}
\newpage

\tableofcontents
\newpage


\section{Background}

\subsection{Introduction}

The brain is the only human organ that has spawned a wide variety of disciplines originating in and focused on completely independent approaches that don't truly intersect each other for a common theory. Scientific disciplines like
\begin{itemize}
  \item psychology
  \item behavioral science
  \item cognitive science
  \item psychiatry
  \item artificial intelligence
  \item machine learning
  \item neurology
  \item neuroscience
  \item neuroanatomy
\end{itemize}
are all devoted to studying different aspects of this organ with the psychology type sciences taking a higher level (symbolic) approach and the neuroanatomy type sciences - a lower level (circuit) approach. The hope is that they will intersect somewhere in the middle, fully explaining higher level cognitive phenomena through lower level physiological and morphological such. However, this hasn't happened as of yet and when it does, it will be analogical to finally creating a comprehensive coherent theory demystifying questions about the human mind which are currently only asked in the field of philosophy.

More engineering oriented fields like artificial intelligence and machine learning linger in the middle between the two ends with attempts to reproduce high level phenomena by both more abstract methods (Lisp, A* search) as well as methods resembling the natural structure of the brain (artificial neural networks, neocognitron). The ultimate goal of such fields is achieving some practical use of the implemented algorithms - be it human-dependent tasks like image/voice/text recognition and classification or self-driving cars and real time translation. Their approaches usually stem in developing an algorithm for the job which strives to achieve maximal accuracy rate on specific commonly-accepted data set. For this purpose, obtaining good empirical results on the given data set is often sufficient for adoption of the developed techniques and results in narrow specialization of the implemented algorithms. However, many results in neuroscience suggest the existence of a universal approach that could be the potential solution in all different areas of AI application - from voice and image recognition to language comprehension and translation \cite{Rauschecker1995, Buchel1998, Ramachandran1998}. Even though a well-specialized implementation will perform better on the specific area of application, one must then advocate for modelling and reproducing more general cognitive behavior and use more mathematical rigor in order to draw useful conclusions from such generality.

To construct such a universal algorithm however is not an easy task. This is one of the reasons for multiple disappointments and periods of loss of interest and high criticism in the area of artificial intelligence that are commonly referred to as the AI winters. In order to build something, one should completely understand it, and the lack of complete or at least sufficient understanding of the brain is among the primary reasons for the occurrence of so many attempts to explain it.
\begin{figure}[!htbp]
\centering
\includegraphics[width=0.75\textwidth]{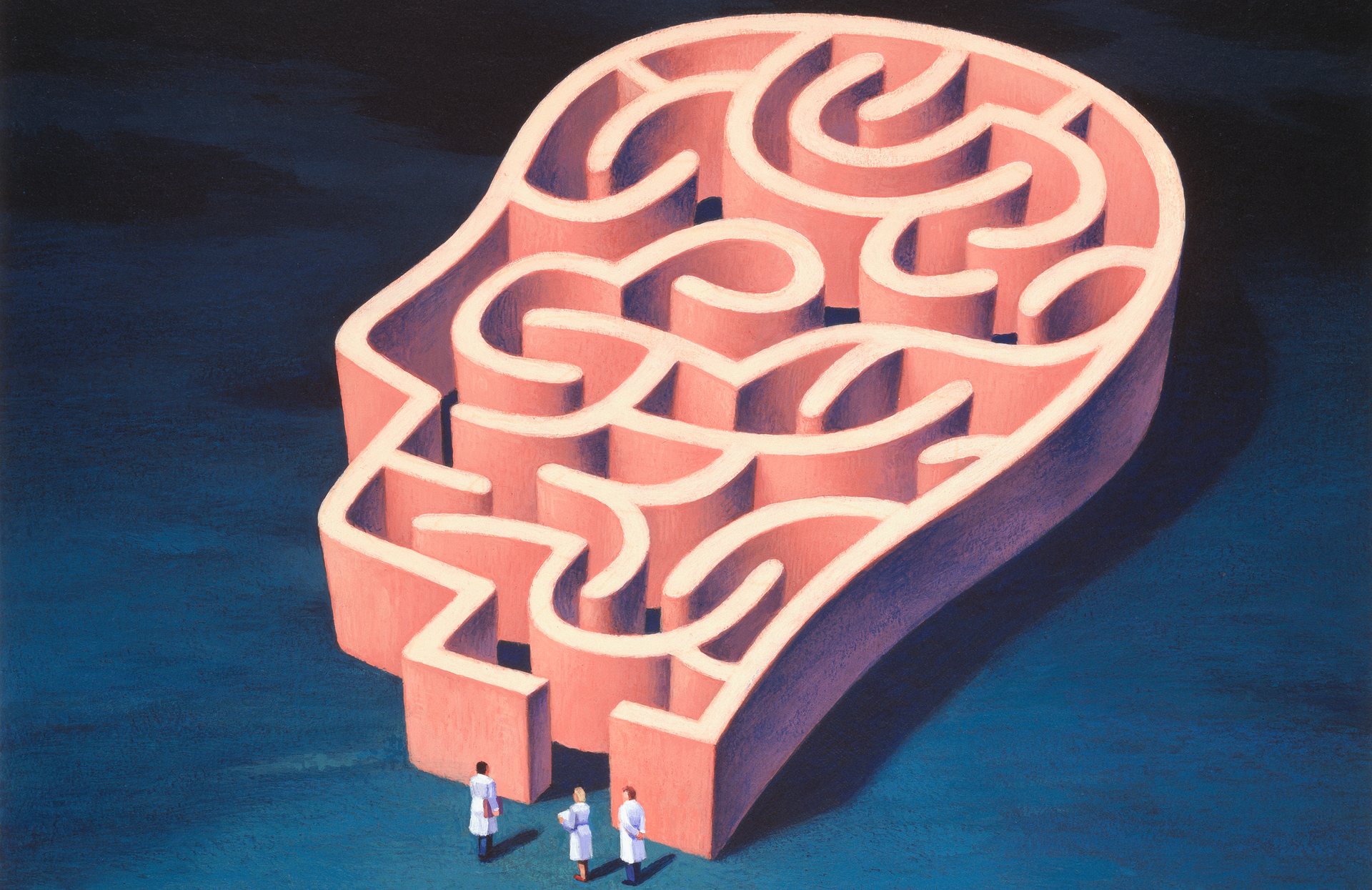}
\caption{The complexity of the brain studied from many different scientific angles gets all participants separated and lost in the end.}
\label{labyrinth}
\end{figure}
Figure \ref{labyrinth} is just one of many depictions of the same problem - each discipline is like a flash of light illuminating different angle of the same giant object and none of them sees the whole of it or how large it really is. And the realization about all the limitations that have to be faced is by far not restricted only to the engineering field. In psychiatry, using psychiatric drugs has so many side effects because using them to treat complex psychiatric disorder is a bit like trying to change a car's engine oil by opening a can and pouring it all over the engine block - some of it will dribble into the right place but a lot of it will do more harm then good \cite{Anderson2016}. This is how neurological issues related with the brain are solved at the present day after all the progress in medicine from the last centuries. General developments in neuroscience like fMRI have been really helpful in pinpointing active areas of the brain to certain stimuli but the map is very crude and uses increased blood density rather than actual neuron spikes. Even if we were able to record the complete history of every spike of every neuron and store the entire neural circuitry or connectome as proposed by \cite{AlivisatosChunChurchGreenspanRoukesYuste2012}, the data generated just from the brain of the smallest organisms with a minimal diversity of possible behaviors like a fruit fly would include approximately $10^4$ neurons to analyze.

It is therefore vital to be under no illusion about the difficulties involved here. Nevertheless, the possibility of existence of such a universal method alone is attractive enough so that we would like to make a step in this direction. In particular, we will investigate the mathematical modelling setting of the scientific fields above, concentrating on neural networks as the least common denominator among all of them. We will study the reasons and derivation and include further references for the major neuron models and the resulting network models in our investigation. Using these as a starting point, we will then propose a family of models that could be best motivated with a set of cognitive behaviors they must exhibit. We have compiled these cognitive phenomena with the expectation that they are general enough to be shaped into solutions of different areas of application like the ones we mentioned before. The validity of the models then depends on the validity of these requirements which is studied in depth both analytically and numerically through the formulated models. We will discuss it in a final section but in order to be able to talk about the summary of existing models, we need some additional background for the relevant life and computational sciences in this first section.

\subsection{Biological neural networks} \label{bnns}

The human brain comprises 100 billion neurons as its principal cellular elements, each one connected with around 10000 others. Consequently, the potential complexity of the resulting network is vast in terms of both possibilities and interpretation. It is called by many the most complex machine on Earth and perhaps the universe \cite{Independent2014, Economist2011}. In addition to the combinatorial size of possibilities for connectivity, the neurons differentiate in many different types with different functions and by far don't constitute the entire brain. Other cellular types are also present and even estimated to be five times more than the neurons (90\% of the brain), namely several types of glial cells that play a crucial role in the maintenance of the neural network, neural development and in the generation of myelin which is used for isolation and is one of the main reasons for fast impulse conductance. Despite their essential supportive nature, we will not discuss them in more detail and instead focus on the main communication infrastructure. This section will begin with an overview of the morphology and physiology of the neuron, the main elementary node or element of the brain, then provide a short description of the ways in which individual neurons communicate with each other in pairs, networks, and systems, and finish with plasticity as one of the most characteristic neural network properties.

The 100 billion neurons in the brain share a number of common features. The anatomical variation of these neurons is large, but the general morphology and their electrical and ligand dependant responsiveness allows these cells to be classified as \emph{neurons} (coined in 1891 by Wilhelm von Waldeyer) \cite{Llinas2008}. Neurons are different from most other cells in the fact that they are polarized and have distinct morphological regions, each with specific functions.
\begin{figure}[!htbp]
\centering
\includegraphics[width=0.85\textwidth]{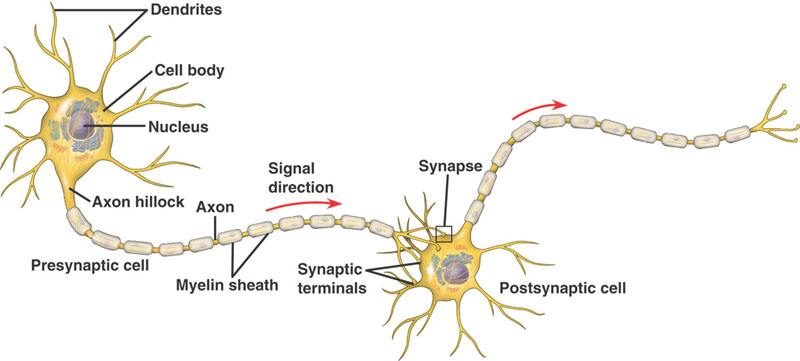}
\caption{Neuron morphology and physiology.}
\label{neuron}
\end{figure}

Neurons are generically characterized by a central cell body or \emph{soma} that comes in different shapes. The soma contains the cell nucleus and most of the genomic expression and synthetic machinery producing the proteins, lipids, and sugars that constitute neuron's cytoplasm and membranes. For a general neuron, the soma extends into inputs and outputs as in figure \ref{neuron}.

The region where a neuron receives connections from other neurons or the input pole consists of extensively branching tree-like extensions of the soma membrane known as \emph{dendrites} (coined in 1889 by William His from dendros meaning "tree" in Greek). The dendrites arise directly from the cell body in vertebrate neurons similarly the ones in the figure. However they arise from the axon in invertebrate neurons. Furthermore, the body is also a receiving site in most neurons.

The output pole, called the \emph{axon} (coined in 1896 by Rudolph Albert von Kolliker) arises as a single structure from the soma (and occasionally from a dendrite). The axon conducts propagating electrochemical signals termed \emph{action potentials} (also spikes or impulses) that are usually initiated at the axon base or \emph{hillock} and move away from the soma to the \emph{terminal} regions of the neuron. Axons can be rather long extending up to about a meter in some human sensory and motor nerve cells. A \emph{synapse} in the terminal region of the axon is the place where one neuron forms a connection with another (called \emph{postsynaptic} neuron, the first one being respectively \emph{presynaptic}) and conveys information through the process of synaptic transmission.

However, it is important to note that there are many exceptions to these general rules of neuronal organization. There are some dendrites that also serve as output systems \cite{PinaultSmithDeschenes1997}. In some neurons, e.g. peripheral sensory neurons, the input occurs via axons. Neurons can have many types of branching or no branches at all, examples being receptors cells in the carotid glomus, in gustatory system in vertebrate tongue or as photoreceptors in the retina \cite{Llinas2008}.

Last important morphological feature is that the presynaptic cell is not directly connected to the postsynaptic cell - the two are rather separated by a gap known as the \emph{synaptic cleft}. Therefore, the communication between a presynaptic and a postsynaptic neuron (\emph{synaptic transmission}) is realized through a \emph{neurotransmitter}, i.e. a chemical messenger released through an action potential at the presynaptic terminal (a process called \emph{exocytosis}) that diffuses at the synaptic cleft and binds to selective \emph{receptors}. The binding to the receptors leads to a change in the permeability of ion channels in the membrane and in turn a change in the membrane potential of the postsynaptic neuron known as a \emph{postsynaptic synaptic potential (PSP)} which may then lead to an action potential in the postsynaptic neuron. This brings to the conclusion that signaling among neurons is associated with changes in their electrophysiological properties.

Now let us provide more details and important terminology regarding the eletrophysiological properties of a neuron. The change in the PSP caused by a presynaptic action potential may be a decrease in the polarized state of the membrane called \emph{depolarization} or an increase called \emph{hyperpolarization}. The membrane potential in the absence of any presynaptic stimulation is about -60 mV inside with respect to the outside of the postsynaptic cell and is called \emph{the resting potential}. Hyperpolarization then makes the potential inside the cell even more negative while sufficiently large depolarization is the one to trigger an action potential. The action potential is associated with a very rapid depolarization to achieve a peak value of about +40 mV in the brief period of 0.5 msec \cite{Byrne2016}. The peak is followed by an equally rapid period of \emph{refractoriness} when the neuron cannot be excited which associated also with a \emph{repolarization} phase that completes a depolarization-repolarization cycle.

The voltage at which the depolarization becomes sufficient to trigger an action potential is called a \emph{threshold}. It can be reached through multiple presynaptic spikes leading to smaller changes in the membrane potential which accumulate in time (\emph{temporal summation} which together with spatial summation over the dendritic area is also called \emph{synaptic integration}) or through a single suprathreshold spike but the resulting action potential is identical in amplitude, shape, and duration. The action potential will also not change if the input stimulus leads to a depolarization much larger than the threshold. The presynaptic spikes can also cause hyperpolarization which reduces the change of action potential on the postsynaptic side. The PSP can therefore be divided in to \emph{excitatory postsynaptic potential (EPSP)} which is the one responsible for the depolarization and \emph{inhibitory postsynaptic potential (IPSP)} which is the one responsible for the hyperpolarization. An IPSP is called inhibitory because it tends to prevent the postsynaptic neuron from firing an action potential thus regulating the ability of excitatory signal to bring about this same event. The conclusion from this is that synaptic transmission has two basic forms, namely \emph{excitation} and \emph{inhibition}.

Generally, a single action potential in a presynaptic cell does not produce an EPSP large enough to reach threshold and necessarily trigger an action potential. But longer-duration of even subthreshold stimulus can lead to multiple action potentials (spike sequences or \emph{spike trains}) with frequency depending on the intensity of the stimulus. The same is true for receptor cells from pressure of touch to intensity of light as well as motor cells where the output frequency determines the contraction level of muscle \cite{Llinas2008}. This, together with the identical properties of action potentials suggests that the nervous system encodes information in frequency of isotropic pulses instead of their amplitude and shape.

\begin{figure}[!htbp]
\centering
\includegraphics[width=0.75\textwidth]{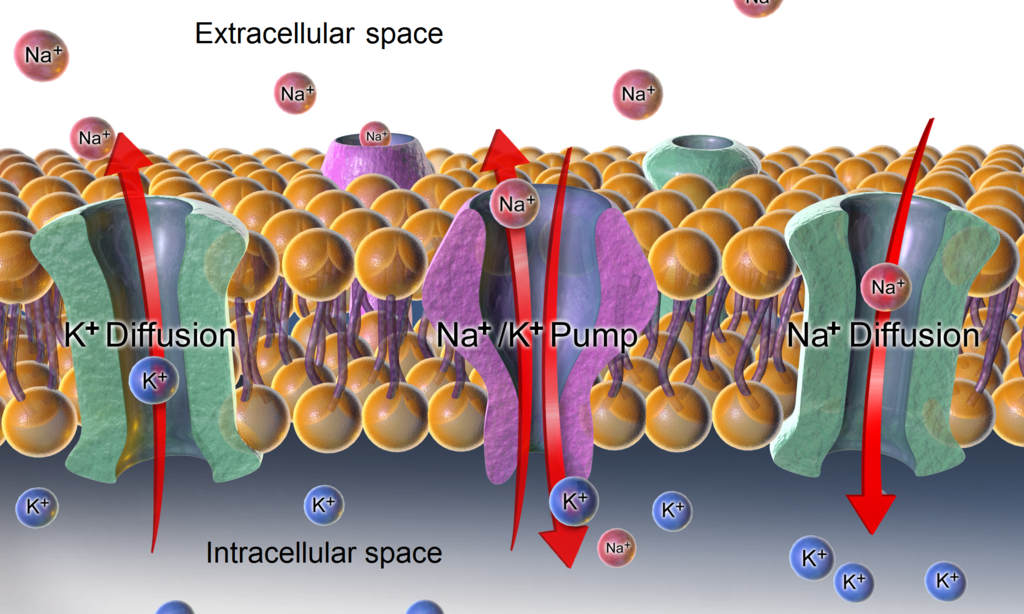}
\caption{Ion gates comprise ion pumps that maintain nontrivial equilibrium potential of the membrane and ion channels which are responsible for the active electric properties of a neuron \cite{Blausen2014}.}
\label{ion_gates}
\end{figure}

The generation of action potentials is an example of the active electrical properties of neurons because they are brought about by active, voltage-dependent means \cite{Llinas2008}. More specifically, the membrane potential might be affected by activation of ion channels which is caused by the neurotransmitter release on the presynaptic side. The passive electrical properties of neurons are also very important for describing their overall dynamics in mathematical models presented in later sections. The cell membrane as a bilipid layer is a nearly perfect insulator but has \emph{ion gates} as proteins embedded into it. These ion gates can be \emph{ion pumps} or \emph{ion channels}. There are also \emph{non-gated 'leakage'} channels (constantly open channels permeable to potassium) which we mention for completeness but later on we will only consider the ion gates.

Ion pumps permanently (for an energy cost) transport ions from one side of the membrane to the other (depending on the ions transported and therefore the ion pumps transporting them, e.g. sodium is less while potassium is more concentrated inside) maintaining a voltage difference through a difference in the ion concentrations on the inside and on the outside of the neuron. An ion channel will then reach an equilibrium when there is no more flow of ions between the two sides. The membrane potential at equilibrium is the \emph{reversal/Nernst potential} of that particular ion assuming a single ion dominated system. Among the main ion species found in the central nervous system are sodium, potassium, calcium, and chloride ions.

Similarly the neuron's morphology, there are many exceptions when it comes to its physiological properties. Although the conductance of most ion (voltage-gated) channels is increased by membrane depolarization, the conductance of some channels is increased when the membrane is hyperpolarized. In addition to axonal spikes, action potentials can also travel along dendrites either in the direction of the soma (centripetal spikes) or away of the soma and reach the most distal dendritic branches (centrifugal spike) depending mostly on the distribution of ion channels over the dendritic tree. Axons can also branch out into parallel pathways early in the spike propagation or later on into a distal dendrite, in both cases sending very similar spike trains but with different resulting conduction due to different axonal diameters and possible spike failure.

Now let's look at the network level and more specifically at the general types of network connectivity that is experimentally observable. There are three major classes of connections in all areas of the brain - \emph{feedforward} (bottom-up) connections transfer activation to neurons which are further along the processing path that started with external input, \emph{lateral} (recurrent) connections are used for communication among neurons that are considered to be within the same stage along the processing path, and \emph{feedback} (top-down) connections which project activation back. The feedforward and feedback connections are typically comparable in terms of numbers while the lateral connections outnumber each. The \emph{feedforward networks} are built only from feedforward connections and exclude the possibility of cycles, the \emph{recurrent networks} include lateral connections as well, and the \emph{fully recurrent networks} allow for connections in any direction among all neurons. This is also the reason why when talking about fully recurrent networks, one talks simply about connections and and forgets the overall categorization.

To complete the various anatomical organizations in the brain, we must include a description of the system level structure. The emphasis of any network model described later on is mainly on \emph{neocortical circuits} so we will briefly outline some principal organizations of the \emph{neocortex}, the outer and developed later in evolution layer of the \emph{cerebral cortex}. Even though the neocortex or simply \emph{cortex} can be horizontally split into four lobes responsible for different cognitive functions and modalities, its detailed structure is similar among all of them which distinguishes it significantly from older parts of the brain like the \emph{brainstem} \cite{Trappenberg2010}. The structure observed in the neocortex comprises of six (to ten depending on enumeration) \emph{layers} where the fourth layer is generally viewed as an input layer (due to predominant number of white matter afferents) while the fifth layer projects mostly outwards and is therefore viewed as an output layer. The \emph{white matter} consists mostly of axons and connects distant areas which implies that it must be used for global communication. The second and third layers are thought to be responsible for long range lateral (within layer) communication. The sixth layer neurons mostly project into the first layer. The neurons are also arranged in \emph{columns} which are thought of to be the functional units (with neurons within the column storing redundant information) since similar connectivity patterns are observed within and among these columns.

Finally, perhaps the most significant property of neural networks which makes them an evolutionary advantage in the long term is their \emph{synaptic plasticity}. A persistent (ten minutes or longer) increase in synaptic transmission efficacy is called \emph{long-term potentiation (LTP)} and a decrease of the same kind is called \emph{long-term depression (LTD)}. There could be multiple physiological and biophysical mechanisms for realizing such synaptic plasticity, among them changes of the number of release sites, the probability of neurotransmitter release, the number of transmitter receptors, and the conductance and kinetics of ion channels all of which have been demonstrated for specific cell types in multiple papers \cite{Trappenberg2010}. Additionally, a spike timing as well as calcium dependent plasticity was observed empirically in some cells and is examined in further detail in some of the models and learning rules to follow.

\subsection{Artificial neural networks} \label{anns}

No doubt a large portion of the models developed concerning neural networks are algorithms without a rigorous mathematical treatment but achieving a very good measurable accuracy at a specific task. These are computational models within the field of machine learning that deserve at least a brief overview here together with any computational terminology involved in the following sections. With the increasing advanced and complexity of such algorithms or \emph{artificial neural networks (ANNs)}, we should also make an effort to understand them a bit more rigorously and not just their biological counterparts. We will start with an overview of the basic terminology and concepts used in machine learning and therefore important also for the ANNs and our numerical implementations later on. We will then move to the classical ideas and developments and into their latest extensions with deep learning. We will complete the section with some existing criticisms about all these computational models.

All machine learning algorithms are based on induction as opposed to (e.g. mathematical) deduction. As the name of the discipline implies, they are learning to perform a specific task from empirical data which cannot be algorithmically specified in any straightforward way. If the learning they perform is \emph{supervised}, the task is to predict an output value or \emph{label} from an input value given a \emph{training set} of examples as input-output value pairs. It is called supervised due to the presence of the training labels that teach the network the right answers as opposed to \emph{unsupervised learning} where these are not provided. There are also paradigms in between like \emph{reinforcement learning} where the right answers are not provided but the algorithm now termed \emph{agent} still receives feedback in the form of a \emph{reward or punishment} from the environment as well as \emph{semi-supervised learning} where both supervised and unsupervised approaches are combined during the \emph{training phase} with both \emph{labelled} and \emph{unlabelled data}.

The two major tasks in supervised learning are \emph{classification} where the algorithm must predict a discrete label value and \emph{regression} where the value is continuous. As such, they define supervised learning as a task of approximating a function over all possible inputs from finitely many known values. While this is also studied in much depth in statistics, the point here is to find algorithms that could gradually learn (approximate and use to predict labels for new inputs) the function that generalizes best over all data. For this purpose, a \emph{test set} is often used in addition to the training set in order to avoid \emph{overfitting}, i.e. performing too well on the training set and a lot worse on a 'fresh new' test set). \emph{Underfitting} is another although generally smaller danger. Overfitting is often caused by too many free parameters or too small training set so it could also be reduced by a technique called \emph{cross-validation} which helps in selecting these parameters by partitioning the training set and obtaining validation errors through testing on each partition and training on the rest. Comparison in performance of the algorithms in terms of \emph{accuracy rate} or \emph{test error} is then possible on standardized \emph{datasets} that are shared and known in the machine learning community. Major tasks in unsupervised learning are \emph{clustering} (while there are no explicit labels for the inputs, similarities among them are still inferable), \emph{dimensionality reduction} (removing unnecessary information for easier interpretation or in the case of ANNs - compression performed by \emph{autoencoders}), and \emph{outlier detection}.

It is surprising how many practical applications (from navigating a road to speech recognition) can be translated into the tasks described above and that all of the tasks can also be performed well by ANNs among many other machine learning algorithms. This is at least the case after multiple periods of abandonment due to their realized limitations and earlier overestimation of their capabilities typical for the artificial intelligence field.

The very first major work in the direction of pure computation with regard to the nervous system was done by a neurophysiologist Warren McCulloch and a mathematician Walter Pitts who showed that a simple model of a neuron (simply a threshold sum of inputs described also in later sections) can reproduce the basic AND/OR/NOT logical functions \cite{McCullochPitts1943}. It was very influential especially because of the belief that logical reasoning could solve AI \cite{Kurenkov2015} but lacked the crucial ability to learn so was later on extended by a neurobiologist Frank Rosenblatt in his \emph{perceptron} model to include weights (and a bias term similar to a current injected straight to the neuron's soma) that could make one input more influential than another \cite{Rosenblatt1958}.

The guiding principle for learning that he used was based on another fundamental work by the psychologist Donald Hebb who conjectured that synaptic changes are driven by correlated activity of pre- and postsynaptic neurons, i.e. neurons that fire together also wire together \cite{Hebb1949}. More formally, the strength of a synapse from neuron A to neuron B will increase if the firing of neuron A often contributes to the firing of neuron B through that synapse. \emph{Hebbian learning} is the way this conjecture is usually referred to and can be generalized in various ways, one in particular to also reduce the strength of the synapse should the neuron A repeatedly fail to contribute to the firing of neuron B.

The perceptron then learns by readjusting the weights in the direction of lower error if its output does not match the desired label and thus correlates better the desired and actual outputs. Due to the presence of sum thresholding or \emph{activation function} for the sum of inputs, the perceptron performs binary classification which can be extended to multi-category classification by adding a \emph{unit} or neuron for each category to form a \emph{layer}. This layer is an \emph{output layer} fully connected to the inputs and the resulting two layer network is one of the most canonical examples of an ANN - a perceptron ANN approximating a vector-valued function. One way to classify the vector of input states is thus to choose the maximum weighted sum among the outputs.

An even more daring deviation from biological plausibility is Widrow and Hoff's \emph{ADALINE} \cite{Widrow1960}. Even though its main contribution is in implementing neurons in electrical circuits through resistors with memory (memistors), it explores the option of simplifying the perceptron altogether by dropping the activation function and simply outputting the weighted sum of each neuron. This linear form is easier to study mathematically because it removes jump discontinuities and nonlinearities from the model neuron. In addition, the learning/approximation problem can be restated as optimization problem where the derivative of the error can be used in a stochastic or other gradient descent.

Although \emph{connectionism} as the idea that complex computations can be performed by many simpler connected units comes with these first computational models, the models themselves are fundamentally limited in their computational power. A major critic of the perceptron for instance is that it cannot reproduce the XOR logical function since it violates the linear separability condition \cite{MinskyPapert1988}. A solution to this, extension to the perceptron, and an eventual exit of the first AI winter is the addition of \emph{hidden layers} and \emph{backpropagation} learning. If there are multiple layers between the input and output layer responsible for multiple levels of abstraction and feature extraction, much more complex computations including approximating the XOR functions can be performed. Since the first weight optimization approach described above can no longer be applied (we only know the error at the output layer), backpropagation of the error from the output layer to the weights of all hidden layers using chain rule splits and redistributes the correction in the entire network. This approach together with the greater computational power of the multilayer ANNs was shown in a sequence of papers most notable from which is the description by Rumelhart, Hinton, and Williams \cite{RumelhartHintonWilliams1986}. A result of major mathematical importance to once and for all solve any approximation issues is the proof that multilayer ANNs are \emph{universal approximators}, i.e. with sufficiently many hidden units in a middle layer they can approximate any Borel-measurable function between finite dimensional spaces to an arbitrary degree of accuracy \cite{HornikStinchcombeWhite1989}.

The multilayer ANNs with backpropagation learning are the state of the art ANNs until the present which after a second AI winter returned with a few small modifications as \emph{deep neural networks or DNNs}. These modifications involve more careful weight initialization and choice of activation function \cite{GlorotBengio2010}. The paper which is among the first to talk about "Deep learning" is again due to Hinton \cite{HintonOsinderoTeh2006} and essentially considers a variant of DNN, namely 'deep belief network' described below, which is optimally trained layer-by-layer through a semi-supervised learning (unsupervised learning to initialize the weights). Such variations of ANNs/DNNs are what remains to be discussed until the end of this section. The major ANN classification is based on the previous connectivity classification of feedforward, recurrent and fully recurrent networks. The DNN variations are then based on modality specialization (CNNs for visual input, LSTMs for sequential input) and connectivity (DBNs as fully recurrent networks).

As a variation of an ANN/DNN which is very successful in computer vision, a \emph{convolutional neural network or CNN} was first applied to handwritten zip code recognition by LeCun \cite{LeCunBoserDenker1989}. The main practical improvement that a CNN introduced involves the addition of a \emph{convolution layer} where instead of each neuron having a unique weight for each of its inputs, we have neurons with weights that are identical (and identically modified) at multiple disjoint subregions of the same size within the previous layer. In this way we don't have to learn the same feature like specific edge orientation by multiple neurons and the neuron's selectivity becomes invariant with respect to the position of the feature in the image. This draws inspiration from Fukushima's neocognitron \cite{Fukushima1980} and in return from Hubel and Wiesel's emprical results on the hierarchical organization of the visual nervous system \cite{HubelWiesel1962} since the complex cell observed and modelled there pertain a certain degree of spatial invariance. A second practical improvement that resembles also the hierarchical arrangement of edge-sensitive simple, complex to hypercomplex cells there is the addition of a \emph{pooling layer} as a form of down-sampling so that a layer is projected in a smaller size successive layer. In this way the number of free parameters and therefore chance of overfitting as well as the training time is reduced while a neuron from a later layer 'sees' the features from the previous layer and thus a larger region in the original input image.

Another somewhat successful variation of an ANN/DNN is the \emph{deep belief network or DBN} which as their name speaks is capable of internal representations of the perceived inputs. The groundwork for a DBN is a \emph{Boltzmann machine} which is a \emph{graphical (graph) model} performing \emph{energy-based learning} with nodes as hidden and visible stochastic variables where the visible variables determine the probabilities of the hidden variables. The essential symmetry feature of Bolzmann machines is that the probability of a visible variable (input) can be calculated from a hidden variable (output) which allows for the original input to be generated from an internal representation. This models are called \emph{generative graphical models}. DBNs are then multilayer networks of restricted Boltzmann machines (only with feedforward/feedback connections) which were introduced and later on sped up in performance by Hinton, Neal, and Dayan. In successive papers, they included recognition weights (feedforward) and generative weights (feedback) and a 'wake-sleep' algorithm for unsupervised training \cite{HintonDayanFreyNeal1994}.

A variation of a recurrent neural network, which excels in processing sequential information such as text or speech recognition, is the \emph{Long Short Term Memory or LSTM}. It has significant advantages over its preceding competitors like time-delay neural networks (separate weights for multiple discrete times within a sliding window) and recurrent networks with backpropogation through time in the fact that the latter two suffer from inability to learn long-term information - the first due to the fixed window length and the second due to limitations of backpropagation for deep learning \cite{Kurenkov2015}. The LSTM solution by \cite{HochreiterSchmidhuber1997} contains special types of units called Constant Error Carousels (CECs) that have an activation function, an identity function, and a connection to itself with weight of $1.0$.  In this way errors as derivatives of identity (chain rule) don't vanish in the long term and the network also discovers and learns correlations that are very distant in time. CECs are then connected to several nonlinear adaptive units with possibly multiplicative activation functions for learning nonlinear behavior.

A recurring and often criticized issue in all of these models is their biological plausibility \cite{Crick1989}. Since they are purely computational, they are optimized for accuracy on comparable datasets like MNIST, Imagenet, TIMIT, to name a few. Since they don't necessarily model natural phenomena and only draw inspiration from such, they get the engineering freedom of deviating away as much as they need as long as they achieve their original goal. At the same time biological resemblance is what distinguishes neural networks from other machine learning techniques like random forests and support vector machines which also perform with very high accuracy ratios but don't assume such resemblance. In this sense, ANNs are a middle ground between features taken from nature and features taken from engineering ingenuity. The problem of balancing the two makes a large difference in the final performance of each computation model. It is therefore possible that engineering constructions like
\begin{itemize}
	\item the backpropagation learning of an ANN
	\item the identical weights on a region of a fixed size of a CNN
	\item the probabilistic (although not necessarily) feedback connections of a DBN
	\item the CECs of a LSTM
\end{itemize}
could all have more natural and universal explanations. If a computational deviation is necessary, it would still be preferable to try to find the most natural solution possible and explicitly state any unavoidable computational reductions.

Other sources of significant criticism concern the lack of feedback connections in most of these models as well as the phenomenon of \emph{catastrophic interference/forgetting} whereby a completely new input pattern may lead to very fast forgetting of the already learned one. Considering feedback connections is probably important at least since such connections are empirically observed to be approximately the same in number as the feedforward connections. However, most of the ANN models ignore them which implies ignoring a significant part of the overall BNN (biological neural network) dynamics. A group of fully recurrent networks that take a very different direction from DNNs and which we only mention here for completeness are self-organizing or Kohonen maps (SOMs) \cite{Kohonen1982} as well as models from the Active Resonance Theory (ART) \cite{CarpenterGrossberg1988}. The ART proposal also solves the catastrophic interference problem but is statistically inconsistent and uses centralized (instead of parallel and purely local) policies. By local policies we mean policies that make use of entirely local information within a synapse or neuron and not global information about the network state of any kind. A more natural solution would therefore resemble swarm intelligence (SI) where decentralized decisions lead to complex self-organized systems.

\section{Mathematical models of neural networks}

\subsection{Realistic neuron models}

In our review of neuron models we will mainly include their mathematical formulation and sometimes their construction. For any additional mathematical study we will include the main references where this study was performed. Some predictions and matched observations in the cognitive sciences are also mentioned with further directions to their main sources for a deeper investigation.

One of the most important tasks in modelling neurons together with any resulting neural networks is to capture neuron's essential features for a desired cognitive level behavior while choosing among many different levels of biological plausibility. While the belief that the more we pertain to the complex empirical reality the greater the chance that we will reach our goal is not entirely false, in most cases this will fail in practice due to the computational resources for and multiple possible interpretations of the complexity that we have added. However, essential features of the neuron might still reside in the more realistic models even though there is already a wide variety of reduced to computationally minimized models that have made their selections of what to retain and what to lose. For this reason, it is important that we include these neuron models as well.

\subsubsection{Conductance-based models}

The most biologically detailed yet still sufficiently general neuron models are the ones including dynamics arising from the currents that pass through the ion channels in the cell membrane. The membrane conductance then results in action potentials and becomes the main focus of this type of models. The Hodgkin-Huxley neuron introduced in \cite{HodgkinHuxley1990} is a milestone in computational neuroscience and a starting point for many easy extensions and therefore cannot be skipped in any study of conductance-based models of a neuron.
\begin{figure}[!htbp]
\centering
\includegraphics[width=0.65\textwidth]{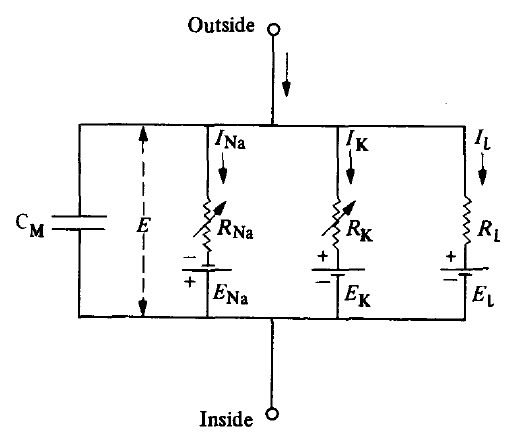}
\caption{Electrical circuit analogue to the Hodgkin-Huxley neuron model as presented in \cite{HodgkinHuxley1990}.}
\label{hh_circuit}
\end{figure}
While studying the giant axon of the squid, Hodgkin and Huxley found three types of current: sodium, potassium, and a leak current of Cl$^-$ ions which also takes care of other channels. Since the membrane lipid bilayer isolates the ions inside and outside the cell body it plays the role of a capacitor. Since the ion pumps lead to difference in the concentration of ions and therefore a nonzero equilibrium potential (Nernst potential), they play the role of a battery. The three currents from the ion channels and the current charging the capacitor comprise the total applied current.
\begin{equation} \label{hh_total_current}
I(t) = I_C(t) + \sum_{k=1}^{3} I_k(t) = I_C(t) + I_{Na}(t) + I_K(t) + I_L(t)
\end{equation}

We can easily construct an equation for a membrane conductance of a passive neuron, i.e. including only the $I_C$ and $I_L$ terms using the above analogies with electrical circuits. From the definition of current $I=dQ/dt$ and the definition of capacity with $Q=VC$ where $Q$ is the charge, $V$ is the voltage, and $C$ is the capacitance, we get that $I_C=dQ/dt=CdV/dt$. From Ohm's law $V-E = \Delta V = IR$ where $E$ is the resting potential (steady state voltage for a passive membrane) and $R$ is the resistance of the leak channel, we get that $I_{L} = \Delta V/R = (V-E)/R$. The passive membrane equation would then follow from the conservation of current $I_C+I_L=0$. For each of the ion channels we have a similar equation to the leak channel and can use conductance as the reciprocal of resistance $g_L=1/R$ to simplify notation with the only difference that the leak channel conductance is a constant while the the other channels can open and close implying conductance that depends on time and voltage. This means that we can simply use their maximum conductances $g_{Na}$ and $g_K$ multiplied by a number between 0 and 1 which will represent the probability of a ion channel to be open. We can then expand the currents and rearrange (\ref{hh_total_current}) like
\begin{equation*}
C\frac{dV}{dt} = I_C(t) = I(t) - (g_{Na}m^3h(V-E_{Na}) + g_Kn^4(V-E_K) + g_L(V-E_L))
\end{equation*}
where the sodium channels have two coefficients (gating variables) $m$ and $h$ since they have activation and inactivation gates and all such coefficients are raised to powers to match experimental data. $E_{Na}$ and $E_K$ are now the reversal potentials and the quantities $V-E_K$ are the driving forces for each ion. To complete the model, we add differential equations to govern the three gating variables which are based on protein kinetics with activation rate $\alpha$ and inactivation rate $\beta$:
\begin{align} \label{cbm1}
\begin{dcases}
& C\frac{dV}{dt} = I(t) - (g_{Na}m^3h(V-E_{Na}) + g_Kn^4(V-E_K) + g_L(V-E_L))\\
& \frac{dm}{dt} = \alpha_m(V) (1-m) - \beta_m m\\
& \frac{dn}{dt} = \alpha_n(V) (1-n) - \beta_n n\\
& \frac{dh}{dt} = \alpha_h(V) (1-h) - \beta_h h
\end{dcases}
\end{align}

This model by Hodgkin and Huxley captures the essence of spike generation by sodium and potassium ion channels but it could easily be extended to other types of ion channels. All a protagonist of biological realism has to do is add an equation for the dynamic of the new ion channel and a term in the main equation sum. A general form system can then be
\begin{align} \label{cbm2}
\begin{dcases}
& C\frac{dV}{dt} = I(t) - \sum_{k=1} I_k(t) = I(t) - \sum_{k=1} g_k m_k^{p_k} h_k^{q_k}(V-E_k)\\
& \frac{dm_k}{dt} = \alpha_{m,k}(V) (1-m_k) - \beta_{m,k} m_k\\
& \frac{dh_k}{dt} = \alpha_{h,k}(V) (1-h_k) - \beta_{h,k} h_k
\end{dcases}
\end{align}
The ion channels that have been added in various models include:
\begin{enumerate}
  \item Sodium channels
  \begin{enumerate}
    \item Fast sodium current $I_{Na}$ is similar to the one in the Hodgkin-Huxley model.
    \item Persistent sodium current $I_{NaP}$ lack the inactivation gate and therefore the variable $h=1$.
  \end{enumerate}
  \item Potassium channels with current $I_{K}$
  \begin{enumerate}
    \item Rapidly inactivating potassium current with constant 1 $I_{A1}$
    \item Rapidly inactivating potassium current with constant 2 $I_{A2}$
    \item Slowly inactivating potassium current with constant 1 $I_{K2a}$
    \item Slowly inactivating potassium current with constant 2 $I_{K2b}$
  \end{enumerate}
  \item Calcium channels with current $I_{Ca^{2+}}$
  \begin{enumerate}
    \item Low threshold calcium current $I_T$ - no inactivation gate
    \item High threshold calcium current $I_L$ - inactivating current where the channel shuts down after depolarization
  \end{enumerate}
\end{enumerate}
Different combinations of these give rise to different dynamics which in some cases may only be quantitatively different but in others also qualitatively \cite{GerstnerKistler2002}. A good example for this is the case of low threshold calcium current which predicts and explains a phenomenon of post-inhibitory rebound, i.e. an action potential arising from an inhibitory input. Another model, this time with the addition of rapidly inactivating potassium current, is known as the Connor-Stevens neuron model \cite{ConnorStevens1971}.
\comment[inline]{PDE version: derivation, bistability, periodicity - no inhibition}

\subsubsection{Multi-compartment models}

A second direction towards greater biological plausibility of a neuron model one could take is to add spatial variance within the cell. Complex neuron geometries and natural differences in the membrane potential across different compartments can bring about different dynamical behavior which can then be accounted for by such models. Such differences in current are especially noticeable along long and thin dendrites. Considering the above models as single-compartment models, the step to make must now be clear - simply separate the neuron into different regions, each governed by the same single-compartment model but with different ion channel concentrations (potential and gating variables) and add longitudinal resistance among the compartments in addition to the transversal resistance of each compartment's ion channels. Of course, this assumes that the variation in potential within a compartment is either negligible or produces a small enough error that can be ignored for interpretation of the results.

An example for a non-branching cable would be
\begin{equation} \label{mccbm1}
C\frac{dV_m}{dt} = I_m(t) - \sum_{k=1} I_k^m + g_{m,m+1} (V_{m+1}-V_m) - g_{m,m-1} (V_{m-1}-V_m)
\end{equation}
where $I_m$ is the current flowing into compartment $m$ and the last two terms are responsible for the interaction with the previous and respectively next compartment (with one of the two terms missing in the case of the cable ends). The constant $g_{m,m'}$ determines the conductance (and therefore resistance) from compartment $m$ to compartment $m'$. The sum of currents can be replaced with the ion currents similarly to the way we did it in (\ref{hh_total_current}) but with different channel conductance for each compartment. This can then be extended to various branching-based coupling by adding removing terms similar to the last two terms \cite{DayanAbbot2005}.

Multi-comparment models can also be used with simpler single-compartment models like the LIF neuron introduced in later sections which could leverage some of the computational challenges they pose when increasing the comparmental resolution to simulate a complex dendritic structure. A further major disadvantage of such multi-compartment models is the large number of free parameters where multiple combinations of very different values can produce the same plausible biological behavior \cite{TaylorGoaillardMarder2009}.

The best approximation of the potential dynamics on a long and narrow dendritic cable can be reached by making a step beyond such multi-compartment models and into cable theory using a nonlinear diffusion equation called the cable equation \cite{Rall2011}. The best approximation of the potential dynamics along an entire dendritic tree can then be achieved with multi-compartment models that use the cable equation instead of a number of ODEs (simpler approach above), thus accounting for spatial variations within a single-compartment model as well. Similar models are sometimes referred to as Rall-type models and some of them made interesting and true predictions like the dendro-dendritic spikes \cite{RallShepherdReeseBrightman1966} but they are beyond the scope of our review.

\subsubsection{Reduced conductance-based models}

A major advantage of the very biologically detailed models proposed above is the easy interpretation of all quantities and comparability with experimental data. However, a major drawback for the analysis of networks of such neurons is the computational challenges they pose for simulation as well as the complexity of their interpretation on a network-wide scale. One major attempt to amend this is the reduction of the dimensionality of the dynamical system representing a full conductance-based model into usually 2-dimensional such with one variable that could be loosely related with the membrane potential and one variable which is usually called a "relaxation" variable.

The first and currently classical such reduction is the Fitzhugh-Nagumo neuron
\begin{align} \label{rcbm1}
\begin{dcases}
& \dot{u} = c(v + u - u^3/3 + z)\\
& \dot{v} = -(u - a + bv) / c\\
& 1-2b/3 < a < 1, 0 < b < 1, b < c^2
\end{dcases}
\end{align}
which can be obtained from the damped oscillator
\begin{equation*}
\ddot{u} + k \dot{u} + u = 0
\end{equation*}
by replacing the damping constant with a quadratic term of $u$ to obtain a Van der Pol oscillator
\begin{equation*}
\ddot{u} + c (u^2 - 1) \dot{u} + u = 0
\end{equation*}
and then performing a Lienard transformation \cite{Fitzhugh1961}. The two unknowns are $u$ for membrane potential (or simply voltage) and $v$ for a recovery variable (also called accommodation or refractoriness), $a$, $b$, and $c$ are constant coefficients within certain bounds and $z$ represents an external input stimulus.

Phase plane analysis shows that this model retains the qualitative features of the Hodking-Huxley model (\ref{cbm1}) namely threshold-spike dynamics, rafractoriness period, spike accommodation and other physiological phenomena observed at the cardiac muscle excitation. When perturbed sufficiently far from its equilibrium point at the resting potential, the neuron will essentially perform an excursion through the phase space and return to this equilibrium point. This trajectory then represents the action potential of the neuron. In addition to its retained properties in lower dimensionality, (\ref{rcbm1}) also shows that (\ref{cbm1}) is a particular member of a large class of nonlinear systems showing excitable and oscillatory behavior.

Further reductions build on top of (\ref{rcbm1}) in various ways. Most widely used as such is the Hindmarsch-Rose neuron
\begin{align} \label{rcbm2}
\begin{dcases}
& \dot{u} = v - F(u) + z - w\\
& \dot{v} = G(u) - v\\
& \tau \dot{w} = H(u) - w
\end{dcases}
\end{align}
where $z$ retains the meaning if input stimulus, $F(u)$ is a cubic and $G(u)$ is a quadratic polynomial of $u$ both generalized from the Fitzhugh-Nagumo equations but with addition of a third unknown $w$ asymptotically converging towards a linear function $H(u)$ \cite{RoseHindmarsh1989}. The model (\ref{rcbm2}) could reproduce all complex neuron behavior presuming we have found correct functions for $F$, $G$, and $H$ \cite{Izhikevich2004}.

Yet another model that has become popular due its exact and measurable parameters is the Morris-Lecar neuron \cite{MorrisLecar1981} and a recent one that represents very well all membrane potential dynamics while remaining very computationally efficient is the Izhikevich neuron namely
\begin{align} \label{rcbm3}
\begin{dcases}
& \dot{u} = 0.04u^2 + 5u + 140 - v + z\\
& \dot{v} = a(bu - v)
\end{dcases}
\end{align}
with the auxiliary after-spike resetting
\begin{align*}
\text{if } u \geq 30\text{mV then }
\left\{
	\begin{array}{ll}
		u \leftarrow c \\
		v \leftarrow v + d
	\end{array}
\right.
\end{align*}
where $u$ and $v$ are the membrane potential and recovery variable as usual \cite{Izhikevich2003, Izhikevich2004}. According to calculations in \cite{Izhikevich2004}, all reduced conductance-based (also called Hodgkin-Huxley type) models described here are performing at least an order of magnitude better than the original Hodgkin-Huxley model (\ref{cbm1}) when it comes to simulating a time window of 1ms (calculated in floating point operations per second or FLOPS). At the same time most of them are capable of reproducing the larger part of the most prominent features of biological spiking neurons. The least amount of FLOPS for the simulation is held by the minimal neuron models that we review in the next part.

\subsection{Minimal neuron models}

The computationally most effective neuron models usually retain only threshold and sometimes subthreshold dynamics of the membrane potential. They do this in an abstract fashion excluding entirely the electrophysiological dynamics that cause the (sub)threshold behavior on the first place. Due to the simple form of their equations, we usually immediately decompose the soma input into a sum of multiple presynaptic contributions and specify the nature of these contributions as well as the neuron output.

There are various assumptions about the nature of the communicated signal or neural code \cite{DayanAbbot2005}. Neurons communicate via action potentials as nearly identical pulses, thus the most realistic variant of communicating neurons and their resulting networks would be to preserve this feature. If no significant information is encoded into the particular times and frequencies of such spike trains, we could simplify them by taking averages in time and deriving spike-count rates. The resulting neuron models are called firing-rate models. If no significant information is encoded in a local population of neurons, i.e. they have a nearly identical response and similar connectivity, we can take averages within a population and deriving population rates. The resulting neuron models are called population models. Clearly, the absence of time coding (mutually independent spikes) and/or population coding (mutually independent firing probabilities for individual neurons) is a strong assumption with some experiments that already contradict them \cite{DayanAbbot2005}.

\subsubsection{Spiking (pulsed) models}

Spiking neuron models are believed to comprise the third generation of neural networks after the rate-based or continuous second generation (reviewed next) and the binary first generation (reviewed last) \cite{Maass1997}. Despite these bold claims they have had some difficulty in finding their predicted application in reproducing cognitive phenomena on a sufficiently high level, mostly due to computational restrictions as well.

One of the major neuron models which are minimized in terms of the biophysical mechanisms but bring about action potentials is the leaky integrate and fire model first introduced by \cite{Lapicque1907}. It is very limited in terms of physiological complexity but reproduces both the threshold phenomena and the subthreshold dynamics of a neuron fairly well.

The derivation is similar to and simpler than the derivation of (\ref{cbm1}). From the conservation of charge we have this time for $k=1$
\begin{equation*}
I(t) = I_C(t) + \sum_{k=1}^{1} I_k(t) = I_C(t) + I_L(t) = C \frac{dV}{dt} + g_L(V-E_L)
\end{equation*}
and taking simpler resting potential $E_L=0$ together with $g_L=1/R$
\begin{equation*}
I(t) = C \frac{dV}{dt} + \frac{(V-0)}{R} = C \frac{dV}{dt} + \frac{V}{R}
\end{equation*}
Finally, multiplying with $R$ and replacing $\tau = R C$ as the time scale constant of the leaky integrator
\begin{equation*}
R I(t) = R C \frac{dV}{dt} + V = \tau \frac{dV}{dt} + V
\end{equation*}
The final leaky integrate and fire neuron then reads
\begin{equation} \label{lifm1}
\tau \frac{dV}{dt} = - V(t) + R I(t)
\end{equation}
with the additional condition that when V reaches a threshold value $V_{threshold}$ it will be reset back to a value $V_{reset}$ to take care of the threshold dynamic of the neuron. After resetting, the model is governed by the same ODE until another discrete fire/reset event occurs. The combination of the fire condition and the leaky integration (decaying to $R I$ steady state) defines the leaky integrate and fire neuron model.

One major problem with the LIF neuron is oversimplification and the lack of reproducibility of rather important dynamics like refractoriness (the neuron can be re-excited for an arbitrarily small interval after the last fire event for sufficiently large input) and spike-rate adaptation (the neuron will not reduce its firing frequency given a long constant input which is observed empirically). On the positive side, the model is simple enough so that it is possible to find its explicit solution for constant input which is a nice and rare mathematical property to have. This derivation together with more elaboration on the model's upsides and downsides can be found in \cite{DayanAbbot2005}.
\comment[inline]{PDE version: weak form, blow-up, asymptotics, properties, well-posedness}

A straightforward extension of the LIF model could be a general nonlinear integrate and fire model like
\begin{equation} \label{lifm2}
\tau \frac{dV}{dt} = - F(V(t)) + G(V(t)) I(t)
\end{equation}
with the same reset condition \cite{GerstnerKistler2002}. A special case of (\ref{lifm2}) is the quadratic integrate and fire model
\begin{equation} \label{lifm3}
\tau \frac{dV}{dt} = - C(V(t)-V_{reset})(V-V_C) + R I(t)
\end{equation}
which under a change of variable can also be called the theta model of a biological neuron (with very interesting stability analysis on the unit circle, also included in \cite{GerstnerKistler2002}).

The second major neuron model which is minimized in terms of the biophysical mechanisms that bring about action potentials is the spike response model or SRM \cite{Gerstner1995}. It is represented by the integral equation
\begin{equation} \label{srm}
u_i(t) = \eta(t-t_i) + \sum_j w_{ij} \sum_k \epsilon_{ij} (t - t_i, t - t_j^{k}) + \int_0^\infty \kappa(t-t_i,s)I(t-s)ds
\end{equation}
where a neuron indexed by $i$ has a single quantity $u_i$ with resting value $u_{rest}=0$ responding to a presynaptic spike at a fixed time $t_i$ and an external input current $I(t)$ (the last term). The presynaptic neurons with respect to neuron $i$ are indexed by $j$, have strengths $w_{ij}$ and each one has spikes indexed by $k$ at times $t_j^k<t$. The time evolution of the response to an incoming spike is represented by $\epsilon$ while the form of the action potential and the after potential is described by $\eta$. An output spike is then triggered if after the summation of presynaptic neurons and their spikes at previous times, the value $u_i$ reaches a (not necessarily fixed) threshold $\alpha=\alpha(t-t_i)$. Here $t_i$ is fixed and is the time of the last action potential
\begin{equation*}
t_i = \max{t_i^k < t}
\end{equation*}

To add refractory period $\Delta$, we can simply set the threshold to a very high value until time $t_i + \Delta$. The notation can be simplified if we introduce the total PSP quantity
\begin{equation*}
h(t|t_i) = \sum_j w_{ij} \sum_k \epsilon_{ij} (t - t_i, t - t_j^{k}) + \int_0^\infty \kappa(t-t_i,s)I(t-s)ds
\end{equation*}
in which case the model becomes
\begin{equation} \label{srm2}
u_i(t) = \eta(t-t_i) + h(t|t_i)
\end{equation}
Further interpretation of the SRM and its properties is presented in \cite{GerstnerKistler2002}.

The integrate-and-fire (LIF) neuron is a special case of the spike response (SRM) neuron. To see this, we need to take $V_i=u_i$ as the potential of the $i$-th neuron and generalize the input current of (\ref{lifm1}) to
\begin{equation} \label{lif2srm}
\tau \frac{du_i}{dt} = - u_i(t) + R I(t) = - u_i(t) + R (\sum_j w_{ij} \sum_k \beta(t-t_j^k) + I_i(t))
\end{equation}
where $I(t)$ is external current applied to the soma and $\beta$ are postsynaptic current pulses at times $t_j^k$. More details about this comparison can also be found in $\cite{GerstnerKistler2002}$.

\subsubsection{Firing-rate (rate-based) models}

The most detailed way to represent interneuron communication and therefore a neural network is by preserving the neuron spikes used for communication among nodes since there could be timing-related patterns like synchronized and correlated firing and therefore dynamics which can only be captured by this type of communication. However, if we assume that this is not the case, we can greatly reduce computational costs due to number of parameters and time scales as well as the difficulty of any analytic interpretation, all by replacing the spikes with firing rates.

Variations in the action potential's duration, amplitude, and shape are very small and are typically ignored in models of spiking networks which consider them as identical events occurring at single time instances $0 \leq t_i \leq T$. The neuron response function $\rho(t)$ can then be defined as
\begin{equation} \label{nrfun}
\rho(t)=\sum_{i=1}^{n}\delta(t-t_i)
\end{equation}
and can be used to obtain any sequence of responses of amplitude defined by $h(t)$ as
\begin{equation}
\sum_{i=1}^{n}h(t-t_i) = \int_{-\infty}^{+\infty}\rho(t-\tau)h(\tau)d\tau = \rho \star h
\end{equation}
We can derive the firing rate (which in this case means a spike-count rate) from the neuron response function (\ref{nrfun}) with a continuum assumption as
\begin{equation} \label{frate}
r(t) = \lim_{\Delta t \to 0} \frac{1}{\Delta t} \int_{t}^{t+\Delta t}\rho(\tau)d\tau
\end{equation}
or alternatively from $\langle \rho(t) \rangle$ averaged from many experimental trials where the neuron is subjected to the same stimulus. In this sense, $r(t)$ can also be interpreted as the probability density of firing.

The most basic firing rate model with current dynamics is constructed by \cite{DayanAbbot2005}:
\begin{align} \label{frm1}
\begin{dcases}
& \tau_s \frac{dI_s}{dt} = -I_s + \sum_{i=1}^{N_u} w_i u_i = -I_s + \mathbf{w} \cdot \mathbf{u}\\
& v = F(I_s) = (I_s - \alpha)^+
\end{dcases}
\end{align}
where $u$ is used for input firing rate and $v$ for output firing rate with vectors $\mathbf{u}$ and $\mathbf{v}$ for collections of input and output units. The $w_i$ coefficients are the weights/strengths of the synapses that receive input firing rates $u_i$ and are constants here but generally could depend on time to account for neuronal plasticity. The first equation then relates the total input current at the soma $I_s$ to the firing rates received from a total of $N_u$ dendrites. It is a simple dynamic equation with a decay to a steady state at the total input firing rate. The second equation relates the output firing rate $v$ with the total input current through an activation function $F$ which is usually taken to be the positive part function $w_+(x):=\max(x,0)$. This is an instant equation with $\alpha$ playing the role of a threshold constant.

It is important to note that (\ref{frm1}) is simplified and somewhat collapsed version of multiple neurophysiological phenomena:
\begin{enumerate}
  \item The amplitude and sign of the synaptic input $i$ are determined by $w_i$ which also absorbs the probability of a neurotransmitter release from a presynaptic terminal.
  \item The firing rate $u_i$ is the time average of the actual spike train as described in the definition of $r(t)$ and encompasses all the dynamics of the synaptic conductance of the presynaptic spike like active and passive properties of the dendritic cables, linear summation of multiple spikes, etc.
  \item The activation function $F$ takes care of converting units of $I_s$ into units of firing rate through an extra constant multiplier, rendering the weights $w_i$ dimensionless in the final model.
\end{enumerate}

If the second equation is not assumed to be instant, i.e. the dependence of the output firing rate on the soma current is dynamic, the firing rate model extends to
\begin{align} \label{frm2}
\begin{dcases}
& \tau_s \frac{dI_s}{dt} = -I_s + \sum_{i=1}^{N_u} w_i u_i = -I_s + \mathbf{w} \cdot \mathbf{u}\\
& \tau_r \frac{dv}{dt} = -v + (I_s - \alpha)^+
\end{dcases}
\end{align}
where $\tau_r$ is the time scale determining how rapidly $v(t)$ can approach its steady state when considering constant input or can follow the changes in $I_s$ otherwise.

Finally, considering large difference in the two time scales $\tau_s$ and $\tau_r$ and in particular $\tau_r \gg \tau_s$, we can perform quasi-steady state approximation with the assumption of $I_s \approx const$ with respect to the second equation to obtain the single equation firing rate model
\begin{equation} \label{frm3}
\tau_r \frac{dv}{dt} = -v + (\mathbf{w} \cdot \mathbf{u} - \alpha)^+
\end{equation}

\subsubsection{Population models}

The firing rates used in the previous section are spike-count rates which work well under the assumption that the time averaging window is very small with respect to the time scale of any change in the input. This is not the case in many practical situations where the reaction times are short. However, we could use a firing rate also as an average over population instead of average over time. We then obtain a generalized neuron which represents a local population aggregate and interactions with other generalized neurons are equivalent to interactions with other subpopulations.

First and most important population model to look at is the Wilson-Cowan model \cite{WilsonCowan1972}. A local population aggregate is assumed to have similar properties and connections which is confirmed from observations about nearly identical responses within relatively small volumes of brain tissue and the presence of local redundancy. Under the additional assumption of random but dense local connectivity and close spatial proximity, we can neglect spatial interactions and deal simply with the temporal dynamics of a population aggregate and more specifically with the portion of the aggregate which is active per unit time.

Considering excitatory and inhibitory populations then leads to two such unknowns or respectively $E(t)$ and $I(t)$ with 0 as resting potential state. To obtain the model then, we have to derive expressions for the proportion of cells which are sensitive (and not refractory) and the cells which receive at least a threshold excitation. If refractory cells have refractory period of $r$ msec their proportion and respectively the proportion of sensitive cells are
\begin{equation} \label{wc_active}
\int_{t-r}^{t} E(t')dt' \Rightarrow 1 - \int_{t-r}^{t} E(t')dt'
\end{equation}
and similarly for the inhibitory population. The cells that receive at least a threshold excitation are given by the functions $S_e$ and $S_i$ which are called subpopulation response functions and represent the expected proportion of cells in a subpopulation which would respond to a given level of excitation if none of them were originally refractory. These functions can be defined from a distribution of per-cell thresholds $D(\theta)$ or per-cell synapse numbers $C(\omega)$ as
\begin{equation} \label{wc_distributions}
S(x) = \int_{0}^{x(t)} D(\theta)d\theta \qquad\text{or}\qquad S(x) = \int_{\theta/x(t)}^{\infty} C(\omega)d\omega
\end{equation}
where per-cell threshold distribution assumes constant synapse number and vice versa and $x(t)$ is the average excitation that all cells are subjected to (reasonable assumption for sufficiently large number of synapses). In the second case all cells with at least $\theta/x(t)$ synapses will receive threshold excitation which together with the desire to make $S(x)$ a sigmoidal function (0 and 1 for no and full subpopulations as asymptotic values) is the reason for the integral bounds.

As a third multiplier in the final equations, we should add an average level of excitation generated in an excitatory or resp. inhibitory cell at time $t$
\begin{align} \label{wc_average}
A_E = \int_{-\infty}^{t} \alpha(t-t')(c_1 E(t') - c2 I(t') + P(t')) dt'\\
A_I = \int_{-\infty}^{t} \alpha(t-t')(c_3 E(t') - c4 I(t') + Q(t')) dt'
\end{align}
which we can obtain if we assume the the effect of stimulation decays with rate $\alpha$, each cell sums its input synapses with average number of excitatory and inhibitory synapses given by the constants $c$, and external input $P(t)$ or $Q(t)$. From (\ref{wc_active}), (\ref{wc_distributions}), and (\ref{wc_average}), the Wilson-Cowan model now follows:
\begin{align} \label{pm1}
\begin{dcases}
& E(t+\tau) = [1 -\int_{t-r}^{t} E(t') dt'] S_e A_e\\
& I(t+\tau') = [1 -\int_{t-r'}^{t} I(t') dt'] S_i A_i
\end{dcases}
\end{align}
Here we take the proportion of cells which are both sensitive and above threshold at time $t$. Both these events are assumed to be uncorrelated with mutually independent probabilities of each.

\comment[inline]{PDE version: something general first, dynamical system, Fokker-Planck equation, numerical (FVM) approximation, asymptotics, evolution, well-posedness, generalized relative entropy, consequences, QSSA and reduction to 1D Fokker-Planck}

Other population models have been developed with LIF as well as SRM neurons as comprising units. These result in PDEs tracking the change in neuron state across time and space which in the first case is a membrane potential density and in the second case is a rafractory density (due to the state of refractoriness of an SRM neuron) \cite{GerstnerKistler2002}.

Besides Wilson-Cowan, there are other integral formulations of the population dynamics, one particularly by Gerstner:
\begin{equation} \label{pm2}
A(t) = \int_{-\infty}^{t} P_I(t|t') A(t') dt'
\end{equation}
where $P_I(t|t')$ is a kernel representing the probability density that a neuron fires at time $t$ with an input $I(t')$ given that it's last spike was at $t'$ and $A(t)$ is the population activity at a given time instance \cite{Gerstner2000}. This equation can be derived under the assumptions that
\begin{enumerate}
	\item The total number of neurons in the population remains constant.
	\item The neurons exhibit no adaptation, i.e. the state of neuron $i$ depends only on the most recent firing at time $t_i'$.
	\item The neurons are independent on a sufficiently small time scale, i.e. the number of spikes in the interval $(t,t+\Delta)$ which is of this time scale for each neuron is an independent random variable. Then by the law of large numbers, for a sufficiently large network they converge in probability to their expectation values which are the ones we will consider.
\end{enumerate}

The second assumption implies that the probability of a neuron which has fired at $t'$ to fire again at $\hat{t} \in [t',t]$ is given by $\int_{t'}^{t} P_I(s|t')ds$. Then define the probability that a neuron which has fired at $t'$ survives without firing until $t$ as $S_I(t|t') = 1 - \int_{t'}^{t} P_I(s|t')ds$.

The third assumption helps us consider the model in the thermodynamic limit of $N \rightarrow \infty$ where $N$ is the size of the population. Then the average of the fraction of neurons at time $t$ which have fired their last spike at $\hat{t} \in [t_0,t_1]$, $t_0<t$, $t_1<t$ is $\int_{t_0}^{t_1} S_I(t|t') A(t') dt'$ where $A(t')\Delta t'$ is the fraction of neurons that have fired (not necessarily last) in $\hat{t} \in [t', t'+\Delta t']$ and $S_I(t|t')$ of these are expected to survive from $t'$ to $t$ without firing.

Finally, the first assumption together with an assumption that all neurons have fired at some point in the past (including $-\infty$ for all that haven't fired in a finite time) implies that if we extend the lower bound to $-\infty$ and the upper bound to $t$ we will obtain
\begin{equation} \label{pm2_form1}
1 = \int_{-\infty}^{t} S_I(t|t') A(t') dt'
\end{equation}
which will hold $\forall t$ because the number of neurons is constant and all of them have fired in the interval $[-\infty,t]$.

Differentiating (\ref{pm2_form1}) and using the facts that by definition of $S_I(t|t')$, $P_I(t|t')=-\frac{d}{dt}(S_I(t|t'))$ and trivially that $S_I(t|t)=1$, we arrive at
\begin{equation*}
0 = S_I(t|t) A(t) + \int_{-\infty}^{t} \frac{d}{dt} S_I(t|t') A(t') dt' = A(t) - \int_{-\infty}^{t} P_I(t|t') A(t') dt'
\end{equation*}
The equation (\ref{pm2}) is highly nonlinear since the kernel $P_I(t|t')$ depends nonlinearly on the total input current which in turn depends nonlinearly on the population activity $A(t)$.

\subsubsection{Binary models}

The absolute minimal neuron model worth mentioning here for completeness is the McCulloch-Pitts (MCP) neuron \cite{McCullochPitts1943} also discussed in the ANN introduction. In fact, it is so biologically unrealistic that it belongs to the separate category of artificial neurons. This neuron is basically a threshold function (a threshold $\theta$ gate in terms of digital logic) like
\begin{equation} \label{dm1}
v =
\left\{
	\begin{array}{ll}
		1 \text{ if } \sum_i u_i > \theta \\
		0 \text{ otherwise}
	\end{array}
\right.
\end{equation}
and it is one of the few artificial neurons that are often used in ANNs. It is also called a perceptron (more specifically when adding weights $w_i$ to the terms in the sum) which is also used as a name of all networks derived from this neuron (single or multilayer perceptron as MLP). Such networks are the first generation of neuron network models according to \cite{Maass1997}.

\subsection{Extension to networks}

Any neuron model studied previously can be coupled with other neurons of the same or even different type to form a network with shared dynamics like oscillations, synchronization, and possibly chaotic behavior \cite{ErichsenBrunnet2008, AmbrosioAziz2012, HandaSharma2016}. However, the number of neurons that we could couple in a computationally feasible manner depends significantly on the complexity of the neuron model with multi-compartment models in one end of the spectrum and minimal models in the other. Studying networks of possibly simplified neurons as principal units instead of detailed single neuron models is motivated by the fact that there are phenomena emerging only from couplings of neuron dynamics that cannot be reproduced otherwise. General cognitive capabilities like perception, learning, or memory recall are all within this category.

\subsubsection{Activity - study of the potential dynamics}

\begin{figure}[!htbp]
\centering
\includegraphics[width=0.9\textwidth]{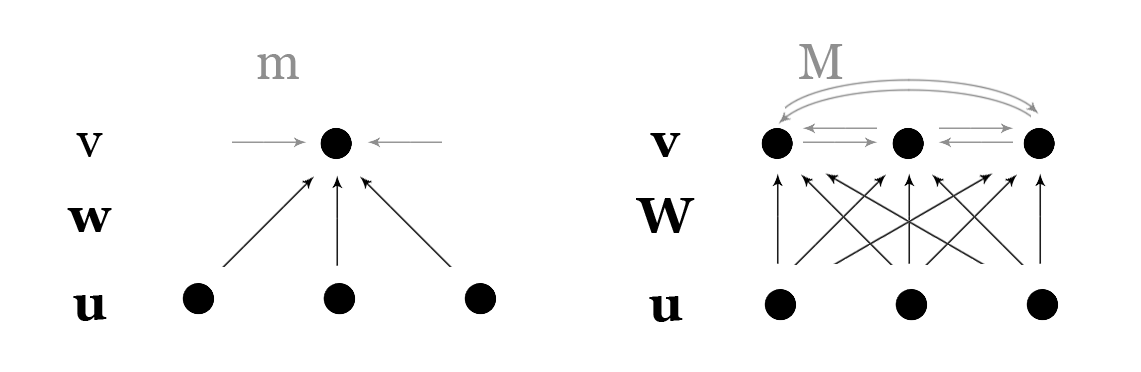}
\caption{Construction of network models from single neuron models using vector notation.}
\label{neuron2network}
\end{figure}

We will construct the various types of networks (feedforward, recurrent, etc.) from the firing-rate neurons (\ref{frm1}, \ref{frm3}), eventually taking the middle ground in the choice of minimal neuron models. The basic idea behind the construction of these networks is shown on figure \ref{neuron2network}. 
For the feedforward network we only have to generalize (\ref{frm3}) to a vector of unknowns therefore obtaining a matrix of weights $\mathbf{W}$, a constant vector of thresholds $\boldsymbol{\alpha}$, and two layers of units  - $\mathbf{u}$ inputs and $\mathbf{v}$ outputs. For layered recurrent networks with feedforward and lateral connection only (no feedback connections), we also add interactions within the layer $\mathbf{v}$ of outputs where $\mathbf{M}$ contains the strengths or weights of the lateral connections. For separate handling of excitatory and inhibitory interactions which we should require if we assume Dale's law (units can either be only excitatory or only inhibitory with respect to others), we only need to split the units within the output layer ($\mathbf{v}$) into excitatory $\mathbf{v_E}$ and inhibitory $\mathbf{v_I}$ and the lateral connections ($\mathbf{M}$) into $\mathbf{M_{EE}}$ and $\mathbf{M_{IE}}$ with weights greater than or equal to 0 and $\mathbf{M_{EI}}$ and $\mathbf{M_{II}}$ with weights less than or equal to 0.

In the case of fully recurrent networks, the inputs $\mathbf{u}$ would be unknowns depending on the outputs $\mathbf{v}$ through feedback connections. However, we don't need to add more equations and a matrix for the weights of the feedback connections. What we can do instead is use to our advantage the fact that adding feedback connections would then include all possible types of connections and there will be no need of differentiating among them anymore. Thus, we simply collapse all classes into one and consider some of the $v_i$ neurons as input units and the rest as hidden units allowing for any possible connection between each two $v_i$ and $v_j$. To generalize the resulting models, we should add a term $\mathbf{U}$ to represent external input (sometimes called bias or control/somatic input/current) to each neuron. Finally, from (\ref{frm1}) we can derive a second type of recurrent network called additive network which reflects the dynamics of total input current or membrane potential rather than the output of a neuron.

Performing all of the constructions above, we obtain the following firing rate network models:
\begin{enumerate}
	\item Feedforward networks
		\begin{equation} \label{fr_ffn}
		\tau_r \frac{d\mathbf{v}}{dt} = -\mathbf{v} + (\mathbf{W} \cdot \mathbf{u} - \boldsymbol{\alpha})^+
		\end{equation}
	\item Recurrent networks
	\begin{enumerate}
		\item Layered networks with feedforward and lateral connections
			\begin{equation} \label{fr_rn_lln}
			\tau_r \frac{d\mathbf{v}}{dt} = -\mathbf{v} + (\mathbf{W} \cdot \mathbf{u} + \mathbf{M} \cdot \mathbf{v} - \boldsymbol{\alpha})^+
			\end{equation}
		\item Excitatory-inhibitory networks
			\begin{align} \label{fr_rn_ein}
			\begin{dcases}
			& \tau_E \frac{d\mathbf{v_E}}{dt} = -\mathbf{v_E} + (\mathbf{W_E} \cdot \mathbf{u} + \mathbf{M_{EE}} \cdot \mathbf{v_E} + \mathbf{M_{EI}} \cdot \mathbf{v_I} - \boldsymbol{\alpha})^+\\
			& \tau_I \frac{d\mathbf{v_I}}{dt} = -\mathbf{v_I} + (\mathbf{W_I} \cdot \mathbf{u} + \mathbf{M_{IE}} \cdot \mathbf{v_E} + \mathbf{M_{II}} \cdot \mathbf{v_I} - \boldsymbol{\alpha})^+
			\end{dcases}
			\end{align}
		\item Fully recurrent networks
		\begin{enumerate}
			\item Output-based networks
				\begin{equation} \label{fr_rn_frn1}
				\tau_r \frac{d\mathbf{v}}{dt} = -\mathbf{v} + (\mathbf{U} + \mathbf{M} \cdot \mathbf{v} - \boldsymbol{\alpha})^+
				\end{equation}
			\item Additive networks
				\begin{equation} \label{fr_rn_frn2}
				\tau_s \frac{d\mathbf{I}}{dt} = -\mathbf{I} + \mathbf{U} + \mathbf{M} \cdot (\mathbf{I} - \boldsymbol{\alpha})^+
				\end{equation}
		\end{enumerate}
	\end{enumerate}
\end{enumerate}
Due to excessive liberty in the interpretation of $\mathbf{U}$, the fully recurrent networks (\ref{fr_rn_frn1}, \ref{fr_rn_frn2}) can also be viewed as generalizations of all the rest. In particular, interpreting the external input $U_i$ as the total feedforward input for the neuron $v_i$ from another layer $\mathbf{u}$, we can define $\mathbf{U} = \mathbf{W} \cdot \mathbf{u}$ and obtain (\ref{fr_rn_lln}) from (\ref{fr_rn_frn1}). In addition, setting the lateral connection weights to zero ($\mathbf{M}=\mathbf{0}$) can give us a feedforward network (\ref{fr_ffn}) and fixing the sign of the weights $\mathbf{M}$, setting $\mathbf{U} = [\mathbf{U_E}, \mathbf{U_I}]^T$, and assuming the same time scale $\tau_E=\tau_I=\tau_r$ can achieve the separation of interactions in (\ref{fr_rn_ein}), all starting from (\ref{fr_rn_frn1}). In general, (\ref{fr_rn_frn1}) and all models derivable from it as special cases are models based on the neuron output dynamics. In contrast, (\ref{fr_rn_frn2}) considers the total input current $\mathbf{I}$ as the dynamical variable which can be interpreted also as thresholded potential with the neuron output firing rate given as $\mathbf{v}=(\mathbf{I} - \boldsymbol{\alpha})^+$. While (\ref{fr_rn_frn2}) can be obtained from (\ref{fr_rn_frn1}) if the decay coefficients (in this case 1) are equal and (\ref{fr_rn_frn1}) from (\ref{fr_rn_frn2}) if the weight matrix $\mathbf{M}$ is invertible, the two are not equivalent in general \cite{Hirsch1989}. Until the end of this section, we will perform some mathematical analysis on these network models that will reveal interesting cognitive level phenomena they reproduce and will visualize different mathematical techniques used for their study.

We can find an explicit solution for a linear version of (\ref{fr_rn_lln}) with $\boldsymbol{\alpha} = \mathbf{0}$ and through some technical assumptions extend our conclusions about their stability and cognitive capabilities to the nonlinear case \cite{DayanAbbot2005}. For this purpose, it is necessary to assume the the matrix $\mathbf{M}$ is symmetric. In particular, consider the network
\begin{equation} \label{fr_rn_linear}
\tau_r \frac{d\mathbf{v}}{dt} = -\mathbf{v} + \mathbf{W} \cdot \mathbf{u} + \mathbf{M} \cdot \mathbf{v}
\end{equation}
The symmetry of $\mathbf{M}$ implies that its eigenvalues $\lambda_i$ are real and therefore can be ordered. The eigenvectors $\mathbf{e_i}$ (satisfying $\mathbf{M} \cdot \mathbf{e_i} = \lambda_i \mathbf{e_i}$) associated with any two distinct eigenvalues are orthogonal (or even orthonormal if normalized, i.e. $\mathbf{e_i} \cdot \mathbf{e_j} = \delta_{ij}$ for $\lambda_i \neq \lambda_j$) and the eigenvectors associated with an eigenvalue with greater multiplicity are linearly independent (so we can still choose a basis of orthonormal vectors). This implies that we can express and replace $\mathbf{v}$ in (\ref{fr_rn_linear}) as
\begin{equation*}
\mathbf{v}(t) = \sum_{i=1}^N c_i(t) \mathbf{e_i} \Rightarrow \tau_r \sum_{i=1}^N \frac{dc_i}{dt} \mathbf{e_i} = -\sum_{i=1}^N (1-\lambda_i) c_i(t) \mathbf{e_i} + \mathbf{U}
\end{equation*}
where $N = \dim(\mathbf{v})$ and split into decoupled ODEs after a dot product with $\mathbf{e_j}$ and using the orthogonality property
\begin{equation*}
\tau_r \frac{dc_j}{dt} = -(1-\lambda_j) c_j(t) + \mathbf{e_j} \cdot \mathbf{U} \Rightarrow c_j(t) = \frac{\mathbf{e_j} \cdot \mathbf{U}}{1-\lambda_j} (1 - e^{(-\frac{t(1-\lambda_j)}{\tau_r})}) + c_j(0) e^{(-\frac{t(1-\lambda_j)}{\tau_r})}
\end{equation*}
This analytic solution could be used to identify $\lambda_j < 1$ as the region of stability where the system approaches the steady state
\begin{equation*}
\mathbf{v}(t) = \sum_{i=1}^N c_i(t) \mathbf{e_i} \rightarrow \mathbf{v_\infty} = \sum_{i=1}^N \frac{\mathbf{e_i} \cdot \mathbf{U}}{1-\lambda_i} \mathbf{e_i}
\end{equation*}
The steady state solution hence involves the projection of the input vector $\mathbf{U}$ onto the eigenvectors of the recurrent matrix $\mathbf{M}$ and is amplified by a factor of $1/(1-\lambda_i) > 1$ along the $i$-th dimension if $0 < \lambda_i < 1$. This leads to some cognitive capabilities like selective amplification when a subset of eigenvectors $\mathbf{e_k}$ have eigenvalues $\lambda_k$ less than but close to $1$ in comparison to the rest. In this case, they dominate the sum and the steady state solution becomes
\begin{equation*}
\mathbf{v_\infty} \approx \sum_{k=1}^K \frac{\mathbf{e_k} \cdot \mathbf{U}}{1-\lambda_k} \mathbf{e_k} + \sum_{k=K+1}^N \frac{\mathbf{e_k} \cdot \mathbf{U}}{1-\lambda_k} \mathbf{e_k} \approx \sum_{k=1}^K \frac{\mathbf{e_k} \cdot \mathbf{U}}{1-\lambda_k} \mathbf{e_k}
\end{equation*}
and the network amplified the projection of the input into the subspace spanned by all $\mathbf{e_k}$. Another cognitive capability of such networks is input integration which can be seen if we consider the eigenvalues $\lambda_k = 1$ in which the equations for $c_k$ become
\begin{equation*}
\tau_r \frac{dc_k}{dt} = -(1-1) c_k(t) + \mathbf{e_k} \cdot \mathbf{U} = \mathbf{e_k} \cdot \mathbf{U} \Rightarrow c_k(t) = c_k(0) + \frac{1}{\tau_r} \int_0^t \mathbf{e_k} \cdot \mathbf{U}(t') dt'
\end{equation*}
Now $c_k$ can remember previous input in the sense that if an input $\mathbf{U}(t)$ is nonzero initially but zero for an extended period afterwards, all the remaining $c_i$ where $\lambda_i < 1$ will settle to zero and we will have
\begin{equation*}
\mathbf{v}(t) \approx \sum_{k=1}^K \frac{1}{\tau_r} \int_0^t \mathbf{e_k} \cdot \mathbf{U}(t') dt' + 0 \approx \sum_{k=1}^K \frac{1}{\tau_r} \int_0^t \mathbf{e_k} \cdot \mathbf{U}(t') dt'
\end{equation*}
where we have assumed $c_j(0)=0$, $\forall j=1, \cdots N$. This means that the projection integrals will be preserved in the absence of input and could be considered as sustained activity or remembrance of the integral of a prior input.

Our specific interest in network models is in such cognitive level phenomena - the ones that they could reproduce and the way they would explain them. The activity dynamics of these networks is studied extensively by \cite{DayanAbbot2005} who compiled results from numerous papers before. Feedforward networks with the above definition can calculate coordinate transformations needed in visually guided reaching tasks \cite{SalinasAbbott1995}. Recurrent networks have the ability to perform selective amplification, input integration and sustained activity for instance in maintaining memory trace of the horizontal eye position \cite{Seung1996}. Extending to the nonlinear recurrent network above (mostly to avoid negative firing rates which are introduced by (\ref{fr_rn_linear})) retains this selective amplification of input stimulus, as well as sustained activity, e.g. memory of a perceived input at the time of uniform current input. In addition, it is shown to perform selective dampening and therefore could describe both simple and complex cells responses in visual processing which are respectively selective and nonselective towards particular position of the stimulus \cite{SompolinskyShapley1997}. The selective amplification and dampening also produce winner-takes-all perception of multiple stimuli where one of two overlapping stimuli is ignored.

Stability analysis can be performed on (\ref{fr_rn_ein}) using a phase plane portrait and linearization around an equilibrium point. Study of the parameters can also reveal the possibility of a Hopf bifurcation \cite{DayanAbbot2005}. A deeper analysis elaborates on limitations of the additive inhibitory interactions and implements a multiplicative effect inhibition \cite{BattagliaTreves1998}. However, we will skip these here for the sake of more complex tasks and refer the interested reader to \cite{DayanAbbot2005}. A model like (\ref{fr_rn_ein}) is an important alternative when more realism is needed for the inhibitory interactions since it allows for different (empirically slower) time scale and (empirically larger) magnitude with respect to the excitatory interactions.

We are mostly interested in (\ref{fr_rn_frn2}), the study of which brings to conclusions also for the potential-based versions of the previous networks (\ref{fr_ffn}, \ref{fr_rn_lln}, \ref{fr_rn_ein}). In particular, by careful choice of the external input $\mathbf{U}$ and the measured output firing rates $\mathbf{v}$ one can partition the neurons into input units (units with nonzero external input), output units (units with measured firing rates), and hidden units (all the rest). These additive networks are also the most widely studied in literature in comparison to (\ref{fr_rn_frn1}) and any specific cases thereof, most typically with consideration of a constant or clamping input $\mathbf{U}(t) = \mathbf{U}$ (with some recent studies on oscillatory input as well \cite{ZhangWangLiu2014}). In this case we can reuse our old notation as
\begin{equation} \label{fr_rn_frn2n}
\tau_r \frac{d\mathbf{u}}{dt} = -\mathbf{u} + \mathbf{W} \cdot \mathbf{a(u)} + \mathbf{U}
\end{equation}
where $\mathbf{u}$ is the neuron potential, $\mathbf{W}$ is the recurrent weight matrix, $\mathbf{a(u)}$ is a general activation function for the firing rate such that $\mathbf{a}(\mathbf{u}) = (a(u_1),\cdots,a(u_n))^T$ and $\mathbf{U}$ is an external input to the neuron. The existence and uniqueness of solutions of (\ref{fr_rn_frn2}) can easily be established since the activation function $a(x)=(x-\alpha)^+$ and therefore the entire vector field is globally Lipschitz-continuous with Lipschitz constant of 1. Further results are well established for discontinuous activation functions of (\ref{fr_rn_frn2n}). The stability of (\ref{fr_rn_frn2}, \ref{fr_rn_frn2n}) plays a pivotal role in models of memory recall, pattern recognition and sequence prediction. A stable additive network is called an attractor network (also autoassociative memory, Hopfield network after \cite{Hopfield1982, Hopfield1984}, or Cohen-Grossberg-Hopfield network after \cite{CohenGrossberg1983}) due to the existence of an attractor for its dynamics. It is the ultimate goal for this section and offers a mathematical explanation for major cognitive phenomena like the ones just mentioned.

Under certain conditions, the most general recurrent network (\ref{fr_rn_frn2n}) is stable and can be shown to converge globally through the use of the energy (Lyapunov) functional
\begin{equation} \label{fr_ef}
E(\mathbf{u}) = - \frac{1}{2} \mathbf{a(u)}^T \mathbf{W} \mathbf{a(u)} - \mathbf{U}^T \mathbf{a(u)} + \sum_{i=1}^{N_u} \tilde{E}(u_i)
\end{equation}
The functional $E(\mathbf{u})$ was introduced by \cite{CohenGrossberg1983} with last term
\begin{equation*}
\tilde{E}(u_i) = \int_0^{u_i} s a'(s) ds
\end{equation*}
and by \cite{Hopfield1984} with last term
\begin{equation*}
\tilde{E}(u_i) = \int_0^{a(u_i)} a^{-1}(s) ds
\end{equation*}
where the second last term can be obtained from the first via change of variable $y=a(s)$ assuming that $a(0)=0$. This energy has negative semidefinite time derivative under the assumption that the matrix $\mathbf{W}$ is symmetric and that the activation function is monotone nondecreasing $a'(u) \geq 0$. This can be seen as
\begin{align*}
& \frac{dE(\mathbf{u})}{dt} = \sum_{i=1}^N \frac{dE}{du_i} \frac{du_i}{dt} = \sum_{i=1}^N [(- \sum_j^N w_{ij} a(u_j) - U_i + u_i) \frac{da(u_i)}{du_i}] \frac{du_i}{dt} =\\
& = - \sum_{i=1}^N a'(u_i) (\frac{du_i}{dt})^2 \leq 0 \text{ if } a'(u_i) \geq 0
\end{align*}
in the first case and as
\begin{align*}
& \frac{dE(\mathbf{u})}{dt} = \sum_{i=1}^N \frac{dE}{da(u_i)} \frac{da(u_i)}{du_i} \frac{du_i}{dt} =\\
& = \sum_{i=1}^N (- \sum_j^N w_{ij} a(u_j) - U_i + a^{-1}(a(u_i))) \frac{da(u_i)}{du_i} \frac{du_i}{dt} =\\
& = - \sum_{i=1}^N a'(u_i) (\frac{du_i}{dt})^2 \leq 0 \text{ if } a'(u_i) \geq 0
\end{align*}
in the second case. In order to show that the system is stable using LaSalle's invariance principle, we also need $E(\mathbf{u})$ to be continuous and bounded from below, so that we have convergence to an invariant set. This is the case for any saturating continuously differentiable activation function. In the case where the activation function is bounded from below but is at least locally Lipschitz continuous like $a(x)=(x-\alpha)^+$, we need to take the derivative of the energy along the trajectories as
\begin{align*}
D^+E(\mathbf{u_0}) = \lim_{h \rightarrow 0^+} \inf \frac{E(\mathbf{u_0} + h \mathbf{\dot{u}(u_0)}) - E(\mathbf{u_0})}{h}
\end{align*}
and replace the Riemann integral with a Radon integral in the definition of (\ref{fr_ef}). Finally, if the activation function and therefore the energy functional is only continuous, its derivative can be taken as
\begin{align*}
D^+E(\mathbf{u_0}) = \lim_{h \rightarrow 0^+} \inf \frac{E(\mathbf{u}(h, \mathbf{u_0})) - E(\mathbf{u_0})}{h}
\end{align*}
where $\mathbf{u_0}=\mathbf{u}(0)$ and $\mathbf{u}(t, \mathbf{u_0})$ is the trajectory with initial condition $\mathbf{u_0}$ \cite{Hale1969}. Clearly, the previous linear case with $a(x)=x$ is smooth but not bounded from below which is why it can play the role of an integrator.

We are paying large attention to the stability analysis of these network models because it represents a cognitive phenomenon on its own. When the activity dynamics of a network converge to a few fixed/equilibrium/steady states without external influence, we can compare this to the cognitive process of memory recall. Memory recall can then be considered as a part of the process of general pattern recognition, or even prediction if the recognition is of spatio-temporal patterns. More generally, attractor networks or networks whose dynamics settle towards an attractor, can be seen as a way to store and retrieve information based on the information's full or partial content as initial condition (content-based rather than address-based information retrieval). Any recalled purely spatial memories are equilibrium points within the phase space containing any possible network activity as a state. The retrieved spatio-temporal memories (including sequences of states) are general attractors within this phase space, like limit cycles for oscillations (chewing, walking, other rhythmic outputs), lines or other subspaces, or even fractals in the case of chaotic network behavior.

Let us return to the specific network models described here by considering spatial (static) memories or equilibrium points. If the initial activity is within the basin of attraction of a particular memory, the memory will be gradually recalled as the steady state for any further evolution or a process of pattern matching. A typical autoassociative network like (\ref{fr_rn_frn2}) has the additional properties that it satisfies conditions to make its convergence to an equilibrium point possible. Its evolution is studied from a given initial condition $\mathbf{v}(0)$. If this initial condition is within the basin of attraction of a memory state $\mathbf{v}_m$ for $m \in [1;N]$ with $N$ as the total number of memories stored in the network, it will converge and the memory recall will be successful. Recalling the wrong memory is another case of unsuccessful recall in addition to divergence of the state. Any further analysis of such associative memory is an optimization problem over the weights where one maximizes the capacity as the total number of recallable memories $N$ and the measure of the basin of attraction for each memory. Another problem that occurs in networks making use of the Lyapunov functional (\ref{fr_ef}) is the occurrence of spurious equilibrium points which are not memories.

A synthesis of a network like (\ref{fr_rn_frn1}) to store $N$ memories is derived by \cite{DayanAbbot2005} in the case of binary outputs. A memory $\mathbf{v}_m$ is an equilibrium point of (\ref{fr_rn_frn1}) and therefore must satisfy
\begin{equation*}
\mathbf{v}_m = (\mathbf{M} \cdot \mathbf{v}_m - \boldsymbol{\alpha})^+ = \mathbf{a}(\mathbf{M} \cdot \mathbf{v}_m)
\end{equation*}
where we have assumed no external input $\mathbf{U}=\mathbf{0}$.
We are interested in constructing the weights of $\mathbf{M}$ and maximizing the number $N$ of vectors $\mathbf{v}_m$ that can simultaneously satisfy this equation for an appropriate choice of $\mathbf{M}$. Considering binary values, we can require that each $\mathbf{v}_m$ has $\gamma N$ active units, i.e. units with value 1, and $(1-\gamma) N$ inactive units, i.e. units with value 0. We can call $\gamma$ the sparseness of each memory since higher $\gamma$ allows for storing more memories with less information for each. To guarantee the existence of equilibrium points, we need the Lyapunov function (\ref{fr_ef}) which in turn must be bounded from below (which can be achieved if the activation function $\mathbf{a}$ saturates) and requires $\mathbf{M}$ to be symmetric. Taking the weight matrix to be
\begin{equation} \label{syn1}
\mathbf{M} = \mathbf{K} - \frac{\mathbf{11}}{\gamma N} \text{ where } \mathbf{K} \cdot \mathbf{v}_m = \lambda \mathbf{v}_m
\end{equation}
i.e. $\mathbf{v}_m$ are the degenerate eigenvectors of a matrix $\mathbf{K}$, we get
\begin{align*}
& \mathbf{v}_m = \mathbf{a}(\mathbf{M} \cdot \mathbf{v}_m) = \mathbf{a}((\mathbf{K} - \frac{\mathbf{11^T}}{\gamma N}) \cdot \mathbf{v}_m) = \mathbf{a}(\lambda \mathbf{v}_m - \frac{\mathbf{1} (\mathbf{1}^T \cdot \mathbf{v}_m)}{\gamma N}) =\\
&= \mathbf{a}(\lambda \mathbf{v}_m - \frac{\mathbf{1} (\gamma N)}{\gamma N}) = \mathbf{a}(\lambda \mathbf{v}_m - \mathbf{1})
\end{align*}
Solving this for each component gives the conditions
\begin{equation*}
0 = a(- 1) \qquad 1 = a(\lambda - 1)
\end{equation*}
on the activation function which are satisfied for $a(x)=(x-\alpha)^+$ for $\lambda=2$. It remains to find $\mathbf{K}$ that satisfies at least approximately $ \mathbf{K} \approx \lambda \mathbf{v}_m$ but also such that $M$ is a symmetric matrix. This can be done under certain assumptions on randomness of the patterns as is done in \cite{DayanAbbot2005}.

More thorough survey on synthesis/design methods for attractor networks with predetermined equilibrium points, basins of attraction, convergence speed, etc. is presented in \cite{MichelFarrell1990}. Different approaches vary significantly in terms of computational requirements, capacity, stability, and restrictions. For instance, the early synthesis by \cite{Hopfield1982} with
\begin{equation} \label{syn2}
w_{ij} = \sum_m (2 u_m^i - 1) (2 u_m^j - 1)
\end{equation}
yields a capacity of $0.15N$. Since this construction of the weight matrix is done for (\ref{fr_rn_frn2n}), a memory is represented by $\mathbf{u}_m$ instead of $\mathbf{v}_m$. We can see that if $u_m^i=u_m^j=0$ or $u_m^i=u_m^j=1$ for some $m$ the weight $w_{ij}$ will increase by $1$ and that if they are different it will decrease by $1$. Further synthesis methods include the outer product method, the projection learning rule, and the eigenstructure method \cite{MichelFarrell1990}. All these methods go in the direction of creating and modifying attractors in the form of instant or gradual learning which is a subject of the next section.

\subsubsection{Connectivity - study of the synapse plasticity}

The synaptic strength or synaptic plasticity as well as the overall network connectivity are determined by activity-independent and activity-dependent mechanisms. It is a widely accepted belief that activity-independent mechanisms are responsible for the original targeting of axons, layers of enervation, etc. while the specialization and development of neuronal input selectivity and cortical mapping are also determined by activity-dependent mechanisms. This is suggested by experimental results from the hippocampus, neocortex, and cerebellum areas in the brain that link synaptic strength with neural activity \cite{DayanAbbot2005}. Most of the models encompassing some kind of adaptation (LTP and or LTD) are using some form of Hebbian learning. Examples of non-Hebbian synaptic plasticity are also possible and could involve learning based only on the presynaptic or postsynaptic firing. An example for nonsynaptic plasticity could be intrinsic plasticity of a neuron based solely on it activity.

A few important considerations valid for any adaptable models when it comes to plasticity are:
\begin{itemize}
	\item We should impose certain bounds on the synaptic strengths - a bound or a synaptic saturation should prevent uncontrolled growth of the synaptic strengths as well as the possibility of them crossing zero, i.e. becoming inhibitory from excitatory or vice versa.
	\item We should find a way to preserve diversity among the strengths through some synaptic competition since if all of them converge to zero or a maximum value the information contained in the connectivity is lost.
	\item The plasticity is realized by updating a synaptic strength according to a learning rule which can be a dynamical equation or a discrete update step.
\end{itemize}
Until the end of this section, we will review the most widely used learning rules providing dynamics for the weights, starting with the simplest but most impractical and ending with the most practical but highly specialized among them.

Expanding on the rate-based networks (\ref{fr_ffn}) build earlier, we again consider the weights $w_{ij}=w_{ij}(t)$ as synaptic strengths however this time not as constant coefficients. Since $w_{ij}$ now depend on the activity, i.e. the rates $v_i$ and $u_j$, we need another equation for its dynamics. However, since generally the learning/connectivity time scale $\tau_w$ is a lot slower than the perception/activity time scale $\tau_r$, we can use QSSA and assume $v_i = \mathbf{w_i} \cdot \mathbf{u}$ instantly reaches its steady state in the first equation so that the simplest network model capable of adaptation is
\begin{equation} \label{afr_ffn}
\tau_w \frac{d\mathbf{w_i}}{dt} = v_i \mathbf{u} = (\mathbf{w_i} \cdot \mathbf{u}) \mathbf{u}
\end{equation}
where the equations are indexed by $i$, i.e. we have preferred index notation rather than tensor notation. This network uses a basic Hebbian learning rule
\begin{equation} \label{hebb_basic}
\tau_w \frac{d\mathbf{w_i}}{dt} = F(v_i, \mathbf{u}) = v_i \mathbf{u}
\end{equation}
since it can be immediately verified that simultaneous activity on the inputs $\mathbf{u}$ and the output $v_i$ has positive effect on the derivative of $\mathbf{w_i}$. We can see now that we can produce a different network model (same QSSA but different plasticity dynamics) for each learning rule we select.

Two prominent learning rules of the Hebbian type are the correlation and covariance rules which we can obtain if we use respectively
\begin{equation} \label{hebb_corr}
F(v_i, \mathbf{u}) = \langle v_i \mathbf{u} \rangle \Rightarrow \tau_w \frac{d\mathbf{w_i}}{dt} = \mathbf{Q} \cdot \mathbf{w_i}
\end{equation}
\begin{equation} \label{hebb_covar}
F(v_i, \mathbf{u}) = (v_i - \theta_v) \mathbf{u} \text{ or } F(v_i, \mathbf{u}) = v_i (\mathbf{u} - \mathbf{\theta_u}) \Rightarrow \tau_w \frac{d\mathbf{w_i}}{dt} = \mathbf{C} \cdot \mathbf{w_i}
\end{equation}
with the input correlation matrix $\mathbf{Q} = \langle \mathbf{u} \mathbf{u} \rangle$ and the input covariance matrix $\mathbf{C} = \langle (\mathbf{u} - \langle \mathbf{u} \rangle) (\mathbf{u} - \langle \mathbf{u} \rangle) \rangle = \langle \mathbf{u} \mathbf{u} \rangle - \langle \mathbf{u} \rangle ^2 = \langle (\mathbf{u} - \langle \mathbf{u} \rangle) \mathbf{u} \rangle$.
Some interesting observations to make here are that both covariance rules (\ref{hebb_covar}) contain thresholds (output threshold $\theta_v$ or input thresholds $\theta_u$) that separate the LTD from LTP activity (negative to positive derivative, i.e. a nullcline for $\mathbf{w_i}$). The first to the two equations modifies the synapse strengths only if they have nonzero presynaptic activities while the second only if the postsynaptic activity is not zero.

All learning rules until here are unstable and therefore not very practical. An improved and stable rule is the Bienenstock-Cooper-Monroe (BCM) rule
\begin{align} \label{hebb_bcm}
\begin{dcases}
& \tau_w \frac{d\mathbf{w_i}}{dt} = F(v_i, \mathbf{u}) = v_i \mathbf{u} (v_i - \theta_v)\\
& \tau_{\theta} \frac{d\theta_v}{dt} = v_i^2 - \theta_v
\end{dcases}
\end{align}
that also supports LTD in addition to LTP through the threshold $\theta_v$. Allowing $\theta_v$ to grow more rapidly than $v_i$ with $\tau_{\theta} < \tau_w$ is the way to achieve stability here. One could also use rules that make use of normalization of the weights (thus also taking care of competition among the synapses and limited resource within a neuron) but this makes use of global information outside of the scope of the individual synapse and is no longer biologically meaningful. A stable, competitive, and local rule is the Oja rule
\begin{equation} \label{hebb_oja}
\tau_w \frac{d\mathbf{w_i}}{dt} = F(v_i, \mathbf{u}) = v_i \mathbf{u} - \alpha v_i^2 \mathbf{w_i}
\end{equation}
where $\alpha>0$ is a constant. It is local because $w_{ij}$ only depends on $v_i$, $u_j$, and $w_{ij}$ while in a general normalization we would also need $w_{ik}$ and $u_k$ for $k \neq j$ to complete the computation. It is stable and competitive because if we multiply the equation (\ref{hebb_oja}) by $\mathbf{w_i}$ and use $v_i = \mathbf{w_i} \cdot \mathbf{u}$, the resulting equation
\begin{equation*}
\tau_w \frac{d|\mathbf{w_i}|^2}{dt} = 2 v_i \mathbf{w_i} \cdot \mathbf{u} - 2 \alpha v_i^2 |\mathbf{w_i}|^2 = 2 v_i^2 (1 - \alpha |\mathbf{w_i}|^2)
\end{equation*}
has $|\mathbf{w_i}|^2=\frac{1}{\alpha}$ as a steady state which is finite (stability) constant (competition - to increase one weight and preserve the constant we must decrease another weight).

A final and very important learning rule that we will look at is the timing based rule
\begin{equation} \label{hebb_timing}
\tau_w \frac{d\mathbf{w_i}}{dt} = F(v_i, \mathbf{u}) = \int_0^{+\infty} H(\tau) v_i(t) \mathbf{u}(t-\tau) + H(-\tau) v_i(t-\tau) \mathbf{u}(t) d\tau
\end{equation}
This rule stems from an attempt to relate rate-based Hebbian learning back to spike-based empirical results where a significant synaptic potentiation was observed if the presynaptic spike is followed by a postsynaptic spike within a sufficiently short window\cite{GerstnerKistler2002}. If the postsynaptic spike is followed by the presynaptic spike, the synaptic strengths are rather depressed and if the window is too large no change in the synaptic plasticity takes place. The rule is an approximation where $\tau$ is the difference between the pre- and postsynaptic firing rate evaluation times. Here $H(\tau)$ describes the weight change rate as a function of the difference $\tau$ and a good candidate for it is anything of the form resembling a bounded version of $\frac{1}{x}$, $x \in \mathbb{R}$. If this function always has the sign of $\tau$, the first term in (\ref{hebb_timing}) represents LTP and the second LTD. It is stable if we add proper bounds and competitive if we consider spiking network models. Most importantly, it depends on the order of the pre- and postsynaptic activity.

Although the above models are only feedforward networks that have been enhanced with plasticity, they are already capable of a number of cognitive level learning phenomena like the development of ocular dominance in cells of the primary visual cortex, orientation selectivity, and trace learning \cite{DayanAbbot2005}. On the other hand, adding lateral connections like (\ref{fr_rn_lln}) can reproduce the appearance of ocular dominance strips on the output layer and other additional phenomena.

There could be different ways of adding lateral connections, namely one could add fixed lateral connections keeping the feedforward connections plastic, fix the feedforward connections and add plastic lateral connections or have plasticity for all connections. In many cases using simply a feedforward network might lead to redundancy in the output layer where each neuron will obtain the same weights and input selectivity for instance using (\ref{hebb_oja}). One way to deal with this is by adding plastic lateral connections with anti-Hebbian learning rule (depressing correlated firing on the second layer to reduce redundancy in all firing there, also observed in nature) in addition to the feedforward Oja rule
\begin{align} \label{afr_rn}
\begin{dcases}
& \tau_w \frac{dw_{ij}}{dt} = v_i u_j - \alpha v_i^2 w_{ij}\\
& \tau_M \frac{dm_{ij}}{dt} = - v_i v_j + \beta m_{ij}
\end{dcases}
\end{align}
The second equation is the anti-Hebbian basic rule, i.e. the learning rule (\ref{hebb_basic}) with a reversed sign and with an additional $\beta m_{ij}$ term to make sure all weights don't converge to zero (the now reversed instability of Hebbian learning). This model network is stable yet not trivial across neurons with a suitable choice of a $\beta>0$ constant.

If the synaptic modification is modelled as a discrete rather than continuous process which would be preferable if the input consists of temporal sequence of patterns, we could use a discrete learning/update rule
\begin{equation} \label{hebb_discrete}
\mathbf{w_i} \rightarrow \mathbf{w_i} + \epsilon F(v_i, \mathbf{u})
\end{equation}
To obtain the discrete version of any of the continuous learning rules above, one only needs to use the correct $F(v_i, \mathbf{u})$.
A this point it's very easy to distinguish between unsupervised and supervised learning in terms of the models' formalism. In the case of unsupervised learning which is what we have implicitly considered above, $v_i$ will be generated by the network through $v_i = \mathbf{w_i} \cdot \mathbf{u}$ which can be replaced in $F(v_i, \mathbf{u})$ to obtain an unsupervised learning rule. In the case of supervised learning, $v_i$ is already given as a desired output value $v_i(t)=\bar{v_i}$ which is then replaced in $F(v_i(t), \mathbf{u}(t)) = F(\bar{v_i}, \mathbf{u}(t))$.

Two important examples where supervised Hebbian learning rules are not performing well are binary classification through a perceptron neuron and functional approximation through a firing rate neuron. If we consider a perceptron model like (\ref{dm1}) as a binary classifier, it is possible for the perceptron to classify inputs perfectly if and only if there exists a hyperplane separating the input space into half spaces of inputs with each one of the binary values. This condition is known as the condition of linear separability. The case of functional approximation is possible when the firing rate of a neuron is given by a function of a stimulus parameter $v(s) = \mathbf{w} \cdot \mathbf{f}(s) = \sum_i w_i f_i(s)$. The approximated function is a linear combination of the functions of the stimulus for the firing rates if the input neurons $u_i$. In both examples, supervised Hebbian learning rules will not work well because they do not depend on the actual performance of the network. For this reason we can introduce error-correcting learning rules where we would start with initial guess for the weights, compare the actual neuron output $v(u^k)$ in response to the input $u^k$ with the desired output $\bar{v}^k$ and adapt the weights to reduce the error between the two. Here we have dropped the $i$ index since we consider one neuron, and have added the $k$ index since we consider a discrete sequence of inputs in time. The binary classification can be augmented with the perceptron learning rule
\begin{equation} \label{hebb_perceptron}
\mathbf{w} \rightarrow \mathbf{w} + \frac{\epsilon_w}{2} (v^m - v(\mathbf{u}^m)) \mathbf{u}^m \text{ and } \theta \rightarrow \theta - \frac{\epsilon_w}{2} (v^m - v(\mathbf{u}^m))
\end{equation}
and the functional approximation with the delta learning rule
\begin{equation} \label{hebb_delta}
\mathbf{w} \rightarrow \mathbf{w} - \epsilon_w \nabla_w E
\end{equation}
where $\theta$ is the threshold from (\ref{dm1}) and $E(s)$ is the square error function which makes the delta rule a an update step in a gradient descent.

A step beyond the rate-based learning rules discussed above and into plasticity for spiking models is taken by \cite{GerstnerKistler2002} and is recommended for more biological realism.

\section{Proposed model}

\subsection{Formulation}

We will now construct and motivate a family of rate-based models of increasing complexity. These have significant morphological difference with the regular neural network models in the previous section but can, under suitable conditions, be reduced to or rather interpreted as such networks therefore attaining their previous results. The emphasis of course is on new and more general cognitive phenomena that they should describe.

\subsubsection{Construction of a family of models}

As we have mentioned before, it is essential to pick the right features from the more complex and biologically realistic models while still abstracting from everything that could be replaced with computationally effective mechanism. In the model we are going to propose in this section, we do exactly this by extracting further possible features from the synaptic integration as well as the distal dendritic tree.

\begin{figure}[!htbp]
\centering
\includegraphics[width=0.75\textwidth]{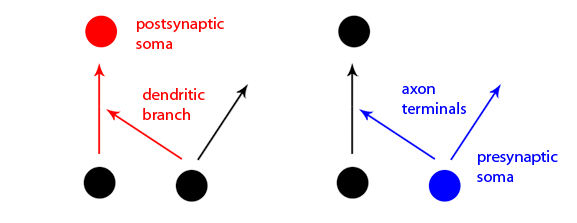}
\caption{Comparison between a simple circuit with metaconnections and an actual neuron's morphology.}
\label{metaconnections}
\end{figure}

Moving in the direction of multi-compartment models, we are interested in having a separate dynamic for each branch of the dendritic tree. A separate activation/threshold function $a(\cdot)$ should capture the properties of propagating an action potential towards the soma while the local PSP $e(\cdot)$ should be individually determined from the presynaptic input. To reduce computational and notational complexity, we will restrict ourselves only to the case of single branching, i.e. a dendrite can only branch once where it, together with its subbranches, each have their own dynamic. We shall exclude having an entire cable PDE as such dynamic and instead construct our own abstractions to capture some minimal properties we need. We will consider the neuron somas as simple units while axons and unbranched dendrites will be allowed to be confused together as connections. Finally, subbranches of dendrites but also some axon terminals will be called metaconnections since they can be loosely defined as connections to other connections.

We start constructing a new model from (\ref{fr_ffn}) by first defining a differential equation for each unit, connection, and metaconnection, then adding more specialized dynamics to each equation and finally defining an input-output layer feedforward network. Also as before, we will then gradually increase the complexity of the network topology moving to recurrent and fully recurrent networks. The basic feedforward model with dynamics similar to (\ref{fr_ffn}), $N$ inputs, and $M$ outputs reads
\begin{align} \label{mm1}
\begin{dcases}
& \tau_r \frac{dv_i}{dt} = - v_i + (\sum_{j=1}^N w_j^i c_j^i - \alpha)^+\\
& \tau_r \frac{dc_j^i}{dt} = - c_j^i + (\sum_{\substack{k = 1 \\ k \neq j}}^N w_k^{ji} c_k^{ji} + u_j - \alpha)^+\\
& \tau_r \frac{dc_k^{ji}}{dt} = - c_k^{ji} + (u_k - \alpha)^+,\ k \neq j
\end{dcases}
\end{align}
where $i \in [1;M]$, $j \in [1;N]$, and $k \in [1;N]$. The notation is similar to the one used for the basic rate-based networks before, namely $v_i$ for the $i$-th output and $u_j$ for $j$-th input unit and the $\mathbf{W}$ tensor represents constant weights. However, there are the dynamics for the pathways to the output units denoted by $c_j^i$ for the regular connections from input unit $j$ to output unit $i$ and by $c_k^{ji}$ for the metaconnections from input unit $k$ to the connection from $j$ to $i$.

Changing the dynamics from firing rate outputs to internal potentials $u^i=\sum_{j=1}^N w_j^i c_j^i$, $u_j^i=\sum_{k = 1 k \neq j}^N w_k^{ji} c_k^{ji} + u_j$, $u_k^{ji}=u_k$, and $U_j$ as internal potential of the input unit with its firing output rate determined by $\tau_r \dot{u_j} = - u_j + (U_j-\alpha)^+$, we obtain
\begin{align*}
& \tau_r \frac{du^i}{dt} = \sum_{j=1}^N w_j^i \tau_r \frac{dc_j^i}{dt} = - \sum_{j=1}^N w_j^i c_j^i + \sum_{j=1}^N w_j^i (\sum_{\substack{k = 1 \\ k \neq j}}^N w_k^{ji} c_k^{ji} + u_j - \alpha)^+=\\
& = - u^i + \sum_{j=1}^N w_j^i (u_j^i-\alpha)^+\\
& \tau_r \frac{du_j^i}{dt} = \sum_{\substack{k = 1 \\ k \neq j}}^N w_k^{ji} \tau_r \frac{dc_k^{ji}}{dt} + \tau_r \frac{du_j}{dt} = \sum_{\substack{k = 1 \\ k \neq j}}^N w_k^{ji} [- c_k^{ji} + (u_k - \alpha)^+] + \tau_r \frac{du_j}{dt} =\\
& = - \sum_{\substack{k = 1 \\ k \neq j}}^N w_k^{ji} c_k^{ji} - u_j + u_j + \sum_{\substack{k = 1 \\ k \neq j}}^N w_k^{ji} (u_k^{ji} - \alpha)^+ - u_j + (U_j-\alpha)^+ =\\
\end{align*}
\begin{align*}
& = - u_j^i + \sum_{\substack{k = 1 \\ k \neq j}}^N w_k^{ji} (u_k^{ji}-\alpha)^+ + (U_j-\alpha)^+\\
& \tau_r \frac{du_k^{ji}}{dt} = \tau_r \frac{du_k}{dt} = - u_k + (U_k-\alpha)^+ = - u_k^{ji} + (U_k-\alpha)^+
\end{align*}
While $v_i$ stands for the output of the $i$-th output neuron in the basic rate-based models by \cite{DayanAbbot2005} and therefore also in (\ref{mm1}), i.e. the quantity obtained after applying the activation function, $u^i$ is rather the total accumulated potential of the $i$-th output neuron. Instead of taking the activation function of the linear summation of inputs, we now take linear summation of activated inputs similarly to the transition from (\ref{fr_rn_frn1}) to (\ref{fr_rn_frn2}). We will now abuse our notation and use $u_k$ to imply internal potential of an input unit. The resulting model reads
\begin{align} \label{mm2}
\begin{dcases}
& \tau_r \frac{du^i}{dt} = - u^i + \sum_{j=1}^N w_j^i (u_j^i-\alpha)^+\\
& \tau_r \frac{du_j^i}{dt} = - u_j^i + \sum_{\substack{k = 1 \\ k \neq j}}^N w_k^{ji} (u_k^{ji}-\alpha)^+ + (u_j-\alpha)^+\\
& \tau_r \frac{du_k^{ji}}{dt} = - u_k^{ji} + (u_k-\alpha)^+,\ k \neq j
\end{dcases}
\end{align}
Notice that the internal potentials depend on the weighted sums of the instant output firing rates in the above model. We could also replace $U_k:=(u_k-\alpha)^+$ as external inputs to the system by a second abuse of notation this time reusing $U_k$. Of course, this replacement is only valid for feedforward and lateral networks since if there are connections to the input units, their potentials $u_i$ become unknowns for the system.

The first major extension besides the more detailed network structure is the addition of a "context" term $e$ which replaces the external input term $a(u_k)=(u_k-\alpha)^+$ of (\ref{mm2}) with a function defined by
\begin{equation} \label{e_basic}
e(c,u,C_i) =
	\begin{dcases}
		\frac{c a(u)}{h(C_i)} \qquad \quad \text{if } \quad h(C_i) \neq 0 \\
		\frac{a(u)}{MN} \qquad \quad \text{ otherwise}
	\end{dcases}
\end{equation}
where $C_i = \{c_i^{0,1},\dots,c_i^{N,M}\}$ and
\begin{equation} \label{h_basic}
h(C_i) = \sum\limits_{l=1}^M \sum\limits_{\substack{m = 0 \\ m \neq i}}^N c_i^{ml}
\end{equation}
Similarly to the activation function $a(\cdot)$, $e(\cdot)$ can be called distribution function since it is used to determine the potential distribution among outputs to different units. For the description of this distribution function we have used the convention $c_i^l=c_i^{0,l}$. In essence, the distribution function $e(\cdot)$ determines the potential at a connection by splitting the total output of its presynaptic unit according to already accumulated potential through other metaconnections. If no such potential is present, it will equidistribute the potential which is analogical to broadcasting of the presynaptic unit. Using the activation function $a(u)=(u-\alpha)^+$, we get the form
\begin{align} \label{mm3}
\begin{dcases}
& \tau_r \frac{du^i}{dt} = - u^i + \sum_{j=1}^N w_j^i a(u_j^i)\\
& \tau_r \frac{du_j^i}{dt} = - u_j^i + \sum_{\substack{k = 1 \\ k \neq j}}^N w_k^{ji} a(u_k^{ji}) + e(u_j^i,u_j,u_j^{0,1},\dots,u_j^{N,M})\\
& \tau_r \frac{du_k^{ji}}{dt} = - u_k^{ji} + e(u_k^{ji},u_k,u_k^{0,1},\dots,u_k^{N,M}),\ k \neq j
\end{dcases}
\end{align}

The second major extension that we will consider is the addition of potential storage if the threshold condition is not met. We can immediately reformulate (\ref{mm3}) as
\begin{align} \label{mm4}
\begin{dcases}
& \tau_r \frac{du^i}{dt} = - a(u^i) + \sum_{j=1}^N w_j^i a(u_j^i)\\
& \tau_r \frac{du_j^i}{dt} = - a(u_j^i) + \sum_{\substack{k = 1 \\ k \neq j}}^N w_k^{ji} a(u_k^{ji}) + e(u_j^i,u_j,u_j^{0,1},\dots,u_j^{N,M})\\
& \tau_r \frac{du_k^{ji}}{dt} = - a(u_k^{ji}) + e(u_k^{ji},u_k,u_k^{0,1},\dots,u_k^{N,M}),\ k \neq j
\end{dcases}
\end{align}
The decay of $u^i$ is now being subjected to the threshold condition. If there is no activation $a(u)=(u-\alpha)^+=0$, the potential is stored while it will simply disappear in the previous cases. This will lead to different short term behavior which is also affected by activity from previous times. For a system with potential storage, it makes more sense to use the discontinuous activation/threshold function
\begin{equation} \label{a_basic}
a(u) =
\left\{
	\begin{array}{ll}
		u \text{ if } u > \alpha\\
		0 \text{ otherwise}
	\end{array}
\right.
\end{equation}
which reflects any previously stored potential in its output. Thus, when referring to the form (\ref{mm4}), we will assume the discontinuous activation is used.

Finally, we could block emitting based on highway priority criterion where a (meta)connection would emit only if its input unit emits, i.e. use an extended activation function for all the connections and metaconnections
\begin{equation} \label{a_ext}
\bar{a}(c,u) =
\left\{
	\begin{array}{ll}
		c \text{ if } c > \alpha \text{ and } u > \alpha\\
		0 \text{ otherwise}
	\end{array}
\right.
\end{equation}
The final form of the feedforward network using such dynamics becomes
\begin{align} \label{mm5}
\begin{dcases}
& \tau_r \frac{du^i}{dt} = - a(u^i) + \sum_{j=1}^N w_j^i \bar{a}(u_j^i,u_j)\\
& \tau_r \frac{du_j^i}{dt} = - \bar{a}(u_j^i,u_j) + \sum_{\substack{k = 1 \\ k \neq j}}^N w_k^{ji} \bar{a}(u_k^{ji},u_k) + e(u_j^i,u_j,u_j^{0,1},\dots,u_j^{N,M})\\
& \tau_r \frac{du_k^{ji}}{dt} = - \bar{a}(u_k^{ji},u_k) + e(u_k^{ji},u_k,u_k^{0,1},\dots,u_k^{N,M}),\ k \neq j
\end{dcases}
\end{align}
The activation function $\bar{a}(\cdot)$ will propagate the potential further if the more major dendrite is sufficiently excited and will play the standard role of a threshold function in case this is true.

Before fully motivating the choices of these extensions, we will complete the model definitions with the other network variants similarly to the way we did in the extension of neuron to networks models. In order to make it possible to consider recurrent networks, we need to elaborate more on both the coupling among equations as well as the indexing we employ to describe the different scopes of connectivity. From our construction above, metaconnections can depend only on units, connections only on metaconnections and units, and units only on connections. Units then can receive only feedforward, lateral and feedback connections. Feedforward metaconnections can be defined as connections to feedforward connections from the same layer or to feedback connections from the next layer and the same applies for the feedback and lateral (to lateral inward and outward) metaconnections. Hence, there could be a much larger variety of recurrent network types depending on acceptable connection types. In order to derive some recurrent network models similar to the regular networks before, we must make some simplifying assumptions leading to reduced coupling:
\begin{itemize}
	\item We will consider only metaconnections to feedforward connections for simplicity of indexing and phenomenological study.
	\item Even though we will state a complete definition in terms of connection types, we will later on exclude lateral and/or feedback connections again with the goal of reducing the complexity of our study.
\end{itemize}
Notice that while in (connection only) regular networks removing feedback connections would completely decouple any layer vector equations, here this is no longer the case and the evolution of the network would be different if it comprises two or more layers.

\begin{figure}[!htbp]
\centering
\includegraphics[width=0.75\textwidth]{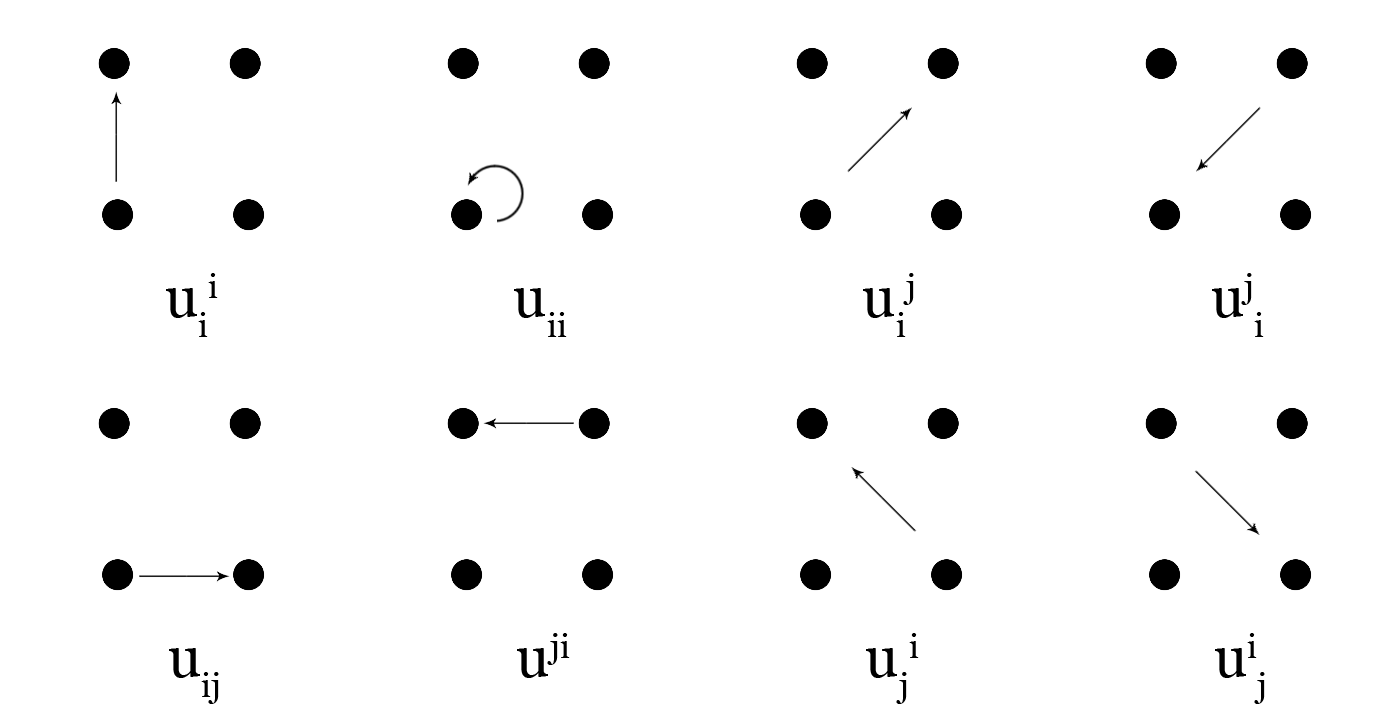}
\caption{Indexing cases for a connection in a 2x2 network.}
\label{indexing}
\end{figure}

In light of this, both the horizontal and vertical order of the index will be used to convey information about the indexed connection as shown in figure \ref{indexing}. The same indexing approach is immediately extendable to metaconnection indexing due to our first simplifying assumption.

Using this indexing convention and the reduced coupling, we can define a minimalistic layered network with feedforward and lateral connectivity as
\begin{align} \label{mm11}
\begin{dcases}
& \tau_r \frac{du^i}{dt} = - a(u^i) + \sum_{j=1}^N w_{j\ }^{\ i} \bar{a}(u_{j\ }^{\ i},u_j) + \sum_{l=1}^M w^{li} \bar{a}(u^{li},u^l)\\
& \tau_r \frac{du_{j\ }^{\ i}}{dt} = - \bar{a}(u_{j\ }^{\ i}, u_j) + \sum_{\substack{k = 1 \\ k \neq j}}^N w_{k\ }^{\ ji} \bar{a}(u_{k\ }^{\ ji}, u_k) + e(u_{j\ }^{\ i}, u_j, u_{j\ }^{\ 0,1}, \dots, u_{j\ }^{\ NM})\\
& \tau_r \frac{du^{li}}{dt} = - \bar{a}(u^{li}, u^l) + \bar{e}(u^{li}, u^l, u^{l1}, \dots, u^{lM})\\
& \tau_r \frac{du_{k\ }^{\ ji}}{dt} = - \bar{a}(u_{k\ }^{\ ji}, u_k) + e(u_{k\ }^{\ ji}, u_k, u_{k\ }^{\ 0,1}, \dots, u_{k\ }^{\ NM}),\ k \neq j\\
\end{dcases}
\end{align}
where $i, l \in [1;M]$ and $j, k \in [1;N]$. An example of an excitatory-inhibitory network model which will also be used later on to obtain a feature detection result is one where we have two groups of neurons where each neuron excites its neighbors within its group and inhibits its neighbors within the other group
\begin{align} \label{mm12}
\begin{dcases}
& \tau_r \frac{du^i}{dt} = - a(u^i) + \sum_{j=1}^N \bar{a}(u_{j\ }^{\ i}, u_j)\\
& \tau_r \frac{du_{j\ }^{\ i}}{dt} = - \bar{a}(u_{j\ }^{\ i}, u_j) + \sum_{\substack{k = 1 \\ k \neq j}}^N \bar{a}(u_{k\ }^{\ ji}, u_k) - \sum_{\substack{k = 1 \\ k \neq j}}^N \bar{a}(v_{k\ }^{\ ji}, v_k) + e(u_{j\ }^{\ i}, \dots)\\
& \tau_r \frac{du_{k\ }^{\ ji}}{dt} = - \bar{a}(u_{k\ }^{\ ji}, u_k) + e(u_{k\ }^{\ ji}, u_k, u_{k\ }^{\ 0,1}, \dots, u_{k\ }^{\ NM}),\ k \neq j
\end{dcases}
\end{align}
with three more equations for the $v$ group respectively and all weight constants set to $1$. Once again, we have to note that in this model we have ignored Dale's law for computational simplicity but we could consider it by separating the two groups into more groups of neurons some of which will have only inhibitory and others only excitatory outputs. We will look into more details for this model in section \ref{feature_detection}.

A reduced fully recurrent network using the above indexing
\small
\begin{align} \label{mm13}
\begin{dcases}
& \tau_r \frac{du^i}{dt} = - a(u^i) + \sum_{j=1}^N w_{j\ }^{\ i} \bar{a}(u_{j\ }^{\ i},u_j) + \sum_{l=1}^M w^{li} \bar{a}(u^{li},u^l)\\
& \tau_r \frac{du_{j\ }^{\ i}}{dt} = - \bar{a}(u_{j\ }^{\ i}, u_j) + \sum_{\substack{k=1 \\ k \neq j}}^N w_{k\ }^{\ ji} \bar{a}(u_{k\ }^{\ ji}, u_k) + \sum_{l=1}^M w^{l\ }_{\ ji} \bar{a}(u^{l\ }_{\ ji},u^l) + e(u_{j\ }^{\ i}, u_j, \dots)\\
& \tau_r \frac{du^{li}}{dt} = - \bar{a}(u^{li}, u^l) + \bar{e}(u^{li}, u^l, u^{l1}, \dots, u^{lM}, u^{l\ }_{\ 0,1}, \dots, u^{l\ }_{\ NM})\\
& \tau_r \frac{du^{i\ }_{\ j}}{dt} = - \bar{a}(u^{i\ }_{\ j}, u^i) + \bar{e}(u^{i\ }_{\ j}, u^i, u^{i1}, \dots, u^{iM}, u^{i\ }_{\ 0,1}, \dots, u^{i\ }_{\ NM})\\
& \tau_r \frac{du_{k\ }^{\ ji}}{dt} = - \bar{a}(u_{k\ }^{\ ji}, u_k) + e(u_{k\ }^{\ ji}, u_k, u_{k\ }^{\ 0,1}, \dots, u_{k\ }^{\ NM}),\ k \neq j\\
& \tau_r \frac{du^{l\ }_{\ ji}}{dt} = - \bar{a}(u^{l\ }_{\ ji}, u^l) + \bar{e}(u^{l\ }_{\ ji}, u^l, u^{l1}, \dots, u^{lM}, u^{l\ }_{\ 0,1}, \dots, u^{l\ }_{\ NM})\\
& \tau_r \frac{du_j}{dt} = - a(u_j) + \sum_{i=1}^M w^{i\ }_{\ j} \bar{a}(u^{i\ }_{\ j}, u^i) + U_j\\
\end{dcases}
\end{align}
\normalsize
can be useful for studying network motifs still distinguishing among the layers. The fact that we have excluded connections of the kind $u_{j\ }^{\ ji}$ and haven't excluded the ones of the kind $u^{j\ }_{\ ji}$ is an engineering-motivated choice which is not of great importance and can be ignored. The indexing of feedforward and feedback metaconnections $u_{k\ }^{\ ji}$ and $u^{l\ }_{\ ji}$ is a collapsed version of the indexing we would otherwise need for a metaconnection because we only allow $ji$ to represent a feedforward connection $u_{j\ }^{\ i}$. The output layer unit $u^i$ can be replaced with $v_i$ as before in which case we can use multilayer (more than two layer) indexing similar to the one above but with $v^{ji}$ instead of $u^{ji}$ to contain information about the starting reference unit. With such extended indexing, $u^{i\ }_{\ j}$ would refer to the feedback connection from $u_i$ in the $\mathbf{u}$ layer while $v^{i\ }_{\ j}$ to the feedback connection from $v_i$ in the $\mathbf{v}$ layer.

The functions $a(\cdot)$ and $e(\cdot)$ are similar for each type of connection with the only difference that we also add extended version of the distribution function
\begin{align}  \label{e_ext}
\bar{e}(c, u, C_i) & =
\begin{dcases}
	\frac{c a(u)}{\bar{h}(C_i)} \text{ if } \bar{h}(C_i) \neq 0 \\[0.5em]
	\frac{a(u)}{MN} \text{ otherwise}
\end{dcases} \quad \bar{h}(C_i) & = \sum\limits_{m=1}^M [u^{im} + \sum\limits_{\substack{n = 0 \\ n \neq i}}^N (u_{i\ }^{\ nm} + u^{i\ }_{\ nm})]
\end{align}
with $C_i = \{u^{i1}, \dots, u^{iM}, u_{i\ }^{\ 0,1}, \dots, u_{i\ }^{\ NM}, u^{i\ }_{\ 0,1}, \dots, u^{i\ }_{\ NM}\}$ where for the network (\ref{mm11}) we more specifically have
\begin{equation*}
C_l = \{u^{l1}, \dots, u^{lM}\} \qquad h(C_l) = \sum_{m=1}^M u^{lm}
\end{equation*}
and for the network (\ref{mm13}) we more specifically have
\begin{equation*}
C_l = \{u^{l1}, \dots, u^{lM}, u^{l\ }_{\ 0,1}, \dots, u^{l\ }_{\ NM}\} \qquad h(C_l) = \sum_{m=1}^M [u^{lm} + \sum\limits_{\substack{n = 0 \\ n \neq j}}^N u^{l\ }_{\ nm}]
\end{equation*}

Finally, a fully recurrent collapsed version can be constructed from (\ref{mm5}) if we arrange all the unknowns in a tensor using the conventions $u_k^{ji}=u_{kji}$, $u_j^i=u_{j0i}$, $u^i=u_{00i}$, $u_i=u_{i00}$ and symmetrize the equations
\begin{align*}
\begin{dcases}
& \tau_r \frac{du_{00i}}{dt} = - a(u_{00i}) + \sum_{j=1}^N w_{j0i} \bar{a}(u_{j0i},u_{j00}), i \neq 0\\
& \tau_r \frac{du_{j0i}}{dt} = - \bar{a}(u_{j0i},u_{j00}) + \sum_{\substack{k = 1 \\ k \neq j}}^N w_{kji} \bar{a}(u_{kji},u_{k00}) + e(u_{j0i},u_{j00},\dots), j,i \neq 0\\
& \tau_r \frac{du_{kji}}{dt} = - \bar{a}(u_{kji},u_{k00}) + e(u_{kji},u_{k00},u_{k01},\dots,u_{kNM}), k,j,i \neq 0, k \neq j\\[10pt]
& u_{000} = 1,\ u_{kj0} = 0, \forall j \neq 0,\ u_{0ji} = 0, \forall j \neq 0, \text{ and } u_{kki} = 0, \forall k, i \neq 0
\end{dcases}
\end{align*}
now for $i \in [0;M]$, $j \in [0;N]$, and $k \in [0;N]$. Taking $a(u_{00i})=\bar{a}(u_{00i}, 1)$, $w_{jji}=0$, and renumbering all the units so that we don't distinguish among input, hidden, and output units (therefore consider connections, metaconnections, and dynamics also to the input units $u_i$), we can see that
\begin{gather*}
\begin{split}
& \tau_r \frac{du_{00i}}{dt} = - \bar{a}(u_{00i}, u_{000}) + 1 \sum_{j=1}^N w_{j0i} \bar{a}(u_{j0i},u_{00j}) + 0 \bar{e}(u_{00i}, u_{000}, u_{001}, \cdots, u_{0NN})\\
& \tau_r \frac{du_{j0i}}{dt} = - \bar{a}(u_{j0i},u_{00j}) + 1 \sum_{k=1}^N w_{kji} \bar{a}(u_{kji},u_{00k}) + 1 \bar{e}(u_{j0i},u_{00j},u_{j01},\dots,u_{jNN})\\
& \tau_r \frac{du_{kji}}{dt} = - \bar{a}(u_{kji},u_{00k}) + 0 \sum_{l=1}^N w_{lki} \bar{a}(u_{lki},u_{00l}) + 1 \bar{e}(u_{kji},u_{00k},u_{k01},\dots,u_{kNN})\\
& u_{000} = 1,\ u_{kj0} = 0, \forall j \neq 0,\ u_{0ji} = 0, \forall j \neq 0, \text{ and } u_{kki} = 0, \forall k, i \neq 0
\end{split}
\end{gather*}
which can be collapsed to
\begin{equation} \label{mm100}
\begin{split}
\tau_r \frac{du_{kji}}{dt} = - \bar{a}(u_{kji},u_{00k}) + c_{kji} \sum_{l=1}^N w_{lki} \bar{a}(u_{lki},u_{l00}) +\\
+ b_{kji} \bar{e}(u_{kji},u_{00k},u_{k01},\dots,u_{kNN}) + U_{kji}
\end{split}
\end{equation}
with the constant tensors
\begin{align*}
& b_{kji}=0 \text{ if } k = 0 \text{ else } 1 \\
& c_{kji}=0 \text{ if } j \neq 0 \text{ else } 1 \\
\end{align*}
the external input tensor $U_{kji}$ and the same predetermined $u_{000} = 1, u_{kj0} = 0, \text{ and } u_{0ji} = 0, \forall j \neq 0$. The fully recurrent model (\ref{mm100}) is a generalization of all previous models which can be derived from it under specific assumptions like $w_{jji}=0$.

Notice that all the introduced models don't consider delay in their dynamics which is a must for actual physical implementation and also widely studied in models of neural networks. The reason for this is that we are more interested in the purely computational side of these models rather than their biological and physical realism similarly to the way artificial neural networks have deviated away from their biological counterparts. The models are also static in terms of network connectivity or synapse plasticity and no learning takes place in the evolution of the network activity. This is because we need to reduce the complexity of the overall system in order to study in depth and validate the newly introduced structure and potential dynamics. For the sake of completeness of this text, we will at least suggest a learning rule that could be coupled with any of the forms above.

To add plasticity to any metanetwork, we simply have to add dynamics for the weights of all connections and metaconnections. Despite the popularity of Hebbian learning, we will suggest a different learning rule based not on correlation of presynaptic and postsynaptic neurons but solely on the activity of presynaptic neurons. The reasoning behind this choice is related with the phenomenon of consolidating information just by recalling it which also causes inherent difficulty in intentional forgetting. A simple such access-based learning rule that satisfies many additional requirements is
\begin{equation} \label{access_based_rule}
\tau_w \frac{dw}{dt} = -uw + u^2
\end{equation}
where $w$ carries the index of $u$, i.e. it is the weight of the connection $u$. Of course the time scale for the weight dynamics $\tau_w$ must be significantly larger than the one for the potential $\tau_r$. The specific form of this learning rule avoids forgetting everything that is not recalled within a fixed time window by simply turning off modifications of the weight for $u=0$ when the connection is not used. Otherwise the weight slowly converges to the current value of the potential but still with rate proportional to the potential and the difference $u(u-w)$. The fact that larger potential has more effect on changing the weight has to do with avoiding simple linear learning, i.e. requiring that higher importance of learned information needs fewer repetitions. The greater effect of larger difference adds the importance of surprise and the difficulty of sustaining large weight with small potential for long. A discrete version of (\ref{access_based_rule}) is the update rule
\begin{equation} \label{access_based_update}
w \rightarrow w + \epsilon (-uw + u^2)
\end{equation}
The learning rule is stable and the convergence of $w$ to $u$ makes more sense if both of them are bounded from above by unity. Otherwise, a more complex rule can be considered where $w$ saturates as $u \rightarrow \infty$. The learning rule is also completely local - it couples only with the local potential and uses no global information from the network. However, it poses more challenges to obtain meaningful stored information with such a learning rule rather than Hebbian rule. For instance, to provide any feedback about the success or failure of an action potential further along the processing path, it is necessary to provide feedback connections from the upper layers back to the lower ones and a fully recurrent network is a crucial requirement. The learning rule also cannot change the sign of the weight for any positive potential and inhibitory connections must be predetermined. Both of these computational drawbacks match actual anatomical observations.

\subsubsection{Motivation for the family of models}

We complete this part with motivation about the currently formulated family of models. The simplest form (\ref{mm1}) has approximating power similar to that of regular multilayer networks (Section \ref{universal_approximator}). The internal potential form (\ref{mm2}) can be interpreted as an expanded graph where each node can represent a unit or a connection, all with the dynamics of an additive fully recurrent network (Section \ref{expanded_graph_interpretation}). This interpretation helps us use existing theory for its well-posedness and multistability and preserve all cognitive capabilities of (\ref{fr_ffn}, \ref{fr_rn_lln}, \ref{fr_rn_ein}) which are only particular cases of an additive network. At the same time, it reveals some of the limitations of the dynamics and the reasons they are also extended in later models in addition to the more elaborate network structure.

The full mathematical development of all later forms, i.e. models with sufficient complexity to motivate our choice of dynamics, is separated from its motivation which is the main topic of this section and starts with the main sections of the next part. Phase space decomposition of (\ref{mm3}, \ref{mm4}) provides with sufficient conditions on manifold-restricted stability (Section \ref{phase_space_decomposition}). Case study of (\ref{mm5}, \ref{mm11}, \ref{mm12}, \ref{mm13}) network motifs extends this to attractors in the full phase space and derives the network level dynamics necessary for the desired extra capabilities we will show here (Section \ref{network_motif_study}). Throughout the entire analysis we use (\ref{mm100}) reduced to the various forms above wherever we don't require more restricted network topology.

The main purpose for introducing this family of models is to be able to describe cognitive level phenomena only through their network level dynamics similarly to the selective amplification, input integration, etc. in (\ref{fr_ffn}), (\ref{fr_rn_lln}), and (\ref{fr_rn_ein}) but for activities involving compression of information and minimization of uncertainty. The following figures are some examples of such activities and network circuits that realize them. One could argue that organisms with a central nervous system that minimizes the uncertainty in the environment have an evolutionary advantageous processing capabilities but this is by no means the starting point for our definitions.

\begin{figure}[!htbp]
\centering
\includegraphics[width=0.5\textwidth]{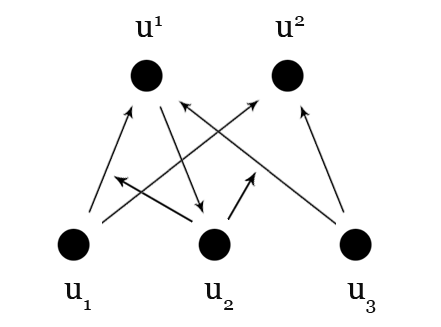}
\caption{Network circuits realizing attention to input.}
\label{circuit1}
\end{figure}

Figure \ref{circuit1} shows a circuit which although oversimplified is exemplary for many aspects of network's behavior. The input $u_{1}u_{2}u_{3}$ could be of different modality: a 3-pixel/photoreceptor grayscale image where each input contains a luminosity value, 3-letter word $ABC$ where each input is binary and reflects whether the letter was parsed from a text stream, etc. The outputs $u^1$ and $u^2$ are stored \emph{concepts} or rather abstractions from the input. The two output connections of $u_1$ respectively to $u^1$ and $u^2$ can be denoted as $u_{1\ }^{\ 1}$ and $u_{1\ }^{\ 2}$ and are the different \emph{interpretations} of the $u_1$ input, one for each concept. Reversely, the input connections $u_{1\ }^{\ 1}$ and $u_{3\ }^{\ 1}$ are \emph{representations} of the concept $u^1$.

The connectivity in the circuit \ref{circuit1} is preselected so that more local definitions can be taken to the surface. If we consider an input like $A\bar{B}\bar{C}$, i.e. $u_1=1$, $u_2=0$, and $u_3=0$, the only active input $u_1$ will \emph{broadcast} its signal to both $u_{1\ }^{\ 1}$ and $u_{1\ }^{\ 2}$. \emph{Uncertainty} results from the input $A\bar{B}\bar{C}$ because no single interpretation of $u_1$ is available. However, due to the connectivity in the circuit, this uncertainty will eventually be \emph{resolved} by activating the metaconnection $u_{2\ }^{\ 11}$ and therefore increasing the potential in $u_{1\ }^{\ 1}$. In this way, $u_2$ or more precisely its interpretation $u_{2\ }^{\ 11}$ is a critical \emph{context} for the right interpretation $u_{1\ }^{\ 1}$ of the input $u_1$. The same would be true for $u_3$ if it was active throughout and the input was rather $A\bar{B}C$ - $u_2$ would be the context to \emph{explain} the additional input $u_3$ as $u_{3\ }^{\ 1}$. Since $u_2$ directs both $u_1$ and $u_3$ to $u^1$, it can be considered as a \emph{feature} of $u^1$ that has greater importance for activating $u^1$ and helping it win against \emph{alternative} (same representations) concepts like $u^2$. Reversely, $u_2$ together with $u_1$ and $u_3$ are \emph{components} (same interpretations) with respect to $u^2$.

Taking a deeper look at the motivation behind some of the dynamics, we have to trace the processing of the input $A\bar{B}\bar{C}$ in more detail. Due to the broadcasting of $u_1$ and no previous stored potential in the two output connections, we have that both $u_{1\ }^{\ 1}=0.5$ and $u_{1\ }^{\ 2}=0.5$. Treating the potential of $u_1$ as a \emph{resource} that is split among the output connections helps in pathological cases where an input is learned so well (many interpretations) that it overtakes or \emph{cannibalizes} the network in terms of activation. In this case, the emitted signal will \emph{diffuse} among the many output connections, so that each connection will have insignificant effect unless it is \emph{contextualized} (learned with more metaconnections to it) and \emph{supported} by the present context. Although $u_{1\ }^{\ 1}=0.5<\alpha$ if the threshold $\alpha=1$, the potential storage as a property of the network will eventually lead to the activation of $u^1$ if $u_1$ keeps firing with the same rate for longer. At the same time, a threshold in general will help avoiding reaction to arbitrarily small inputs (e.g. a diffused input) and bring to sparse overall activity. Emitting through a feedback connection $u^{1\ }_{\ 2}$ represents network's \emph{assumption} and control over its own input as well as the \emph{imagination}/reconstruction of a learned input $AB\bar{C}$ achieved through the recall of the input $B$. The cycle $u^1 \rightarrow u^{1\ }_{\ 2} \rightarrow u_2 \rightarrow u_{2\ }^{\ 11} \rightarrow u_{1\ }^{\ 1}$ carries feedback for $u_{1\ }^{\ 1}$ from its own activation but this and any later analysis is beyond the scope of the current text. Increasing the emitting through $u_{1\ }^{\ 1}$ together with the signal resource policy leads to reduction of the emitting through $u_{1\ }^{\ 2}$ and therefore a form of \emph{weak inhibition} which does not involve specialized inhibitory connections. This is in accordance with empirical observations which show a much smaller number of inhibitory connections (in particular inhibitory interneurons or neurons withing the central nervous system) instead of a ratio of 1 \cite{OkunLampl2009} which is often employed in regular networks that need sufficient control of activation.

\begin{figure}[!htbp]
\centering
\includegraphics[width=0.75\textwidth]{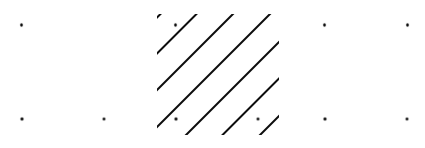}
\caption{Heterogeneity of representations of a triangle.}
\label{triangle_features}
\end{figure}

Taking another step towards credibility of such circuitry, we have to mention some cognitive level properties like the high heterogeneity of representations visualized by figure \ref{triangle_features}. A concept as an abstraction of the input, e.g. a triangle, can be very insensitive (robust) with respect to a lot of noise and corruption of the input as long as such variations are not features, and very sensitive to differences as small as a single extra activated input if this input is a feature of a concept. In the case of triangle, a fourth properly placed point would direct all previous points towards a square concept. Although features are most influential context for the rest of the overall input, they are also contextually influenced (however still broadcasting in comparison to fully resolved/explained inputs) and therefore are not structurally fixed for a given concept.

Another exemplary circuits could demonstrate cognitive level phenomena like vertical attention (towards uncertainty higher in the abstraction hierarchy), competition of concepts by spreading the influence of their features and concentrating the influence of competitive ones, etc. Of course, the case studied currently is an oversimplification of actual network circuits aimed at demonstrating a good amount of such phenomena with unrealistically few connections. Many definitions arising here should in fact be a lot more general:
\begin{itemize}
	\item interpretations/representations of a concept could be combinations of input/output connections also called \emph{internal states}
	\item a concept can be stored as an \emph{active state} (as a set of active units) or an \emph{active pattern} (temporal sequence of states) rather than a single unit (grandmother cell \cite{Gross2002})
\end{itemize}
All demonstrated behaviors are not guaranteed to match any empirical reality and all the models are not guaranteed to reproduce these behaviors. The first is not necessary as long as the network is capable of performing goals of sufficiently high generality, subject to evaluation through numerical implementations. The second is one of the main objectives of our current mathematical analysis together with providing guidance for new possible behaviors and dynamics to produce them.

\subsection{Mathematical analysis}

The mathematical development of the proposed models includes analysis of their approximating power (what kinds of input can be learned and represented), well-posedness (existence and uniqueness of their solutions or restriction to regions where these are provided), stability (multistability and multiperiodicity of simpler forms, partial stability for more complex forms), and overall dynamics (most complex forms, rather from simpler to more complex network motifs).

\subsubsection{Universal approximation} \label{universal_approximator}

The model (\ref{mm1}) is a universal approximator for any Borel measurable function similarly to a multilayer network with an arbitrary number of hidden units. In order to demonstrate this, we need some definitions and theorems from \cite{HornikStinchcombeWhite1989}.

\begin{definition}
For any $r \in \mathbb{N}$, $A^r$ is the set of all affine functions $A: \mathbb{R}^r \rightarrow \mathbb{R} :A(\mathbf{x}) = \mathbf{w} \cdot \mathbf{x} + b$ where $\mathbf{w}, \mathbf{x} \in \mathbb{R}^r$ and $b \in \mathbb{R}$.
\end{definition}

\begin{definition}
For any (Borel) measurable function $G(\cdot):\mathbb{R} \rightarrow \mathbb{R}$ and $r \in \mathbb{N}$, $\Sigma^r(G)$ is a class of functions $\{f: \mathbb{R}^r \rightarrow \mathbb{R} :f(\mathbf{x}) = \sum_{j=1}^q \beta_j G(A_j(\mathbf{x}))\}$ where $\mathbf{x} \in \mathbb{R}^r, \beta_j \in \mathbb{R}, A_j \in A^r, q \in \mathbb{N}$.
\end{definition}

\begin{definition}
For any (Borel) measurable function $G(\cdot):\mathbb{R} \rightarrow \mathbb{R}$ and $r \in \mathbb{N}$, $\Sigma\Pi^r(G)$ is a class of functions $\{f: \mathbb{R}^r \rightarrow \mathbb{R} :f(\mathbf{x}) = \sum_{j=1}^q \beta_j \prod_{k=1}^{l_j} G(A_{jk}(\mathbf{x}))\}$ where $\mathbf{x} \in \mathbb{R}^r, \beta_j \in \mathbb{R}, A_{jk} \in A^r, l_j \in \mathbb{N}, q \in \mathbb{N}$.
\end{definition}

\begin{definition}
A subset $S$ in a metric space $(C^r, \rho_k)$ where for $f,g \in C^r$ we have the metric $\rho_k(f,g)=\sup_{x \in K}|f(x)-g(x)|$ is said to be uniformly dense on compacta in $C^r$ if for every compact subset $K \subset \mathbb{R}^r$ we have that $S$ is $\rho_k$-dense in $C^r$.
\end{definition}

\begin{definition}
Given a probability measure $\mu$ on $(\mathbb{R}^r, B^r)$ define the metric $\rho_{\mu}: M^r \times M^r \rightarrow \mathbb{R}$ by $\rho_{\mu}(f,g) = \inf\{\epsilon > 0 : \mu\{x : |f(x) - g(x)| > \epsilon\} < \epsilon\}$.
\end{definition}

\begin{theorem}
\label{thm_uac}
Let $C^r$ be the set of continuous functions from $\mathbb{R}^r$ to $\mathbb{R}$. Let $G$ be any continuous nonconstant function from $\mathbb{R}$ to $\mathbb{R}$. Then $\Sigma\Pi^r(G)$ is uniformly dense on compacta in $C^r$.
\end{theorem}

\begin{theorem}
\label{thm_uam}
Let $M^r$ be the set of Borel measurable functions from $\mathbb{R}^r$ to $\mathbb{R}$ and $B^r$ the Borel $\sigma$-algebra of $\mathbb{R}^r$. For any continuous nonconstant function $G: \mathbb{R} \rightarrow \mathbb{R}$, every $r \in \mathbb{N}$, and every probability measure $\mu$ on $(\mathbb{R}^r, B^r)$ we have that $\Sigma\Pi^r(G)$ is $\rho_\mu$-dense in $M^r$.
\end{theorem}

Thus a function representing the instant output of a neuron $f(\mathbf{x}) \in \Sigma^r(G) \subset \Sigma\Pi^r(G)$ can approximate any continuous function arbitrary well when the input is restricted to a compact interval $x_i \in [a,b]$ from the first theorem and any Borel measurable function in a probability measure from the second theorem. A network with an input layer size $r$, hidden layer size $q$, and output layer size $1$ has an output unit of the form $f \in \Sigma\Pi^r(G)$, i.e. $f = \sum_{j=1}^q \beta_j \prod_{k=1}^1 G(\mathbf{w}_{jk} \cdot \mathbf{x} + b_{jk}) = \sum_{j=1}^q \beta_j G(\mathbf{w}_j \cdot \mathbf{x} + b_j) = \sum_{j=1}^q \beta_j G(\sum_{i=1}^r w_{ij} x_i + b_j)$. The input units are respectively $x_i$, the weights to a hidden unit are represented by the vector $\mathbf{w}_j$ while its potential is $A_j=\mathbf{w}_j \cdot \mathbf{x} + b_j$ with $b_j$ representing the bias term (external input) for the $j$-th hidden unit. The activation function for a hidden unit is $G(A_j)$ and each weight to the output unit is represented by $\beta_j$. For an output layer of size $s$ we simply take a vector valued function so that $f_i \in \Sigma^r(G)$ for $i \in [0;s]$.

We will now show that (\ref{mm1}) has the same approximating power by considering the output of a neuron $v_i=g(\mathbf{x})$. In particular, consider a linear variant of (\ref{mm1}) with $\alpha=0$
\begin{align*}
\begin{dcases}
& \tau_r \frac{dv_i}{dt} = - v_i + \sum_{j=1}^N w_j^i c_j^i\\
& \tau_r \frac{dc_j^i}{dt} = - c_j^i + \sum_{\substack{k = 1 \\ k \neq j}}^N w_k^{ji} c_k^{ji} + u_j\\
& \tau_r \frac{dc_k^{ji}}{dt} = - c_k^{ji} + u_k,\ k \neq j
\end{dcases}
\end{align*}
Assuming instantaneous influence of the variables we get
\begin{align*}
\begin{dcases}
& v_i = \sum_{j=1}^N w_j^i c_j^i\\
& c_j^i = \sum_{\substack{k = 1 \\ k \neq j}}^N w_k^{ji} c_k^{ji} + u_j\\
& c_k^{ji} = u_k,\ k \neq j
\end{dcases}
\Rightarrow
\begin{dcases}
& v_i = \sum_{j=1}^N w_j^i (\sum_{\substack{k = 1 \\ k \neq j}}^N w_k^{ji} u_k + u_j)\\
& c_j^i = \sum_{\substack{k = 1 \\ k \neq j}}^N w_k^{ji} u_k + u_j\\
& c_k^{ji} = u_k,\ k \neq j
\end{dcases}
\end{align*}
Thus, dropping the index $i$ and considering a single output $v$ ($M=1$) we obtain that
\begin{equation*}
v = g(\mathbf{u}) = \sum_{j=1}^N w_j (\sum_{\substack{k = 1 \\ k \neq j}}^N w_k^j u_k + u_j) = \sum_{j=1}^N w_j (\sum_{\substack{k = 1}}^N w_k^j u_k), \quad w_j^j=1
\end{equation*}
Finally, changing notation to $u_k=x_k$, $w_j=\beta_j$, $k=i$, $G(u)=u$, $q=r=N$, and using $b_j=0$ we can finally see that
\begin{equation*}
g(\mathbf{x}) = \sum_{j=1}^q \beta_j G(\sum_{i=1}^r w_i^j x_i) \Rightarrow g(\mathbf{x}) \in \Sigma^r(G) \subset \Sigma\Pi^r(G)
\end{equation*}
and since $G(u)=u$ is continuous nonconstant function and $x_i \in [0,1]$ by theorem \ref{thm_uac} we can approximate any continuous function. Alternatively, using theorem \ref{thm_uam} for $G(u)=u$ continuous nonconstant function we can conclude that the linear variant of (\ref{mm1}) is a universal approximator to any Borel measurable function on $r$ dimensions. We can consider the original (\ref{mm1}) without the dynamics and threshold on the output units in which case $G(u)=(u-\alpha)^+$ is still continuous and nonconstant. Therefore, we can still apply the theorems and conclude that an input-output layer network of the form (\ref{mm1}) is a universal approximator. An important distinction here is that while the original consideration of \cite{HornikStinchcombeWhite1989} was about any neural network with at least three layers (Section \ref{anns}), the network (\ref{mm1}) has only two layers. This is intuitive since (\ref{mm1}) has more complex dynamics like dynamics on the connections but also fits better the observation that the neocortex has only six layers (Section \ref{bnns}) to perform all its complex operations.

\subsubsection{Expanded graph interpretation} \label{expanded_graph_interpretation}

In the following we will establish some results available from the literature regarding a simplified version of (\ref{mm100}), namely the tensor form of (\ref{mm2}). We will show that (\ref{mm2}) is a particular case of the additive model (\ref{fr_rn_frn2n}) which is the most widely studied neural network with well established theory for its well-posedness and stability. We can derive (\ref{fr_rn_frn2n}) both by reducing the dynamics of (\ref{mm100}) to the dynamics of (\ref{mm2}) or by constructing a tensor form of (\ref{mm2}) similarly to the way it was done for (\ref{mm100}) but using the simpler dynamics. Here we will take the first approach and reduce (\ref{mm100}) to a fully recurrent network of the form
\begin{align*}
& \tau_r \frac{du_{kji}}{dt} = - u_{kji} + c_{kji} \sum_{l=1}^N w_{lki} (u_{lki} - \alpha)^+ + b_{kji} (u_{00k} - \alpha)^+ + U_{kji} =\\
& = - d_{kji} u_{kji} + c_{kji} \sum_{l=1}^N w_{lki} a(u_{lki}) + b_{kji} a(u_{00k}) + U_{kji}
\end{align*}
where $d_{kji}=1$ and $a(u)=(u-\alpha)^+$ is the activation function.

\begin{figure}[!htbp]
\centering
\includegraphics[width=0.75\textwidth]{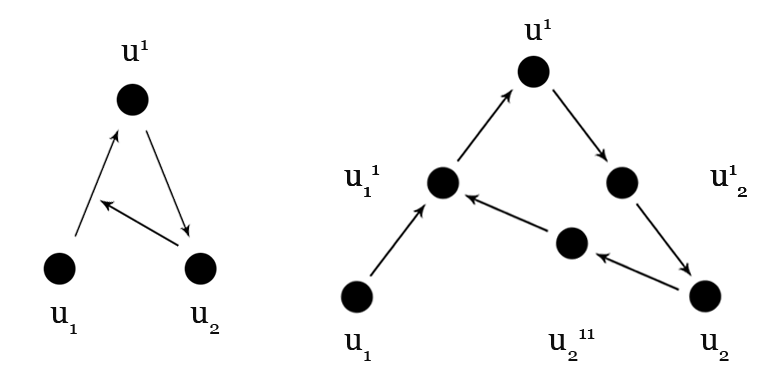}
\caption{Expanded graph version of a simple circuit.}
\label{expanded_graph}
\end{figure}

If we interpret the network as an expanded graph where each unknown, be it a unit, connection, or a metaconnection, is a node of the graph and the dependence among two unknowns is an edge (figure \ref{expanded_graph}), we can order the unknowns in a vector and obtain a sparse weight matrix of $(N+1)^3$ rows/columns. In particular,
\begin{align*}
& \tau_r \frac{du_{kji}}{dt} = - d_{kji} u_{kji} + c_{kji} \sum_{l=1}^N w_{lki} a(u_{lki}) + b_{kji} a(u_{00k}) + U_{kji} =\\
& = - d_{kji} u_{kji} + c_{kji} \sum_{l=0}^N \sum_{m=0}^N \sum_{n=0}^N w^{kji}_{lmn} a(u_{lmn}) + U_{kji} =\\
& = - d_{kji} u_{kji} + \sum_{l=0}^N \sum_{m=0}^N \sum_{n=0}^N \tilde{w}^{kji}_{lmn} a(u_{lmn}) + U_{kji}
\end{align*}
with $w_{lki}=w^{kji}_{lki}$, $\tilde{w}^{kji}_{lmn}=0$ if $l=0$, $m \neq k$ or $n \neq i$, and $\tilde{w}^{kji}_{00k}=b_{kji}$ (from $w^{kji}_{00k}=b_{kji}/c_{kji}$) which is well-defined since $b_{kji}$ depends only on $k$ and implies that $\tilde{w}^{kji}_{00k}=0 \text{ if } k = 0 \text{ else } 1$. Absorbing $c_{kji}$ in addition implies that $\tilde{w}^{kji}_{lmn}=0$ also if $j \neq 0$ with the exception of the case $\tilde{w}^{kji}_{00l}=1$.

After reindexing of $kji$ into $i$ and $lmn$ into $j$ to arrange $u_{kji}$ in a vector, the equation becomes
\begin{equation} \label{mm102}
\tau_r \frac{du_{i}}{dt} = - d_{i} u_{i} + \sum_{j=1}^{(N+1)^3} \tilde{w}_{ij} a(u_{i}) + U_{i} \iff \tau_r \frac{d\mathbf{u}}{dt} = - diag(\mathbf{d}) \mathbf{u} + \mathbf{\tilde{W}} \mathbf{a}(\mathbf{u}) + \mathbf{U}
\end{equation}
which is the additive model (\ref{fr_rn_frn2n}). We can therefore apply all theory for (\ref{fr_rn_frn2n}) to draw conclusions on the existence and uniqueness of its solutions, as well as of its attractors. Most of the available results involve sufficient conditions to guarantee the uniqueness of a stable equilibrium point (monostability) due to the large applicability of such networks to optimization \cite{ZhangWangLiu2014} but we are more interested in the existence of multiple stable equilibrium points (multistability) and multiple stable limit cycles (multiperiodicity) of (\ref{mm102}).

The existence and uniqueness of solutions clearly depends on the choice of activation function $a(\cdot)$. They are guaranteed for a smooth as well as Lipschitz continuous $a(\cdot)$ by the Picard-Lindel\"{o}f theorem from the standard theory for ODEs. In the case of discontinuous activation functions we need to make use of Filippov theory and define a solution (also if it is an equilibrium point or a limit cycle) in the sense of differential inclusion \cite{FilippovArscott1988}.

The existence (but not uniqueness) of attractors is studied predominantly for bounded activation functions (saturating or also called squashing functions) which can be interpreted as a maximal output firing rate. However, unbounded activation functions are also of great relevance and included in the stability results of many papers \cite{ZhangWangLiu2014}. A final distinction in the type of activation function is about monotonicity where the first assumptions are on strictly monotonically increasing $a(\cdot)$ but later on extended to nondecreasing cases.

As discussed in the study of (\ref{fr_rn_frn2n}), the activation function $a(u)=(u-\alpha)^+$ is globally Lipschitz continuous so existence, uniqueness and continuity with respect to initial data are all immediately verifiable and hold for all times. If we take a step of extension towards discontinuous activation functions, we need some more general definitions similar to the ones found in the first investigations of the topic for monostability \cite{FortiNistri2003} and multistability \cite{LiMichelPorod1989}.

\begin{definition}[Convex hull]
For a set $E \subset \mathbb{R}^n$, the smallest convex set containing $E$ is called the convex hull of $E$. The convex hull always exists and is the intersection of all convex sets containing $E$. The closure of the convex hull of $E$ will be denoted as $K[E]$.
\end{definition}

The convex hull is a linear operator, i.e. $K[\alpha E_1+\beta E_2] = \alpha K[E_1] + \beta K[E_2]$. This property is used in the next definition.

\begin{definition}[Solution]
Let us consider the set-valued map $\phi:\mathbb{R}^n \mapsto \mathbb{R}^n$ defined as
\begin{equation*}
\phi(\mathbf{u}) = \bigcap_{\epsilon>0} \bigcap_{\mu(N)=0} K[\mathbf{f}(B(\mathbf{u},\epsilon) \setminus N)] = -diag(\mathbf{d}) \mathbf{u} + \mathbf{\tilde{W}} K[\mathbf{a}(\mathbf{u})] + \mathbf{U}
\end{equation*}
where $N$ is an arbitrary set with measure zero and $B$ is a ball centered at $\mathbf{u}$ with a radius $\epsilon$. A solution of (\ref{mm102}) on an interval $[t_0,t_1]$, $t_0 \leq t_1 \leq +\infty $, with initial condition $\mathbf{u}(t_0)=\mathbf{u_0}$, is an absolutely continuous function $\mathbf{u}(t)$ defined on $[t_0,t_1]$, such that $\mathbf{u}(t_0)=\mathbf{u_0}$, and for almost all $t \in [t_0,t_1]$, $\mathbf{u}(t)$ satisfies the differential inclusion $\dot{\mathbf{u}}(t) \in \phi(\mathbf{u}(t))$.
\end{definition}

Here, $K[\mathbf{a}(\mathbf{u})]=([a_1(u_1^-), a_1(u_1^+)], \cdots [a_N(u_N^-), a_N(u_N^+)])^T$ is a vector of intervals defined by each point of discontinuity of $\mathbf{a}(\mathbf{u})$. Since equilibria are themselves solutions, we define them here as well.

\begin{definition}[Equilibrium point]
An equilibrium point $\mathbf{u^*} \in \mathbb{R}^n$ of (\ref{mm102}) is a constant solution of (\ref{mm102}) $\mathbf{u}(t)=\mathbf{u^*}, \forall t \in [0,+\infty)$. Thus $\mathbf{u^*}$ satisfies
\begin{equation*}
\mathbf{0} \in \phi(\mathbf{u^*}) = -diag(\mathbf{d}) \mathbf{u^*} + \mathbf{\tilde{W}} K[\mathbf{a}(\mathbf{u^*})] + \mathbf{U}
\end{equation*}
\end{definition}

\begin{definition}[Output equilibrium point]
If $\mathbf{u^*}$ is an equilibrium point of (\ref{mm102}), then there exists a vector $\mathbf{v^*} \in \mathbb{R}^n$ such that
\begin{equation*}
\mathbf{v^*} \in K[a(\mathbf{u^*})] \qquad \mathbf{0} = -diag(\mathbf{d}) \mathbf{u^*} + \mathbf{\tilde{W}} \mathbf{v^*} + \mathbf{U}
\end{equation*}
which is called an output equilibrium point of (\ref{mm102}) corresponding to $\mathbf{u^*}$.
\end{definition}

It is important to consider the output equilibria separately from the state equilibria because the activation function is no longer assumed to be continuous, i.e. convergence in the state space does not guarantee convergence in the outputs as was the case before.

\begin{theorem}[Local existence of solutions]
Suppose that $a: \mathbb{R}^n \rightarrow \mathbb{R}^n$ is such that each component $a_i$ of $a$ is nondecreasing, bounded, and almost everywhere continuous with finite left and right limits at each point of discontinuity. Then $\forall \mathbf{u_0} \in \mathbb{R}^n$ there is at least a local solution of $\mathbf{u}(t)$ of (\ref{mm102}) with initial condition $\mathbf{u}(0)=\mathbf{u_0}$. Furthermore, any solution is bounded and hence defined on $[0, +\infty)$.
\end{theorem}

The local existence is a consequence of \cite{FilippovArscott1988} and can also be derived in the case where $a_i$ is not necessarily bounded. The global existence is proved by \cite{FortiNistri2003} with the ultimate goal of providing sufficient conditions for monostability. On the other hand, \cite{LiMichelPorod1989} used phase space decomposition to study multistability of (\ref{mm102}) with $a(u)=sgn(u)$ as an activation function.

There are numerous studies in the case of smooth activation \cite{LiMichelPorod1988, SudharsananSundareshan1991, FangKincaid1996, MichelFarrellPorod1989, ChenAmari2001, GuezProtopopsecuBarhen1988, CaoTao2001} using techniques from custom energy functionals \cite{CaoTao2001} to matrix measures \cite{FangKincaid1996} and providing results on local and global exponential stability \cite{LiMichelPorod1988, SudharsananSundareshan1991, FangKincaid1996, MichelFarrellPorod1989, ChenAmari2001, GuezProtopopsecuBarhen1988}, convergence rate \cite{FangKincaid1996, CaoTao2001}, basins of attraction \cite{MichelFarrellPorod1989, SudharsananSundareshan1991, CaoTao2001}, and network synthesis/design \cite{LiMichelPorod1988, SudharsananSundareshan1991, GuezProtopopsecuBarhen1988}. More recent results using continuous saturating (often piecewise linear) activation functions have identified the possibility of $3^N$ equilibria from which $2^N$ are asymptotically stable and the rest unstable. In particular, \cite{ZengWang2006} analyzed n-neuron network with time-varying delay and found conditions for the existence of $2^N$ locally exponential attractive limit cycles (equilibria being a special case and a corollary). In \cite{ChengLinShih2006,ChengLinShih2007}, the number of possible equilibria was extended to $3^N$ with the $3^N-2^N$ equilibria proven to be unstable by \cite{WangLuChen2010, LuWangChen2011} who also showed that $3^N$ equilibria is a particular case for an activation function with 2 corner points (assuming piecewise linearity), the more general result for $2r$ corner points being $(2r+1)^N$ equilibria, from which $(r+1)^N$ locally exponentially stable and the rest unstable. Similar results on the stable equilibria is shown by \cite{HuangCao2008, ChenZhouSong2010, HuangZhangWang2012} also for discontinuous piecewise constant activation functions with $s$ segments, namely $s^N$ locally exponentially stable equilibrium points (or limit cycles if considering periodic external input) with $a(u)=sgn(u)$ as a particular case with $s=2$.

Choosing a continuous (piecewise linear) activation function with arbitrary number of corners or a discontinuous activation function with arbitrary number of segments, one may argue that we can finally achieve arbitrary high capacity for the number of the stored memories instead of the very limited first estimation of $0.15N$ for (\ref{fr_rn_frn2n}). This is not true since one also needs synthesis procedure in order to store preselected memories and Hopfield's early estimate of $0.15N$ is also due to his early synthesis procedure (\ref{syn2}). In particular, if we consider a large number of memories that are all correlated in some units (vectors correlated in some components), we can obtain large weights and violate the sufficient conditions for the $3^n$ equilibria by \cite{ZengWang2006, ChengLinShih2006, WangLuChen2010}.

There are also some other drawbacks in the associative memory interpretation of (\ref{fr_rn_frn2n}) and therefore (\ref{mm2}) that are indicative of the excessive simplicity of its dynamics:
\begin{itemize}
	\item The existence of unstable equilibria does not have proper interpretation in terms of memories although the instability will leave the equilibrium undetectable in practice due to unavoidable perturbations.
	\item Under the sufficient conditions for existence of stable equilibria \cite{ZengWang2006, ChengLinShih2006}, each equilibrium would also have a counterpart which is symmetric with respect to the origin, leaving only a total of $1^n$ equilibria in the fully positive region which might sometimes be a meaningful restriction.
	\item Results on convergence \cite{ZengWang2006, ChengLinShih2006, WangLuChen2010} might be too restrictive since the main methods to obtain them involve decomposing the phase space and detecting positive invariant regions where the equations remain fully decoupled. This means that an equilibrium along a dimension will remain constant along any other dimension and more interesting cases cannot be considered.
	\item Results on the basins of attraction \cite{ChengLinShih2006, WangLuChen2010} are usually descriptions of convex positive invariant sets bounded by the stable manifolds of unstable equilibria. In many cases this might be too simple definition of the kinds of initial conditions that are pooled together for a robust response, possibly involving partial convergence or a different notion of proximity of clues.
	\item Intensive partitioning of the activation function would also lead to many small attraction regions rendering the network useless for pattern completion, classification, or any other purpose. It is also biologically less meaningful compared to simpler activation functions like $a(u)=(u-\alpha)^+$ that better reflect the firing rate approximation.
\end{itemize}

Because it is unbounded, the activation function $a(u)=(u-\alpha)^+$ has more complex dynamics, fits better observations that neurons rarely operate in upper saturating regime, and handles spurious equilibrium points in winner-takes-all networks \cite{WersingBeynRitter2001}. Results on this type of linear threshold function involve sufficient conditions on boundedness of the trajectories (for symmetric and nonsymmetric connectivity) \cite{WersingBeynRitter2001,YiTanLee2003,ZhangYiYu2008}, determination of global attractive sets  \cite{YiTanLee2003,ZhangYiYu2008}, complete convergence  \cite{YiTanLee2003}, and existence of exponentially stable limit cycles (and therefore equilibria) in invariant sets \cite{ZhangYiYu2008}.

\subsubsection{Phase space decomposition} \label{phase_space_decomposition}

We can derive a tensor form of (\ref{mm3}, \ref{mm4}) from (\ref{mm100}) similarly to the way we did it in the previous section or analogically to the way we obtained (\ref{mm100}) from (\ref{mm5}) to get
\begin{align*}
& \tau_r \frac{du_{kji}}{dt} = - d_{kji} u_{kji} + c_{kji} \sum_{l=1}^N w_{lki} a(u_{lki}) + b_{kji} \bar{e}(u_{kji},u_{00k},\dots) + U_{kji}\\
& \tau_r \frac{du_{kji}}{dt} = - a(u_{kji}) + c_{kji} \sum_{l=1}^N w_{lki} a(u_{lki}) + b_{kji} \bar{e}(u_{kji},u_{00k},\dots) + U_{kji}
\end{align*}
We will now emphasize on the first equation (remembering that $w_{kki}=0$) and write it explicitly with the context term expanded as the sum of indicator functions for each condition
\begin{equation} \label{mm103}
\begin{split}
& \tau_r \frac{du_{kji}}{dt} = - d_{kji} u_{kji} + c_{kji} \sum_{l=1}^N w_{lki} a(u_{lki}) + U_{kji} + \\
& b_{kji} [\chi(u_{k01},\dots,u_{kNN}) \frac{u_{kji}}{\sum\limits_{l=0}^N \sum\limits_{m=1}^N u_{klm}} + (1-\chi(u_{k01},\dots,u_{kNN})) \frac{1}{N(N+1)}] a(u_{00k})
\end{split}
\end{equation}
where we use the indicator function
\begin{equation} \label{chi1}
\chi(u_{k01},\dots,u_{kNN}) =
\left\{
	\begin{array}{ll}
		1 \text{ if } \sum_{l=0}^N \sum_{m=1}^N u_{klm} \neq 0\\
		0 \text{ otherwise}
	\end{array}
\right.
\end{equation}

Decomposing the phase space into regions according to the choice of activation function helps us obtain simpler dynamics and conditions on local stability in each region. This is one of the major approaches in the study of multistability of neural networks in literature \cite{ZhangWangLiu2014} and is done both for continuous and discontinuous activation functions. In our case however, we will also consider affine subspaces where the context term simplifies to broadcasting dynamics and their positive side where the distributed emitting is nonsingular. Without loss of generality, we will also assume that $\alpha=0$, i.e. $a(u)=(u)^+=\max(0,u)$.

More specifically, if we fix $\bar{k} \in [1;N]$, then in the half-space $\{u_{00\bar{k}} \leq \alpha = 0\}$ we have that $\forall j \in [0;N], \forall i \in [1;N]$ the equations simplify to
\begin{equation*}
\tau_r \frac{du_{\bar{k}ji}}{dt} = - d_{\bar{k}ji} u_{\bar{k}ji} + c_{\bar{k}ji} \sum_{l \in \mathcal{N}} w_{l\bar{k}i} u_{l\bar{k}i} + U_{\bar{k}ji}
\end{equation*}
where $\mathcal{N} \subset [1;N]$ is such that $\bar{l} \in \mathcal{N}$ if $u_{\bar{l}\bar{k}i} > \alpha = 0$. The same reduction is true for the entire phase space for the case $\bar{k}=0$ since $b_{0ji}=0$ and the following conclusions extend to this case without effort. If we vary along a specific dimension $u_{\bar{k}\bar{j}\bar{i}}$ and fix all the rest ($k \neq \bar{k}, j \neq \bar{j}, i \neq \bar{i}$), we can find a stable equilibrium along the resulting 1-dimensional manifold as
\begin{equation*}
u_{\bar{k}\bar{j}\bar{i}} = \frac{c_{\bar{k}\bar{j}\bar{i}}}{d_{\bar{k}\bar{j}\bar{i}}} \sum_{l \in \mathcal{N}} w_{l\bar{k}\bar{i}} u_{l\bar{k}\bar{i}} + \frac{U_{\bar{k}\bar{j}\bar{i}}}{d_{\bar{k}\bar{j}\bar{i}}}
\end{equation*}
This partial equilibrium is stable because the differential equation simplifies to the somewhat canonical form $\dot{y}=-y+c$ with a stable equilibrium at $c$. It varies linearly with $u_{l\bar{k}\bar{i}}$ and is constant with respect to all other unknowns. In both cases the dynamical behavior along the other dimensions remains qualitatively the same as long as $u_{00\bar{k}} \leq 0$ but in the second case it also remains quantitatively the same. If we vary $u_{\bar{k}ji}$ and fix all the rest ($k \neq \bar{k}$), we can find a stable equilibrium on the resulting $(N+1)N$-dimensional manifold as
\begin{equation*}
u_{\bar{k}ji} = \frac{c_{\bar{k}ji}}{d_{\bar{k}ji}} \sum_{l \in \mathcal{N}} w_{l\bar{k}i} u_{l\bar{k}i} + \frac{U_{\bar{k}ji}}{d_{\bar{k}ji}}
\end{equation*}
remembering again that $w_{\bar{k}\bar{k}i}=0$ and therefore that the above $N(N+1)$ equations are mutually decoupled. Furthermore, if $c_{\bar{k}ji}=0$ or $u_{l\bar{k}i} \leq \alpha = 0$ for $l, i \in [1;N]$ (and therefore $\mathcal{N} = \emptyset$) holds then $u_{\bar{k}ji} = \frac{U_{\bar{k}ji}}{d_{\bar{k}ji}}$ is an equilibrium in the $N(N+1)$-dimensional manifold as before but constant with respect to all other dimensions, i.e. with identical quantitative behavior along $u_{kji}$ for $k \neq \bar{k}$. Finally, if we are in the subthreshold region $u_{kji} \leq \alpha = 0$ for all unknowns, we can find a stable equilibrium in the phase space
\begin{equation*}
u_{kji} = \frac{U_{kji}}{d_{kji}} \quad \forall k,i \in [1;N], j \in [0;N]
\end{equation*}
which would imply noninteracting units whose potentials balance at the ratio of each external input and decay rate.

If we restrict the dynamics to the subspace $\{\sum_{l=0}^N \sum_{m=1}^N u_{\bar{k}lm}=0\}$ intersected with the remaining half-space $\{u_{00\bar{k}} > \alpha = 0\}$, we get that $\chi(u_{k01},\dots,u_{kNN})=0$ and $\forall j \in [0;N], \forall i \in [1;N]$ the equations simplify to
\begin{equation*}
\tau_r \frac{du_{\bar{k}ji}}{dt} = - d_{\bar{k}ji} u_{\bar{k}ji} + c_{\bar{k}ji} \sum_{l \in \mathcal{N}} w_{l\bar{k}i} u_{l\bar{k}i} + b_{\bar{k}ji} \frac{u_{00\bar{k}}}{N(N+1)} + U_{\bar{k}ji}
\end{equation*}
resulting in partial equilibrium on the $N(N+1)$-dimensional manifold (and its lower dimensional counterparts) of the form
\begin{equation*}
u_{\bar{k}ji} = \frac{c_{\bar{k}ji}}{d_{\bar{k}ji}} \sum_{l \in \mathcal{N}} w_{l\bar{k}i} u_{l\bar{k}i} + \frac{b_{\bar{k}ji} u_{00\bar{k}}}{d_{\bar{k}ji} N(N+1)} + \frac{U_{\bar{k}ji}}{d_{\bar{k}ji}}
\end{equation*}
If $u_{l0\bar{k}} \leq 0$ and $u_{l\bar{k}i} \leq 0$ (or $c_{\bar{k}ji}=0$) for $l \in [1;N], l \neq \bar{k}$, we have in addition that $u_{00\bar{k}} = \frac{U_{00\bar{k}}}{d_{00\bar{k}}}$ and therefore
\begin{equation*}
u_{\bar{k}ji} = \frac{b_{\bar{k}ji} U_{00\bar{k}}}{d_{\bar{k}ji} d_{00\bar{k}} N(N+1)} + \frac{U_{\bar{k}ji}}{d_{\bar{k}ji}}
\end{equation*}
which is fully decoupled and therefore constant with respect to all other dimensions. Notice that all of this holds on a $(N-1)N(N+1)$-dimensional manifold which is the subspace $\{\sum_{l=0}^N \sum_{m=1}^N u_{\bar{k}lm}=0\}$.

\begin{figure}[!htbp]
\centering
\includegraphics[width=1.0\textwidth]{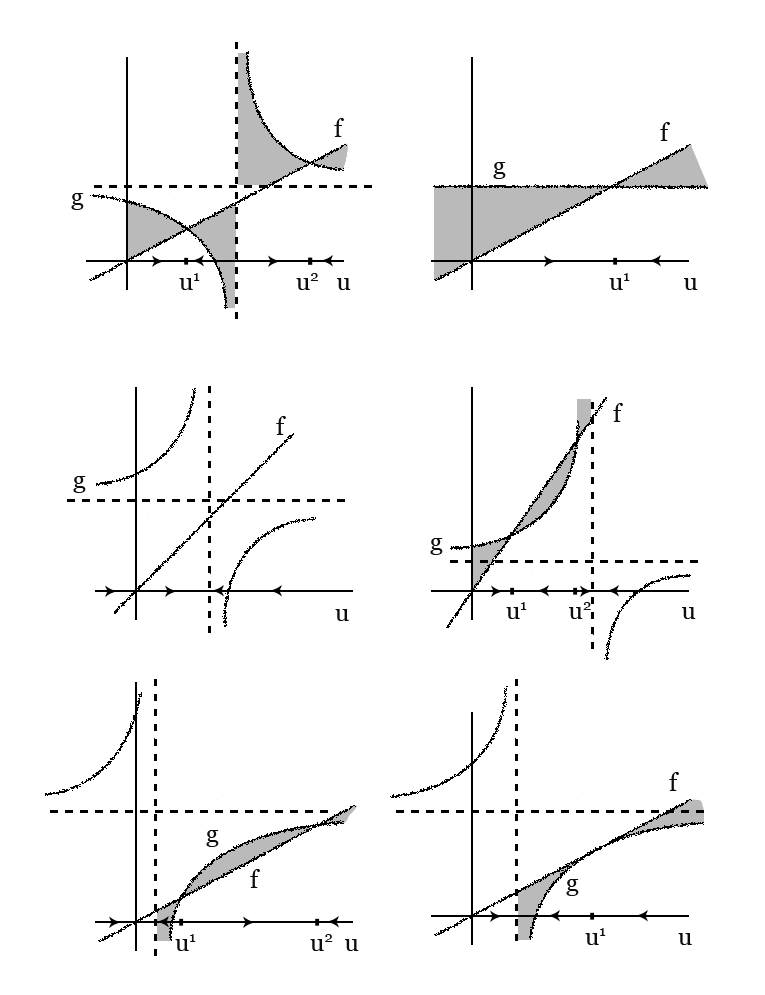}
\caption{The different qualitative cases of the stability of $u_{\bar{k}ji}$: a) two locally stable equilibria; b) one globally stable equilibrium; c) no stable equilibria; d,e) one stable and one unstable equilibrium in reversing order; f) one semistable equilibrium (only one of two cases shown);}
\label{geometrical_setting}
\end{figure}

Up until this point we haven't added much to the dynamics but a scaled down version of a particular interaction with a particular unknown $u_{00\bar{k}}$. The remainder of the phase space, namely the region  $\{\sum_{l=0}^N \sum_{m=1}^N u_{\bar{k}lm} \neq 0\} \cap \{u_{00\bar{k}} > \alpha = 0\}$ is the main regime of operation of the network and leads to equations like
\begin{equation*}
\tau_r \frac{du_{\bar{k}ji}}{dt} = - d_{\bar{k}ji} u_{\bar{k}ji} + c_{\bar{k}ji} \sum_{l \in \mathcal{N}} w_{l\bar{k}i} u_{l\bar{k}i} + b_{\bar{k}ji} \frac{u_{\bar{k}ji} u_{00\bar{k}}}{\sum\limits_{l=0}^N \sum\limits_{m=1}^N u_{\bar{k}lm}} + U_{\bar{k}ji}
\end{equation*}
where we have that $u_{\bar{k}\bar{k}m}=0$ from the definition of (\ref{mm3}). To obtain the partial equilibria now, we will define the remainder sum
\begin{equation} \label{r_sum}
r_{kji} := (\sum_{l = 0}^N \sum_{m = 1}^N u_{klm}) - u_{kji}
\end{equation}
and use geometrical arguments. For the purpose of readability and since we won't modify anything about the indices until the end of the section, we will use a simplified abusive notation that drops them altogether as follows
\begin{align*}
a &:= r_{\bar{k}ji} & c &:= U_{\bar{k}ji} + c_{\bar{k}ji} \sum_{l \in \mathcal{N}} w_{l\bar{k}i} u_{l\bar{k}i}\\
b &:= b_{\bar{k}ji}u_{00\bar{k}} & d &:= d_{\bar{k}ji}
\end{align*}
This implies that the equilibria satisfy
\begin{equation*}
0 = - d u_{\bar{k}ji} + \frac{b u_{\bar{k}ji}}{a + u_{\bar{k}ji}} + c \Rightarrow f(u_{\bar{k}ji}) = d u_{\bar{k}ji} = \frac{b u_{\bar{k}ji}}{a + u_{\bar{k}ji}} + c = g(u_{\bar{k}ji})
\end{equation*}
which has three different cases a), b), and the rest shown on figure \ref{geometrical_setting}.

When $-ba > 0 \Rightarrow b_{\bar{k}ji}u_{00\bar{k}} r_{\bar{k}ji} < 0$, we obtain case a) where we have two stable equilibria $\bar{u}_{\bar{k}ji}^1$ and $\bar{u}_{\bar{k}ji}^2$ on the two sides of a singularity which is located at $u_{\bar{k}ji}=-a= -r_{\bar{k}ji}$. Clearly, the context term will dominate the equation for $\dot{u}_{\bar{k}ji}$ for values $-a \pm \epsilon$ for an arbitrary small $\epsilon > 0$ where it will be positive for $-a+\epsilon$. Since $f<g$ in $(-\infty,\bar{u}_{\bar{k}ji}^1) \cup (-b,\bar{u}_{\bar{k}ji}^2)$ and $f>g$ in $(\bar{u}_{\bar{k}ji}^1,-b) \cup (\bar{u}_{\bar{k}ji}^2,+\infty)$, the stability of each equilibrium can be verified by noticing that the derivative $\dot{u}_{\bar{k}ji}$ increases on its left and decreases on its right side. In particular, on the left of each equilibrium point $u_{\bar{k}ji}=\bar{u}_{\bar{k}ji}^{1/2}-\epsilon$ we have that $f(u_{\bar{k}ji}) < g(u_{\bar{k}ji}) \Rightarrow d u_{\bar{k}ji} < \frac{b u_{\bar{k}ji}}{a + u_{\bar{k}ji}} + c \Rightarrow 0 < d u_{\bar{k}ji} + \frac{b u_{\bar{k}ji}}{a + u_{\bar{k}ji}} + c = \dot{u}_{\bar{k}ji}$ and analogically on the right both implying positively invariant region $(\bar{u}_{\bar{k}ji}^{1/2}-\epsilon, \bar{u}_{\bar{k}ji}^{1/2}+\epsilon)$ and by the arbitrariness of $\epsilon>0$ the stability of $\bar{u}_{\bar{k}ji}^{1/2}$. The stability is local since $f < g$ and $f > g$ each hold on a sufficiently small interval and the basins of attraction are respectively $(-\infty,-a)$ for $\bar{u}_{\bar{k}ji}^1$ and $(-a,\infty)$ for $\bar{u}_{\bar{k}ji}^2$. The asymptote for arbitrary large $u_{\bar{k}ji}$ is $b+c$ which is related with the case b).

When $-ba = 0 \Rightarrow b_{\bar{k}ji}u_{00\bar{k}} r_{\bar{k}ji} = 0$, we obtain case b) where we have one globally stable equilibrium $\bar{u}_{\bar{k}ji}$ and no singularity. Our $g(u_{\bar{k}ji})=b+c$ is constant function equal to the asymptotic limit and since $f(u_{\bar{k}ji}) = d u_{\bar{k}ji}$ where $d$ is always positive (as decay rate), $\bar{u}_{\bar{k}ji}$ will be in the positive half-space of $u_{\bar{k}ji}$ if $b+c = b_{\bar{k}ji}u_{00\bar{k}} + U_{\bar{k}ji} + c_{\bar{k}ji} \sum_{l \in \mathcal{N}} w_{l\bar{k}i} u_{l\bar{k}i} > 0$. The stability of $\bar{u}_{\bar{k}ji}$ can be verified similarly to the way it was done in a) and the equilibrium can be found as
\begin{equation*}
\bar{u}_{\bar{k}ji} = \frac{c_{\bar{k}ji}}{d_{\bar{k}ji}} \sum_{l \in \mathcal{N}} w_{l\bar{k}i} u_{l\bar{k}i} + \frac{b_{\bar{k}ji}}{d_{\bar{k}ji}} u_{00\bar{k}} + \frac{U_{\bar{k}ji}}{d_{\bar{k}ji}}
\end{equation*}

When $-ba < 0 \Rightarrow b_{\bar{k}ji}u_{00\bar{k}} r_{\bar{k}ji} > 0$, we obtain cases c), d), e), and f) according to the number of $f=g$ intersections (0, 1, or 2) that we have. In order to derive conditions to distinguish among these cases, we need to compare the position of the points of $g$ that have the same slope as $f$. This means that we have to solve
\begin{align*}
& d = g'(u_{\bar{k}ji}^{1/2}) = (\frac{b u_{\bar{k}ji}^{1/2}}{a + u_{\bar{k}ji}^{1/2}} + c)' = \frac{b a}{(a + u_{\bar{k}ji}^{1/2})^2}\\
& \Rightarrow u_{\bar{k}ji}^{1/2} = -a \pm \sqrt{\frac{ba}{d}} = -r_{\bar{k}ji} \pm \sqrt{\frac{b_{\bar{k}ji} u_{00\bar{k}}}{d_{\bar{k}ji}} r_{\bar{k}ji}}
\end{align*}
where we always have $u_{\bar{k}ji}^{1/2} \in \mathbb{R}$ and $u_{\bar{k}ji}^{2}<u_{\bar{k}ji}^{1}$ since we already assumed $b_{\bar{k}ji}u_{00\bar{k}} r_{\bar{k}ji} > 0$ and $d_{\bar{k}ji} > 0$. As a result we calculate
\begin{align*}
& f(u_{\bar{k}ji}^{1/2}) = d_{\bar{k}ji} u_{\bar{k}ji}^{1/2} = -d_{\bar{k}ji} r_{\bar{k}ji} \pm \sqrt{b_{\bar{k}ji} d_{\bar{k}ji} u_{00\bar{k}} r_{\bar{k}ji}}\\
& g(u_{\bar{k}ji}^{1/2}) = \frac{b (-a \pm \sqrt{\frac{ba}{d}})}{a - a \pm \sqrt{\frac{ba}{d}}} + c = \frac{\mp ab + b\sqrt{\frac{ba}{d}}}{\sqrt{\frac{ba}{d}}} + c = \mp \sqrt{abd} + b + c=\\
& = \mp \sqrt{b_{\bar{k}ji} d_{\bar{k}ji} u_{00\bar{k}} r_{\bar{k}ji}} + b_{\bar{k}ji} u_{00\bar{k}} + U_{\bar{k}ji} + c_{\bar{k}ji} \sum_{l \in \mathcal{N}} w_{l\bar{k}i} u_{l\bar{k}i}
\end{align*}
Thus, if $f(u_{\bar{k}ji}^{2}) < g(u_{\bar{k}ji}^{2})$ and $f(u_{\bar{k}ji}^{1}) > g(u_{\bar{k}ji}^{1})$ we have case c) with no intersection and the trajectories converging towards the singularity at $-a$. If $f(u_{\bar{k}ji}^{2}) = g(u_{\bar{k}ji}^{2})$ or $f(u_{\bar{k}ji}^{1}) = g(u_{\bar{k}ji}^{1})$, we have case f) with one semistable equilibrium point to the left or to the right of $-a$ with trajectories slowing down and stopping at the point and then accelerating away from it and towards $-a$. Finally, if $f(u_{\bar{k}ji}^{1}) > g(u_{\bar{k}ji}^{1})$ and $f(u_{\bar{k}ji}^{2}) > g(u_{\bar{k}ji}^{2})$ we have case d) and if $f(u_{\bar{k}ji}^{1}) < g(u_{\bar{k}ji}^{1})$ and $f(u_{\bar{k}ji}^{2}) < g(u_{\bar{k}ji}^{2})$ we have case e). In the case d) we have one stable and one unstable equilibrium point $\hat{u}_{\bar{k}ji}$ and $\check{u}_{\bar{k}ji}$ to the left of the singularity with basin of attraction of the stable point being $(-\infty,\check{u}_{\bar{k}ji})$. In the case e), $\hat{u}_{\bar{k}ji}$ is unstable and $\check{u}_{\bar{k}ji}$ stable, both are to the right of the singularity, and the basin of attraction of the stable point is $(\hat{u}_{\bar{k}ji},+\infty)$.

In our use of the model (\ref{mm103}), we want to avoid the singularity surface encountered in the above decomposition as well as the entire dynamics on the negative side of $u_{\bar{k}ji}$ which doesn't make much sense in the phenomenological representation of the context term. Even though the output firing rate must always be positive from its definition and the internal potential is generally allowed to be negative from its derivation, we will require nonnegative internal potential, considering inhibition to play the role of depletion of any positive such. Requiring all $u_{kji}$ to be nonnegative however means that some trajectories of the system are no longer admissible since the autonomous dynamics of the network can drive some units to negative potentials. This wasn't the case for all output firing rate models since $a(u)=(u-\alpha)^+ \geq 0$ and each output firing rate would decay to a positive value. Instead of modifying the dynamics to turn off decrease of potential when approaching zero, we will take the simpler approach of using a resetting condition like
\begin{equation*}
u \leftarrow 0 \text{ if } u < 0 
\end{equation*}
Assuming $u_{kji} \geq 0$ for $k,j,i \in [0;N]$ (with the exception of $u_{000}=1$) implies that
\begin{equation*}
-a = -r_{\bar{k}ji} \leq 0 \qquad
-ba = -b_{\bar{k}ji}u_{00\bar{k}} r_{\bar{k}ji} \leq 0
\end{equation*}
and therefore case a) is no longer possible. More importantly, the singularity is locked on the negative side and not reachable by the constrained dynamics of (\ref{mm103}) since $-a \leq 0$ and if $-a = 0 \Rightarrow -ba = 0$ and we are in case b) which doesn't have a singularity. Being some $\epsilon$ positive distance away from the singularity also allows us to still guarantee the existence and uniqueness of solutions in the positive region. In addition, we are interested in obtaining nontrivial stability in the positive region. Since $-a \leq 0$ cases c) and d) reduce to trivial stability where $u_{\bar{k}ji}$ converge to zero. The same behavior is true for case b) with $b + c \leq 0$, case e) with nonpositive stable equilibrium, and case e) with $f(u_{\bar{k}ji}^{2}) = g(u_{\bar{k}ji}^{2})$. Somewhat more interesting behavior appears in e) if $f(u_{\bar{k}ji}^{1}) = g(u_{\bar{k}ji}^{1})$ and the semistable equilibrium point is in the positive region ($u_{\bar{k}ji}^{1}>0$ since the equilibrium coincides with $u_{\bar{k}ji}^{1}$) where convergence to zero will slow down to some positive value $u_{\bar{k}ji}^{1}$. In case b) with $b + c > 0$ we already concluded the existence of a strictly positive equilibrium which we also specified above but this is also not too interesting since the context term is reduced to just the input from $u_{00\bar{k}}$. Finally, the most interesting case of e) includes nontrivial stable state in the positive region with an unstable equilibrium that could be pushed back to the negative region if one requires $u_{\bar{k}ji}^{1}=0$ leaving just one stable state in the nonnegative region, fixed at zero and thus adding an unstable state at zero, or even kept positive and thus adding a second stable state at zero.

Notice however that a stable equilibrium on an $n$-dimensional manifold does not say anything about the overall stability of the point or about the existence of one, more or no equilibria in the entire phase space. In many of the previously mentioned results on stability, the main approach is to find a region or a bound with fully decoupled dynamics, thus producing an equilibrium in one dimension that is constant with respect to the rest. Satisfying any equilibrium existence conditions in all dimensions then would imply existence of the equilibrium in the full phase space. We could obtain full instead of partial equilibria in most of the cases above if we make some additional simplifying assumptions like using piecewise constant activation function (e.g. $a(u)=sgn(u)$) or saturating piecewise linear activation function (e.g. $a(u)=(|u+1|-|u-1|)/2$) so that the reduced equations will contain constants $\tilde{c}$ for many unknowns in certain saturation regions. The only more complicated case is
\begin{align*}
& \tau_r \frac{du_{\bar{k}ji}}{dt} = - d_{\bar{k}ji} u_{\bar{k}ji} + c_{\bar{k}ji} \sum_{l=0}^N w_{l\bar{k}i} \tilde{c}_{l\bar{k}i} + b_{\bar{k}ji} \frac{u_{\bar{k}ji} \tilde{c}_{00\bar{k}}}{\sum\limits_{l=0}^N \sum\limits_{m=1}^N u_{\bar{k}lm}} + U_{\bar{k}ji} \leq \\
& \leq - d_{\bar{k}ji} u_{\bar{k}ji} + c_{\bar{k}ji} \sum_{l=0}^N w_{l\bar{k}i} \tilde{c}_{l\bar{k}i} + b_{\bar{k}ji} |\tilde{c}_{00\bar{k}}| + U_{\bar{k}ji} \\
& \tau_r \frac{du_{\bar{k}ji}}{dt} \geq - d_{\bar{k}ji} u_{\bar{k}ji} + c_{\bar{k}ji} \sum_{l=0}^N w_{l\bar{k}i} \tilde{c}_{l\bar{k}i} - b_{\bar{k}ji} |\tilde{c}_{00\bar{k}}| + U_{\bar{k}ji}
\end{align*}
which is possible assuming nonnegative potentials and implies conclusions similar to the ones in \cite{ZengWang2006, ChengLinShih2006, WangLuChen2010}. However, we are not interested in full equilibria. Partial equilibria represent convergence only in some subset of the unknowns, which is reasonable when part of the network has to sustain oscillations while another part stabilizes or when two relatively independent parts of the input are recalled separately (e.g. object segmentation). We might also add upper bound as a resetting condition although we already have sufficient conditions on boundedness from \cite{ZhangYiYu2008} which could be satisfied on the nonconverging part of the network.

We obtained some additional stability conditions for (\ref{mm103}) from many alternative methods like requirements on the trace and determinant signs of the Jacobian, bounding the real parts of the eigenvalues in the negative half-plane using Gershgorin disks, and trying some standard energy functionals. In most of these cases the conditions weren't easy to interpret or practical enough. Consider for instance the energy functional (\ref{fr_ef}) used for the additive model (\ref{fr_rn_frn2n}). For the expanded graph interpretation of (\ref{mm103}) with $\chi(u_{k01},\dots,u_{kNN})=1$, it can be extended to
\begin{equation} \label{mm_ef}
E(\mathbf{u}) = \sum_{i,j,k=0}^N [-\frac{1}{2} \sum_{l,m,n=0}^N a(u_{kji}) w^{kji}_{lmn} a(u_{lmn}) - U_{kji} a(u_{kji})  - \hat{E}_{kji} + \tilde{E}_{kji}]
\end{equation}
where the term $\tilde{E}$ is one of the two options for (\ref{fr_ef}), for instance
\begin{equation*}
\tilde{E}_{kji} = d_{kji} \int_0^{u_{kji}} s a'(s) ds
\end{equation*}
and the new term $\hat{E}$ is such that
\begin{equation} \label{mm_ef_extra}
\hat{E}_{kji} = b_{kji} a(u_{00k}) \int_0^{u_{kji}} \frac{s a'(s)}{s + r_{kji}} ds
\end{equation}
Under the conditions of integrability of (\ref{mm_ef_extra}), special symmetry of the weight tensor, and proper boundedness of the functional, we can see that
\begin{align*}
& \frac{dE(\mathbf{u})}{dt} = \sum_{i,j,k=0}^N \frac{dE}{du_{kji}} \frac{du_{kji}}{dt} =\\
& = \sum_{i,j,k=0}^N [(- \sum_{l,m,n=0}^N w^{kji}_{lmn} a(u_{lmn}) - U_{kji} - \dots + d_{kji} u_{kji}) \frac{da(u_{kji})}{du_{kji}}] \frac{du_{kji}}{dt} =\\
& = - \sum_{i,j,k=0}^N a'(u_{kji}) (\frac{du_{kji}}{dt})^2 \leq 0 \text{ if } a'(u_i) \geq 0
\end{align*}
where we can only perform the second passage if we assume that all extra terms (in dots) resulting from the differentiation and the multiple additional dependencies of the integral sum up to zero. Additional impractical requirement from this is the symmetry of the 6-order tensor which doesn't map back to admissible metanetwork in the sense of expanded graph.

Using the more natural but unbounded activation like before on the decay term would yield the continuous tensor form of (\ref{mm4}). The differences in the superthreshold region $\{u_{kji} > \alpha\ = 0\}$ are only that we have assumed a special case $d_{kji}=1$. In the subthreshold part $\{u_{kji} \leq \alpha\ = 0\}$ we have instability where $u_{kji}$ acts like an integrator. However, the accumulated potential does not take effect on output since we have to substract it as $\alpha$. A discontinuous version of the previous activation is therefore more reasonable and also differs in the superthreshold region but requires the use of Filippov theory for discontinuous flows.

\subsubsection{Network motif study} \label{network_motif_study}

The feedforward, reduced lateral, reduced excitatory-inhibitory, and reduced and fully recurrent networks (\ref{mm5}, \ref{mm11}, \ref{mm12}, \ref{mm13}, \ref{mm100}) can be explicitly stated if we use a set identifier function like
\begin{equation} \label{set_indicator_function}
\chi(c,u) = \theta(c) \theta(u) = \mathbf{1}_{\{c > \alpha\}} \mathbf{1}_{\{u > \alpha\}}
\end{equation}
If we exclude unit subthreshold cases where we have unstable but integrator behavior with requirements like $\alpha=0$ and $u_{00k}>0, k \in [1;N]$, we will obtain continuous activation functions in all equations. Furthermore, if we confine the initial conditions and trajectories to the intersection of the nonnegative region $u_{kji} \geq 0, k,j,i \in [0;N]$ and the nonbroadcasting region $\bar{h}(C_i)>0$ where $h$ is defined as (\ref{e_ext}) in accordance with the choice of model, the context term will also simplify to a continuous version of the original. Finally, in the following section we will assume nonnegative external input $U_k \ge 0$ although many of the conclusions easily generalize to cases with negative external input as well. Such reductions are meaningful in engineering applications in the next section and exclude some trivial subthreshold behaviors so that we can concentrate on the most complex form of the context term. The simplest explicit form of (\ref{mm5}) can then be written as
\begin{align} \label{mm5_is}
\begin{dcases}
& \tau_r \frac{du^i}{dt} = - \theta(u^i) u^i + \sum_{j=1}^N w_{j\ }^{\ i} \theta(u_{j\ }^{\ i}) \theta(u_j) u_{j\ }^{\ i} = - u^i + \sum_{j=1}^N w_{j\ }^{\ i} \theta(u_{j\ }^{\ i}) u_{j\ }^{\ i}\\
& \tau_r \frac{du_{j\ }^{\ i}}{dt} = - \theta(u_{j\ }^{\ i}) u_{j\ }^{\ i} + \sum_{\substack{k = 1 \\ k \neq j}}^N w_{k\ }^{\ ji} \theta(u_{k\ }^{\ ji}) u_{k\ }^{\ ji} + \frac{u_{j\ }^{\ i} u_j}{\sum\limits_{l=1}^M \sum\limits_{\substack{m = 0 \\ m \neq j}}^N u_{j\ }^{\ ml}}\\
& \tau_r \frac{du_{k\ }^{\ ji}}{dt} =  - \theta(u_{k\ }^{\ ji}) \theta(u_k) u_{k\ }^{\ ji} + \frac{u_{k\ }^{\ ji} \theta(u_k) u_k}{\sum\limits_{l=1}^M \sum\limits_{\substack{m = 0 \\ m \neq k}}^N u_{k\ }^{\ ml}} = - \theta(u_{k\ }^{\ ji}) u_{k\ }^{\ ji} + \frac{u_{k\ }^{\ ji} u_k}{\sum\limits_{l=1}^M \sum\limits_{\substack{m = 0 \\ m \neq k}}^N u_{k\ }^{\ ml}}\\
\end{dcases}
\end{align}
where we have Lipschitz continuous flow and ignore any equations for $u_{k\ }^{\ ki}$. Notice that we may remove any remaining identifier function from the very beginning since $\theta(c) = \mathbf{1}_{\{c > \alpha\}} = \mathbf{1}_{\{c > 0\}}$ and $\theta(c) c = c$ for $c \ge 0$. However, we will keep it in the equations for as long as possible in order to accommodate for generalization to cases with $\alpha > 0$.

All of the above considerations could guarantee no blow up of (\ref{mm5_is}) even though the network trajectories are not necessary bounded. The nonnegative external input assumption in the current notation $u_k \geq 0$ and an initial condition $u_{k\ }^{\ ji}(0)$ in the nonnegative region now would imply that $u_{k\ }^{\ ji}(t) \geq 0$ for all $t>0$ and the trajectory will remain in the nonnegative region for all times since the same will follow for $u_{j\ }^{\ i}(t)$ and $u^i(t)$. Assuming also that $u_k \leq 1$ implies $e(u_{k\ }^{\ ji}, u_k) \leq 1$ and $a(u_{k\ }^{\ ji}, u_k) \leq N$. Using these bounds and Gronwall's lemma with $\tau_r=1$, we can also obtain
\begin{align*}
& \frac{du_{k\ }^{\ ji}}{dt} \leq - a(u_{k\ }^{\ ji}, u_k) + 1 = 1 - \chi(u_{k\ }^{\ ji}, u_k) u_{k\ }^{\ ji} \leq 1 \text{ since } 0 \leq \chi(u_{k\ }^{\ ji}, u_k) \leq 1 \Rightarrow\\
&  \Rightarrow u_{k\ }^{\ ji}(t) \leq u_{k\ }^{\ ji}(0) + t
\end{align*}
as an upper bound on the trajectory of each metaconnection,
\begin{align*}
& \frac{du_{j\ }^{\ i}}{dt} \leq \sum_{\substack{k = 1 \\ k \neq j}}^N \chi(u_{k\ }^{\ ji}, u_k) u_{k\ }^{\ ji} + 1 \leq 1 + \sum_{\substack{k = 1 \\ k \neq j}}^N u_{k\ }^{\ ji} = 1 + \sum_{\substack{k = 1 \\ k \neq i}}^N u_{k\ }^{\ ji}(0) +(N-1)t \Rightarrow\\
&  \Rightarrow u_{j\ }^{\ i}(t) \leq u_{j\ }^{\ i}(0) + t(1 + \sum_{\substack{k = 1 \\ k \neq j}}^N u_{k\ }^{\ ji}(0)) + \frac{N-1}{2} t^2
\end{align*}
as an upper bound on the trajectory of each connection, and
\begin{align*}
& \frac{du^i}{dt} \leq \sum_{j=1}^N \chi(u_{j\ }^{\ i}, u_j) u_{j\ }^{\ i} + 0 \leq \sum_{j=1}^N u_{j\ }^{\ i} \Rightarrow\\
& \Rightarrow u^i(t) \leq u^i(0) + t \sum_{j=1}^N u_{j\ }^{\ i}(0) + \frac{t^2}{2}(N + \sum_{j=1}^N \sum_{\substack{k=1 \\ k \neq i}}^N u_{k\ }^{\ ji}(0)) + \frac{N(N-1)}{6} t^3
\end{align*}
as an upper bound on the trajectory of each unit. All of this indicates that we can rule out blow up in finite time.

The restrictions to the unit superthreshold and nonbroadcasting regions that produced (\ref{mm5_is}) also characterize the stability with all the results from the previous section. In particular, for all units since $b_{kji}=0$ we are always in the case b) on figure \ref{geometrical_setting} with partial equilibrium state
\begin{equation*}
\bar{u}_{kji} = \frac{c_{kji}}{d_{kji}} \sum_{l \in \mathcal{N}} w_{lki} u_{lki} + \frac{U_{kji}}{d_{kji}}
\end{equation*}
which is nontrivial if $c_{kji} \sum_{l \in \mathcal{N}} w_{lki} u_{lki} + U_{kji} > 0$. For the connections and metaconnections, we are once again interested in nontrivial stability (single stable state at zero) which we get again in case b) if $u_{klm}=0, l \in [0;N], l \neq j, m \in [1;N], m \neq i$ (considering tensor form) with an equilibrium
\begin{equation*}
\bar{u}_{kji} = \frac{c_{kji}}{d_{kji}} \sum_{l \in \mathcal{N}} w_{lki} u_{lki} + \frac{u_{00k}}{d_{kji}} + \frac{U_{kji}}{d_{kji}}
\end{equation*}
but also in case e) if we satisfy the conditions $f(u_{kji}^{1}) < g(u_{kji}^{1})$ and $f(u_{kji}^{2}) < g(u_{kji}^{2})$ with the additional requirement of nonnegative stable equilibrium and case f) with the conditions $f(u_{kji}^{1}) < g(u_{kji}^{1})$ and $u_{kji}^{1}>0$. Using results from the previous section, the main conditions of case e) can be written as
\begin{align*}
& f(u_{kji}^{1/2}) < g(u_{kji}^{1/2}) \Rightarrow -d_{kji} r_{kji} \pm \sqrt{b_{kji} d_{kji} u_{00k} r_{kji}} <\\
& < \mp \sqrt{b_{kji} d_{kji} u_{00k} r_{kji}} + b_{kji} u_{00k} + U_{kji} + c_{kji} \sum_{l \in \mathcal{N}} w_{lki} u_{lki}
\end{align*}
which after taking $d_{kji}=b_{kji}=1$ because of the previous equations becomes
\begin{align*}
& - r_{kji} \pm 2\sqrt{u_{00k} r_{kji}} - u_{00k} < U_{kji} + c_{kji} \sum_{l \in \mathcal{N}} w_{lki} u_{lki}
\end{align*}
Since the square root is guaranteed to be real and nonnegative (due to our previous assumptions), the inequality with a plus implies the inequality with the minus as well and we can obtain a polynomial
\begin{align*}
& r_{kji} - 2\sqrt{u_{00k} r_{kji}} + u_{00k} > - U_{kji} - c_{kji} \sum_{l \in \mathcal{N}} w_{lki} u_{lki}\\
& \bigg(\sqrt{r_{kji}} - \sqrt{u_{00k}}\bigg)^2 > - U_{kji} - c_{kji} \sum_{l \in \mathcal{N}} w_{lki} u_{lki}
\end{align*}
which is always true if $U_{kji} + c_{kji} \sum_{l \in \mathcal{N}} w_{lki} u_{lki}>0$ or otherwise true if
\begin{align} \label{ns_conds}
& \bigg|\sqrt{r_{kji}} - \sqrt{u_{00k}}\bigg| > \sqrt{- U_{kji} - c_{kji} \sum_{l \in \mathcal{N}} w_{lki} u_{lki}}
\end{align}
Using similar approach, the main condition of case f) can be written as
\begin{equation} \label{nss_conds}
f(u_{kji}^{1}) = g(u_{kji}^{1}) \Rightarrow \bigg|\sqrt{r_{kji}} + \sqrt{u_{00k}} \bigg| = \sqrt{- U_{kji} - c_{kji} \sum_{l \in \mathcal{N}} w_{lki} u_{lki}}
\end{equation}
and the extra condition can be summarized as
\begin{equation} \label{nss_conds_extra}
u_{kji}^{1} > 0 \Rightarrow -r_{kji} + \sqrt{u_{00k} r_{kji}} > 0
\end{equation}
We will use the conditions of the three cases of nontrivial stability and all assumptions preceding them until the end of this section. For this reason we have to recall that $r_{kji}$ is the remainder sum defined in (\ref{r_sum}).

In comparison to the simplest form (\ref{mm5_is}), the more complex form of (\ref{mm13}) can be written in terms of indicator functions and with all assumptions above as
\small
\begin{align} \label{mm13_is}
\begin{dcases}
& \tau_r \frac{du^i}{dt} = - u^i + \sum_{j=1}^N w_{j\ }^{\ i} \theta(u_{j\ }^{\ i}) u_{j\ }^{\ i} + \sum_{l=1}^M w^{li} \theta(u^{li}) u^{li}\\
& \tau_r \frac{du_{j\ }^{\ i}}{dt} = - \theta(u_{j\ }^{\ i}) u_{j\ }^{\ i} + \sum_{\substack{k = 1 \\ k \neq j}}^N w_{k\ }^{\ ji} \theta(u_{k\ }^{\ ji}) u_{k\ }^{\ ji} + \sum_{l=1}^M w^{l\ }_{\ ji} \theta(u^{l\ }_{\ ji}) u^{l\ }_{\ ji} + \frac{u_{j\ }^{\ i} u_j}{\sum\limits_{m=1}^M \sum\limits_{\substack{n = 0 \\ n \neq j}}^N u_{j\ }^{\ nm}}\\
& \tau_r \frac{du^{li}}{dt} =  - \theta(u^{li}) u^{li} + \frac{u^{li} u^l}{\sum\limits_{m=1}^M [u^{lm} + \sum\limits_{\substack{n = 0 \\ n \neq l}}^N u^{l\ }_{\ nm}]}\\
& \tau_r \frac{du^{i\ }_{\ j}}{dt} =  - \theta(u^{i\ }_{\ j}) u^{i\ }_{\ j} + \frac{u^{i\ }_{\ j} u^i}{\sum\limits_{m=1}^M [u^{im} + \sum\limits_{\substack{n = 0 \\ n \neq i}}^N u^{i\ }_{\ nm}]}\\
& \tau_r \frac{du_{k\ }^{\ ji}}{dt} =  - \theta(u_{k\ }^{\ ji}) u_{k\ }^{\ ji} + \frac{u_{k\ }^{\ ji} u_k}{\sum\limits_{m=1}^M \sum\limits_{\substack{n = 0 \\ n \neq k}}^N u_{k\ }^{\ nm}},\ k \neq j\\
& \tau_r \frac{du^{l\ }_{\ ji}}{dt} =  - \theta(u^{l\ }_{\ ji}) u^{l\ }_{\ ji} + \frac{u^{l\ }_{\ ji} u^l}{\sum\limits_{m=1}^M [u^{lm} + \sum\limits_{\substack{n = 0 \\ n \neq l}}^N u^{l\ }_{\ nm}]}\\
& \tau_r \frac{du_j}{dt} = - u_j + \sum_{i=1}^M w^{i\ }_{\ j} \theta(u^{i\ }_{\ j}) u^{i\ }_{\ j} + U_j
\end{dcases}
\end{align}
\normalsize
We will use (\ref{mm5_is}) and (\ref{mm13_is}) as starting points for deriving smaller network motifs involving feedforward and feedback connections.

\begin{figure}[!htbp]
\centering
\includegraphics[width=0.75\textwidth]{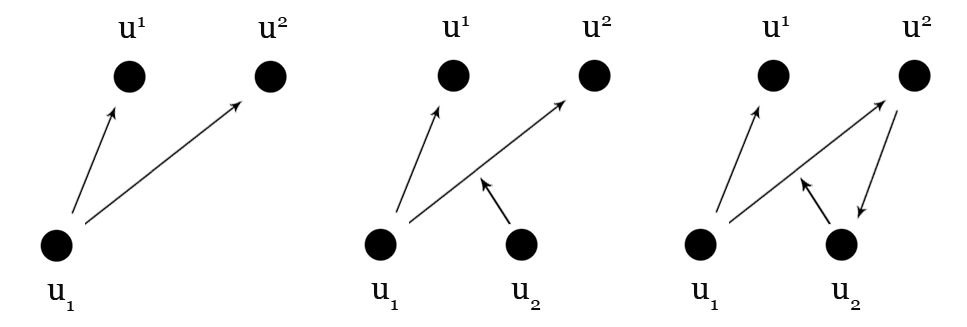}
\caption{Motifs of increasing complexity: a) broadcasting circuit; b) circuit with a metaconnection; c) circuit with a feedback connection;}
\label{circuit2}
\end{figure}

\begin{figure}[!htbp]
\centering
\includegraphics[width=0.5\textwidth]{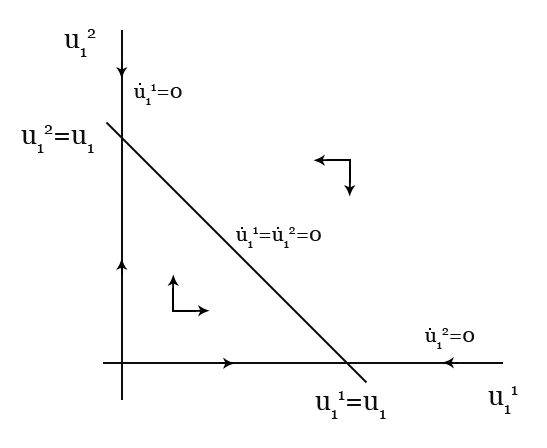}
\caption{Phase portrait of $u_{1\ }^{\ 1}$ and $u_{1\ }^{\ 2}$ from the motif (\ref{mm5_is_v})}
\label{phase_v}
\end{figure}

A study of specific network motifs can help us tie all representations motivating such structure and dynamics with any theoretical conclusions done so far. The simplest circuit with sufficiently complex dynamics is shown in figure \ref{circuit2} case a) and has two connections coming out of one unit. Such a case is a reduced version of (\ref{mm5_is}) with only four equations
\begin{align} \label{mm5_is_v}
\begin{dcases}
& \tau_r \frac{du^1}{dt} = - u^1 + w_{1\ }^{\ 1} \theta(u_{1\ }^{\ 1}) u_{1\ }^{\ 1}\\
& \tau_r \frac{du^2}{dt} = - u^2 + w_{1\ }^{\ 2} \theta(u_{1\ }^{\ 2}) u_{1\ }^{\ 2}\\
& \tau_r \frac{du_{1\ }^{\ 1}}{dt} = - \theta(u_{1\ }^{\ 1}) u_{1\ }^{\ 1} + \frac{u_{1\ }^{\ 1} u_1}{u_{1\ }^{\ 1} + u_{1\ }^{\ 2}}\\
& \tau_r \frac{du_{1\ }^{\ 2}}{dt} = - \theta(u_{1\ }^{\ 2}) u_{1\ }^{\ 2} + \frac{u_{1\ }^{\ 2} u_1}{u_{1\ }^{\ 1} + u_{1\ }^{\ 2}}\\
\end{dcases}
\end{align}
The partial equilibrium states for both units are therefore
\begin{align*}
& \bar{u}^1 = w_{1\ }^{\ 1} u_{1\ }^{\ 1} \text{ if } u_{1\ }^{\ 1}>0 \text{ else } \bar{u}^1 = 0 \Rightarrow \bar{u}^1 = w_{1\ }^{\ 1} u_{1\ }^{\ 1}\\
& \bar{u}^2 = w_{1\ }^{\ 2} u_{1\ }^{\ 2} \text{ if } u_{1\ }^{\ 2}>0 \text{ else } \bar{u}^2 = 0 \Rightarrow \bar{u}^2 = w_{1\ }^{\ 2} u_{1\ }^{\ 2}
\end{align*}
and are stable. The partial equilibrium state for the first connection $u_{1\ }^{\ 1}$ in the simpler case b) of figure \ref{geometrical_setting} is
\begin{equation*}
\theta(\bar{u}_{1\ }^{\ 1}) \bar{u}_{1\ }^{\ 1} = \bar{u}_{1\ }^{\ 1} = u_1 \text{ if } u_{1\ }^{\ 2} = 0 \text{ and } \bar{u}_{1\ }^{\ 1} \neq 0
\end{equation*}
which also conforms to the nonbroadcasting condition $u_{1\ }^{\ 1} + u_{1\ }^{\ 2} > 0$. If the first connection is in the case e), the stability conditions (\ref{ns_conds}) for it are simplified to
\begin{align*}
\begin{dcases}
& \text{ always true if } 0 > 0 \text{ thus not possible}\\
& \big|\sqrt{u_{1\ }^{\ 2}} - \sqrt{u_1}\big| > \sqrt{0 - 0} = 0 \text{ if } 0 \leq 0
\end{dcases}
\Rightarrow \big|\sqrt{u_{1\ }^{\ 2}} - \sqrt{u_1}\big| > 0
\end{align*}
and the corresponding unstable and stable partial equilibrium states are
\begin{align*}
& 0 = - \theta(\bar{u}_{1\ }^{\ 1}) \bar{u}_{1\ }^{\ 1} + \frac{\bar{u}_{1\ }^{\ 1} u_1}{\bar{u}_{1\ }^{\ 1} + u_{1\ }^{\ 2}} \Rightarrow \theta(\bar{u}_{1\ }^{\ 1}) \bar{u}_{1\ }^{\ 1} (\bar{u}_{1\ }^{\ 1} + u_{1\ }^{\ 2}) = \bar{u}_{1\ }^{\ 1} u_1 \Rightarrow\\
& \bar{u}_{1\ }^{\ 1} (\theta(\bar{u}_{1\ }^{\ 1}) \bar{u}_{1\ }^{\ 1} + \theta(\bar{u}_{1\ }^{\ 1}) u_{1\ }^{\ 2} - u_1) = 0 \Rightarrow \bar{u}_{1\ }^{\ 1} = 0 \text{ or } \bar{u}_{1\ }^{\ 1} = u_1 - u_{1\ }^{\ 2}
\end{align*}
The final case $\bar{u}_{1\ }^{\ 1}$ can be in is case f) where the main condition (\ref{nss_conds}) suggests
\begin{equation*}
\big|\sqrt{u_{1\ }^{\ 2}} - \sqrt{u_1}\big| = \pm 0 \Rightarrow u_{1\ }^{\ 2} = u_1
\end{equation*}
and the semistable steady state $u_{kji}^{1} = - u_{1\ }^{\ 2} + \sqrt{u_1 u_{1\ }^{\ 2}} = 0$ can only be trivial since it doesn't satisfy condition (\ref{nss_conds_extra}). We summarize all cases in the analogical conclusions for the second connection $u_{1\ }^{\ 2}$, namely
\begin{align*}
\begin{dcases}
& \bar{u}_{1\ }^{\ 2} = u_1 \text{ if } u_{1\ }^{\ 1} = 0 \text{ and } \bar{u}_{1\ }^{\ 2} \neq 0\\
& \bar{u}_{1\ }^{\ 2} = 0 \text{ or } \bar{u}_{1\ }^{\ 2} = u_1 - u_{1\ }^{\ 1} \text{ if } \big|\sqrt{u_{1\ }^{\ 1}} - \sqrt{u_1}\big| > 0\\
& \bar{u}_{1\ }^{\ 2} = 0 \text{ if } u_{1\ }^{\ 1} = u_1
\end{dcases}
\end{align*}
where $\bar{u}_{1\ }^{\ 2} = u_1$ is the only stable state in the first case, $\bar{u}_{1\ }^{\ 2} = 0$ and $\bar{u}_{1\ }^{\ 2} = u_1 - u_{1\ }^{\ 1}$ are the unstable and stable states in the second case, and $\bar{u}_{1\ }^{\ 2} = 0$ is the semistable state in the third case. Notice that if $u_1 - u_{1\ }^{\ 1} < 0$ in the second case, the stable equilibrium $\bar{u}_{1\ }^{\ 2}$ will no longer be admissible since it will pass on the negative side and we will be left with stable state at zero. All cases of partial equilibria are illustrated on figure \ref{phase_v}. The origin is excluded by the nondistributional condition $u_{1\ }^{\ 1} + u_{1\ }^{\ 2} \neq 0$ and we are left with one partial stable equilibrium $\bar{u}_{1\ }^{\ 2} = u_1$ for fixed $u_{1\ }^{\ 2}=0$ or vice versa. When $|\sqrt{u_{1\ }^{\ 1}} - \sqrt{u_1}| > 0$, we have two partial equilibria for $u_{1\ }^{\ 2}$ along the $u_{1\ }^{\ 1}$ direction with a transcritical bifurcation when $|\sqrt{u_{1\ }^{\ 1}} - \sqrt{u_1}|=0$ and therefore change of their stability after the nonzero equilibrium passes the one at zero.

Some partial nontrivial stable states $(\bar{u}_{1\ }^{\ 1},\bar{u}_{1\ }^{\ 2})$ on a 2-surface can be obtained by solving the algebraic polynomial system
\begin{align*}
\begin{dcases}
& 0 = - \theta(\bar{u}_{1\ }^{\ 1}) \bar{u}_{1\ }^{\ 1} + \frac{\bar{u}_{1\ }^{\ 1} u_1}{\bar{u}_{1\ }^{\ 1} + \bar{u}_{1\ }^{\ 2}}\\
& 0 = - \theta(\bar{u}_{1\ }^{\ 2}) \bar{u}_{1\ }^{\ 2} + \frac{\bar{u}_{1\ }^{\ 2} u_1}{\bar{u}_{1\ }^{\ 1} + \bar{u}_{1\ }^{\ 2}}\\
\end{dcases}
\Rightarrow
\begin{dcases}
& \theta(\bar{u}_{1\ }^{\ 1}) \bar{u}_{1\ }^{\ 1} (\bar{u}_{1\ }^{\ 1} + \bar{u}_{1\ }^{\ 2}) = \bar{u}_{1\ }^{\ 1} u_1\\
& \theta(\bar{u}_{1\ }^{\ 2}) \bar{u}_{1\ }^{\ 2} (\bar{u}_{1\ }^{\ 1} + \bar{u}_{1\ }^{\ 2}) = \bar{u}_{1\ }^{\ 2} u_1\\
\end{dcases}
\end{align*}
and hence
\begin{align*}
\begin{dcases}
& \bar{u}_{1\ }^{\ 1} (\bar{u}_{1\ }^{\ 1} + \bar{u}_{1\ }^{\ 2} - u_1) = 0\\
& \bar{u}_{1\ }^{\ 2} (\bar{u}_{1\ }^{\ 1} + \bar{u}_{1\ }^{\ 2} - u_1) = 0\\
\end{dcases}
\Rightarrow
\begin{dcases}
& \bar{u}_{1\ }^{\ 1} = 0 \text{ or } \bar{u}_{1\ }^{\ 1} + \bar{u}_{1\ }^{\ 2} - u_1 = 0\\
& \bar{u}_{1\ }^{\ 2} = 0 \text{ or } \bar{u}_{1\ }^{\ 1} + \bar{u}_{1\ }^{\ 2} - u_1 = 0
\end{dcases}
\end{align*}
similarly to the way it was done for the partial equilibria on 1-surface. The desired 2-surface partial equilibrium states are then the pairs $(s,u_1-s)$ for $s \in [0,u_1]$. Examining these solutions carefully, we can observe that in the pair $(0,u_1)$, $\bar{u}_{1\ }^{\ 1}$ is in case f) since $|\sqrt{u_{1}} - \sqrt{u_1}|=0$ and $\bar{u}_{1\ }^{\ 2}$ is in case b) while the opposite holds for the pair $(u_1,0)$. All remaining pairs satisfy the conditions (\ref{ns_conds}) and fall into case e) for both components. Plotting the phase plane for ($u_{1\ }^{\ 1}$, $u_{1\ }^{\ 2}$) on figure \ref{phase_v}, we have a line attractor because of the coinciding nullclines $\dot{u}_{1\ }^{\ 1}=\dot{u}_{1\ }^{\ 2}=0$. In the case of three output connections, such a line would become a plane $\bar{u}_{1\ }^{\ 1}+\bar{u}_{1\ }^{\ 2}+\bar{u}_{1\ }^{\ 3}=u_1$ intersecting the three axes at $(u_1,0,0)$, $(0,u_1,0)$, and $(0,0,u_1)$. Finally, the full equilibria are $(\bar{u}_{1\ }^{\ 1}, \bar{u}_{1\ }^{\ 2}, \bar{u}^1, \bar{u}^2) = (s, u_1-s, w_{1\ }^{\ 1} s, w_{1\ }^{\ 2}(u_1-s))$ and lie on an attractor line in a 4-dimensional space. Conclusions on stability of the full equilibria are possible due to the continuous dependence of the partial 1-surface equilibria of all unknowns with respect to the two output connections. For instance, since $u_{1\ }^{\ 1} \rightarrow \bar{u}_{1\ }^{\ 1}$ and $u^1 \rightarrow \bar{u}^1 = w_{1\ }^{\ 1} u_{1\ }^{\ 1}$ in some neighborhood, we have that $(u_{1\ }^{\ 1}, u^1) \rightarrow (\bar{u}_{1\ }^{\ 1}, w_{1\ }^{\ 1} \bar{u}_{1\ }^{\ 1})$.

The interpretation if this motif coincides with the intuition behind it - the system will settle down to a steady state when the potential is properly distributed among the two outputs and any such configuration will remain constant. At the same time if a metaconnection is depleted to zero, it will be sustained in this way unless all remaining output connections settle to zero as well and we obtain broadcasting. Finally, if a connection is depleted to zero it will be sustained in this way unless there is broadcasting or it is forced back into the positive region through excitation by its input metaconnections.

\begin{figure}[!htbp]
\centering
\includegraphics[width=0.8\textwidth]{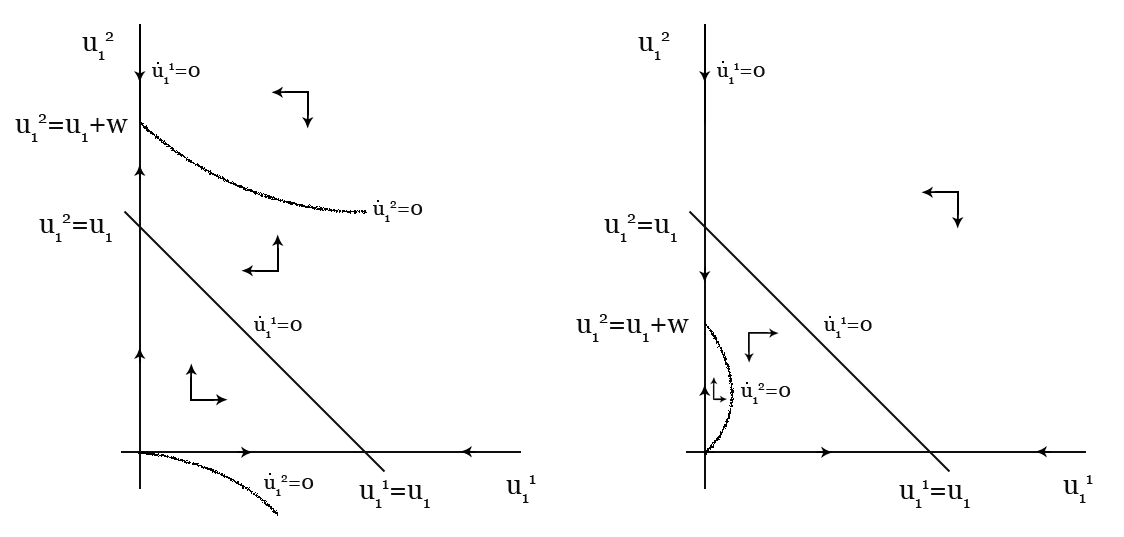}
\caption{Phase portrait of $u_{1\ }^{\ 1}$ and $u_{1\ }^{\ 2}$ from the motif (\ref{mm5_is_k}) where $w = w_{2\ }^{\ 12} \theta(u_{2\ }^{\ 12}) u_{2\ }^{\ 12}$, $u_1 \geq 0$, and: a) $w_{2\ }^{\ 12}>0$; b) $w_{2\ }^{\ 12}<0$; }
\label{phase_k}
\end{figure}

Adding metaconnections, case b) of figure \ref{circuit2} is another reduced version of (\ref{mm5_is}) with five equations and one additional equation to the previous motif (\ref{mm5_is_v}) as
\begin{align} \label{mm5_is_k}
\begin{dcases}
& \tau_r \frac{du^1}{dt} = - u^1 + w_{1\ }^{\ 1} \theta(u_{1\ }^{\ 1}) u_{1\ }^{\ 1}\\
& \tau_r \frac{du^2}{dt} = - u^2 + w_{1\ }^{\ 2} \theta(u_{1\ }^{\ 2}) u_{1\ }^{\ 2}\\
& \tau_r \frac{du_{1\ }^{\ 1}}{dt} = - \theta(u_{1\ }^{\ 1}) u_{1\ }^{\ 1} + \frac{u_{1\ }^{\ 1} u_1}{u_{1\ }^{\ 1} + u_{1\ }^{\ 2}}\\
& \tau_r \frac{du_{1\ }^{\ 2}}{dt} = - \theta(u_{1\ }^{\ 2}) u_{1\ }^{\ 2} + \frac{u_{1\ }^{\ 2} u_1}{u_{1\ }^{\ 1} + u_{1\ }^{\ 2}} + w_{2\ }^{\ 12} \theta(u_{2\ }^{\ 12}) u_{2\ }^{\ 12}\\
& \tau_r \frac{du_{2\ }^{\ 12}}{dt} = - \theta(u_{2\ }^{\ 12}) u_{2\ }^{\ 12} + \frac{u_{2\ }^{\ 12} u_2}{u_{2\ }^{\ 12}} = - \theta(u_{2\ }^{\ 12}) u_{2\ }^{\ 12} + u_2
\end{dcases}
\end{align}
The partial equilibrium states for the two units as well as the metaconnection are obtained in the same way as before, i.e. case b) of figure \ref{geometrical_setting}
\begin{equation*}
\bar{u}^1 = w_{1\ }^{\ 1} u_{1\ }^{\ 1} \qquad \bar{u}^2 = w_{1\ }^{\ 2} u_{1\ }^{\ 2} \qquad \bar{u}_{2\ }^{\ 12} = u_2
\end{equation*}
The conclusions on the 1-surface equilibria of $u_{1\ }^{\ 1}$ are the same as before
\begin{align*}
\begin{dcases}
& \bar{u}_{1\ }^{\ 1} = u_1 \text{ if } u_{1\ }^{\ 2} = 0 \text{ and } \bar{u}_{1\ }^{\ 1} \neq 0\\
& \bar{u}_{1\ }^{\ 1} = 0 \text{ or } \bar{u}_{1\ }^{\ 1} = u_1 - u_{1\ }^{\ 2} \text{ if } \big|\sqrt{u_{1\ }^{\ 2}} - \sqrt{u_1}\big| > 0\\
& \bar{u}_{1\ }^{\ 1} = 0 \text{ if } u_{1\ }^{\ 2} = u_1
\end{dcases}
\end{align*}
For $u_{1\ }^{\ 2}$ we now have 1-surface equilibria given implicitly as
\begin{align*}
& (\theta(\bar{u}_{1\ }^{\ 2}) \bar{u}_{1\ }^{\ 2} - w_{2\ }^{\ 12} \theta(u_{2\ }^{\ 12}) u_{2\ }^{\ 12}) (u_{1\ }^{\ 1} + \bar{u}_{1\ }^{\ 2}) = \bar{u}_{1\ }^{\ 2} u_1 \Rightarrow\\
& \theta(\bar{u}_{1\ }^{\ 2}) (\bar{u}_{1\ }^{\ 2})^2 - (w_{2\ }^{\ 12} \theta(u_{2\ }^{\ 12}) u_{2\ }^{\ 12} - \theta(\bar{u}_{1\ }^{\ 2}) u_{1\ }^{\ 1} + u_1) \bar{u}_{1\ }^{\ 2} - w_{2\ }^{\ 12} \theta(u_{2\ }^{\ 12}) u_{2\ }^{\ 12} u_{1\ }^{\ 1} = 0
\end{align*}
and with modified conditions (\ref{ns_conds}, \ref{nss_conds}) that split into two cases
\begin{align*}
\begin{dcases}
& \bar{u}_{1\ }^{\ 2} = u_1 + w_{2\ }^{\ 12} \theta(u_{2\ }^{\ 12}) u_{2\ }^{\ 12} \text{ if } u_{1\ }^{\ 1} = 0 \text{ and } \bar{u}_{1\ }^{\ 2} \neq 0\\
& \bar{u}_{1\ }^{\ 2} = u_{1/2} \text{ if } \big|\sqrt{u_{1\ }^{\ 1}} - \sqrt{u_1}\big| > \sqrt{-w_{2\ }^{\ 12} \theta(u_{2\ }^{\ 12}) u_{2\ }^{\ 12}}\\
& \bar{u}_{1\ }^{\ 2} = -u_{1\ }^{\ 1} + \sqrt{u_1 u_{1\ }^{\ 1}} \text{ if } \big|\sqrt{u_{1\ }^{\ 1}} - \sqrt{u_1}\big| = \sqrt{-w_{2\ }^{\ 12} \theta(u_{2\ }^{\ 12}) u_{2\ }^{\ 12}}
\end{dcases}
\end{align*}
for $-w_{2\ }^{\ 12} \theta(u_{2\ }^{\ 12}) u_{2\ }^{\ 12} \geq 0 \Rightarrow w_{2\ }^{\ 12} \leq 0$ and
\begin{align*}
\begin{dcases}
& \bar{u}_{1\ }^{\ 2} = u_1 + w_{2\ }^{\ 12} \theta(u_{2\ }^{\ 12}) u_{2\ }^{\ 12} \text{ if } u_{1\ }^{\ 1} = 0 \text{ and } \bar{u}_{1\ }^{\ 2} \neq 0\\
& \bar{u}_{1\ }^{\ 2} = u_{1/2} \text{ otherwise }
\end{dcases}
\end{align*}
for $-w_{2\ }^{\ 12} \theta(u_{2\ }^{\ 12}) u_{2\ }^{\ 12} < 0 \Rightarrow w_{2\ }^{\ 12} u_{2\ }^{\ 12} > 0$. Here $u_{1/2}$ are the two roots of the polynomial in $\bar{u}_{1\ }^{\ 2}$. For a positive weight, their discriminant will always be positive so they will exist and remain distinct, while for a negative weight the discriminant will become zero and ultimately negative so the two roots will merge into one semistable point $\bar{u}_{1\ }^{\ 1} = -u_{1\ }^{\ 2} + \sqrt{u_1 u_{1\ }^{\ 2}}$ and disappear before reappearing when the conditions hold again (due to the absolute values). The second root is the unstable partial equilibrium and is no longer the $u_{1\ }^{\ 1}$ axis (becoming positive or negative for $u_{1\ }^{\ 1}>0$ depending on the weight) and the the first root is the stable partial equilibrium. If $w_{2\ }^{\ 12} u_{2\ }^{\ 12} < -u_1$, the stable root will also be on the negative side and we will have no nontrivial stability. Also, no stability of any kind is asserted in the range $\big|\sqrt{u_{1\ }^{\ 2}} - \sqrt{u_1}\big| < \sqrt{-w_{2\ }^{\ 12}}$ when $w_{2\ }^{\ 12} < 0$. All cases of partial equilibria are illustrated on figure \ref{phase_k}.

If we consider 2-surface steady states, we need to solve the polynomial system
\begin{align*}
\begin{dcases}
& \theta(\bar{u}_{1\ }^{\ 1}) \bar{u}_{1\ }^{\ 1} (\bar{u}_{1\ }^{\ 1} + \bar{u}_{1\ }^{\ 2}) = \bar{u}_{1\ }^{\ 1} u_1\\
& (\theta(\bar{u}_{1\ }^{\ 2}) \bar{u}_{1\ }^{\ 2} - w_{2\ }^{\ 12} \theta(u_{2\ }^{\ 12}) u_{2\ }^{\ 12}) (\bar{u}_{1\ }^{\ 1} + \bar{u}_{1\ }^{\ 2}) = \bar{u}_{1\ }^{\ 2} u_1\\
\end{dcases}
\end{align*}
where each component must satisfy the conditions of one of the three cases before. If $\theta(\bar{u}_{2\ }^{\ 12}) = 0$ we obtain the previous solution namely the one of (\ref{mm5_is_v}). Otherwise, we get
\begin{align*}
\begin{dcases}
& \theta(\bar{u}_{1\ }^{\ 1}) \bar{u}_{1\ }^{\ 1} (\bar{u}_{1\ }^{\ 1} + \bar{u}_{1\ }^{\ 2}) = \bar{u}_{1\ }^{\ 1} u_1 \Rightarrow \bar{u}_{1\ }^{\ 1} = 0 \text{ or } \bar{u}_{1\ }^{\ 1} + \bar{u}_{1\ }^{\ 2} = u_1\\
& (\theta(\bar{u}_{1\ }^{\ 2}) \bar{u}_{1\ }^{\ 2} - w_{2\ }^{\ 12} \theta(u_{2\ }^{\ 12}) u_{2\ }^{\ 12}) (\bar{u}_{1\ }^{\ 1} + \bar{u}_{1\ }^{\ 2}) = \bar{u}_{1\ }^{\ 2} u_1\\
\end{dcases}
\end{align*}
Hence, the second equation when $\bar{u}_{1\ }^{\ 1} = 0$ is
\begin{align*}
& (\theta(\bar{u}_{1\ }^{\ 2}) \bar{u}_{1\ }^{\ 2} - w_{2\ }^{\ 12} \theta(u_{2\ }^{\ 12}) u_{2\ }^{\ 12})(0 + \bar{u}_{1\ }^{\ 2}) = \bar{u}_{1\ }^{\ 2} u_1\\
& \bar{u}_{1\ }^{\ 2} (\theta(\bar{u}_{1\ }^{\ 2}) \bar{u}_{1\ }^{\ 2} - w_{2\ }^{\ 12} \theta(u_{2\ }^{\ 12}) u_{2\ }^{\ 12} - u_1) = 0\\
& \Rightarrow \bar{u}_{1\ }^{\ 2} = 0 \text{ or } \bar{u}_{1\ }^{\ 2} = u_1 + w_{2\ }^{\ 12} \theta(u_{2\ }^{\ 12}) u_{2\ }^{\ 12}
\end{align*}
and the partial 2-surface equilibria are $(0, 0)$ and $(0, u_1 + w_{2\ }^{\ 12} \theta(u_{2\ }^{\ 12}) u_{2\ }^{\ 12})$ where we exclude $(0, 0)$ due to the nondistributional condition $u_{1\ }^{\ 1} + u_{1\ }^{\ 2} \neq 0$. Alternatively, the second equation when $\bar{u}_{1\ }^{\ 1} + \bar{u}_{1\ }^{\ 2} = u_1$ is
\begin{align*}
& (\theta(\bar{u}_{1\ }^{\ 2}) \bar{u}_{1\ }^{\ 2} - w_{2\ }^{\ 12} \theta(u_{2\ }^{\ 12}) u_{2\ }^{\ 12}) u_1 = \bar{u}_{1\ }^{\ 2} u_1\\
& \bar{u}_{1\ }^{\ 2} - w_{2\ }^{\ 12} \theta(u_{2\ }^{\ 12}) u_{2\ }^{\ 12} = \bar{u}_{1\ }^{\ 2}\\
& \Rightarrow \forall \bar{u}_{1\ }^{\ 2} \text{ if } - w_{2\ }^{\ 12} \theta(u_{2\ }^{\ 12}) u_{2\ }^{\ 12} = 0
\end{align*}
with partial equilibria $(u_1-a, a), a \in [0, u_1]$ which exist only if $u_{2\ }^{\ 12}=0$ or $w_{2\ }^{\ 12}=0$ and the metaconnection does not have any effect. In such a case, we will obtain the partial 1-surface equilibria of (\ref{mm5_is_v}) and the previous 2-surface equilibrium $(0, u_1 + w_{2\ }^{\ 12} \theta(u_{2\ }^{\ 12}) u_{2\ }^{\ 12})$ will reduce to $(0, u_1)=(u_1-a, a)$ for $a=u_1$. If the metaconnection effect is turned on, the line attractor is replaced by a point attractor in figure \ref{phase_k} a) or by an unstable equilibrium point in b) depending on the sign of the weight $w_{2\ }^{\ 12}$. Notice that the points $(0,0)$ and $(0,u_1 + w_{2\ }^{\ 12} \theta(u_{2\ }^{\ 12}) u_{2\ }^{\ 12})$ are always points on the locus of the roots since they solve the equation for $\bar{u}_{1\ }^{\ 2}$ which is a polynomial of degree 2 and therefore could have at most two such intersections with the $u_{1\ }^{\ 2}$ axis. We could use this to argue that if $w_{2\ }^{\ 12} \theta(u_{2\ }^{\ 12}) u_{2\ }^{\ 12} = w < -u_1$, the $(0,u_1 + w)$ point would lie on the negative side of the axis (now shown on the figure) and the polynomial nullcline would only intersect the nonnegative region at $(0,0)$. Alternatively, we could obtain the same as a condition on the derivative of the nullcline at $(0,0)$ as
\begin{align*}
& (\theta(\bar{u}_{1\ }^{\ 2}) \bar{u}_{1\ }^{\ 2} - w) (\bar{u}_{1\ }^{\ 1} + \bar{u}_{1\ }^{\ 2}) = \bar{u}_{1\ }^{\ 2} u_1 \Rightarrow \bar{u}_{1\ }^{\ 1}(\bar{u}_{1\ }^{\ 2}) = - \bar{u}_{1\ }^{\ 2} + \frac{\bar{u}_{1\ }^{\ 2} u_1}{\theta(\bar{u}_{1\ }^{\ 2}) \bar{u}_{1\ }^{\ 2} - w}\\
& \frac{d\bar{u}_{1\ }^{\ 1}}{d\bar{u}_{1\ }^{\ 2}} = - 1 - \frac{w u_1}{(\theta(\bar{u}_{1\ }^{\ 2}) \bar{u}_{1\ }^{\ 2} - w)^2} \Rightarrow \frac{d\bar{u}_{1\ }^{\ 1}}{d\bar{u}_{1\ }^{\ 2}}(0) = - 1 - \frac{w u_1}{(0 - w)^2} > 0 \Rightarrow w < - u_1
\end{align*}
The asymptotes of the hyperbola formed by the $u_{1\ }^{\ 2}$ nullcines are $u_{1\ }^{\ 1} + u_{1\ }^{\ 2} = u_1$ and $u_{1\ }^{\ 2} = w$ and therefore the second nullcline of $u_{1\ }^{\ 1}$ will coincide with the asymptote $u_{1\ }^{\ 1} + u_{1\ }^{\ 2} = u_1$ and never intersect these nullclines to form any other equilibria.

If we consider the points on a 3-surface with partial equilibria $(\bar{u}_{1\ }^{\ 1}, \bar{u}_{1\ }^{\ 2}, \bar{u}_{2\ }^{\ 12})$ we obtain the system
\begin{align*}
\begin{dcases}
& \theta(\bar{u}_{1\ }^{\ 1}) \bar{u}_{1\ }^{\ 1} (\bar{u}_{1\ }^{\ 1} + \bar{u}_{1\ }^{\ 2}) = \bar{u}_{1\ }^{\ 1} u_1\\
& (\theta(\bar{u}_{1\ }^{\ 2}) \bar{u}_{1\ }^{\ 2} - w_{2\ }^{\ 12} \theta(\bar{u}_{2\ }^{\ 12}) \bar{u}_{2\ }^{\ 12}) (\bar{u}_{1\ }^{\ 1} + \bar{u}_{1\ }^{\ 2}) = \bar{u}_{1\ }^{\ 2} u_1\\
& \bar{u}_{2\ }^{\ 12} = u_2
\end{dcases}
\end{align*}
which is solved analogically but with equilibria $(0, u_1 + w_{2\ }^{\ 12} \bar{u}_{2\ }^{\ 12}, \bar{u}_{2\ }^{\ 12}) = (0, u_1 + w_{2\ }^{\ 12} u_2, u_2)$ and $(u_1-a, a, u_2), a \in [0, u_1]$ if $u_2=0$ or $w_{2\ }^{\ 12}=0$. The full equilibria are similar to before but with the extra component $\bar{u}_{2\ }^{\ 12}=u_2$. The interpretation is the same when the metaconnection has no effect on the connection because of zero weight or zero throughput - we have a line attractor and connections that settle when the potential of the unit is fully distributed. Differently, when the metaconnection has positive effect it will bias the distribution towards the configuration where only the aided connection is emitted through and the connection with no metaconnection support will converge to zero. When the metaconnection has negative effect, the aided connection $u_{1\ }^{\ 2}$ will settle to reduced value as long as the other connection $u_{1\ }^{\ 1}$ is fully deactivated. Otherwise, even the slightest positive potential $u_{1\ }^{\ 1}>0$ will bring to the full depletion of $u_{1\ }^{\ 2}$ and $u_{1\ }^{\ 1}=u_1$ as the ultimate winner.

Adding feedback connections, case c) of figure \ref{circuit2} is a reduced version of (\ref{mm13_is}) with eight equations
\begin{align} \label{mm13_is_v}
\begin{dcases}
& \tau_r \frac{du^1}{dt} = - u^1 + w_{1\ }^{\ 1} \theta(u_{1\ }^{\ 1}) u_{1\ }^{\ 1}\\
& \tau_r \frac{du^2}{dt} = - u^2 + w_{1\ }^{\ 2} \theta(u_{1\ }^{\ 2}) u_{1\ }^{\ 2}\\
& \tau_r \frac{du_{1\ }^{\ 1}}{dt} = - \theta(u_{1\ }^{\ 1}) u_{1\ }^{\ 1} + \frac{u_{1\ }^{\ 1} u_1}{u_{1\ }^{\ 1} + u_{1\ }^{\ 2}}\\
& \tau_r \frac{du_{1\ }^{\ 2}}{dt} = - \theta(u_{1\ }^{\ 2}) u_{1\ }^{\ 2} + \frac{u_{1\ }^{\ 2} u_1}{u_{1\ }^{\ 1} + u_{1\ }^{\ 2}} + w_{2\ }^{\ 12} \theta(u_{2\ }^{\ 12}) u_{2\ }^{\ 12}\\
& \tau_r \frac{du^{2\ }_{\ 2}}{dt} = - \theta(u^{2\ }_{\ 2}) u^{2\ }_{\ 2} + \frac{u^{2\ }_{\ 2} u^2}{u^{2\ }_{\ 2}} = - \theta(u^{2\ }_{\ 2}) u^{2\ }_{\ 2} + u^2\\
& \tau_r \frac{du_{2\ }^{\ 12}}{dt} = - \theta(u_{2\ }^{\ 12}) u_{2\ }^{\ 12} + \frac{u_{2\ }^{\ 12} u_2}{u_{2\ }^{\ 12}} = - \theta(u_{2\ }^{\ 12}) u_{2\ }^{\ 12} + u_2\\
& \tau_r \frac{du_1}{dt} = - u_1 + U_1\\
& \tau_r \frac{du_2}{dt} = - u_2 + w^{2\ }_{\ 2} \theta(u^{2\ }_{\ 2}) u^{2\ }_{\ 2} + U_2
\end{dcases}
\end{align}
The partial equilibrium states for all units now are
\begin{align*}
\bar{u}^1 &= w_{1\ }^{\ 1} u_{1\ }^{\ 1} & \bar{u}^2 &= w_{1\ }^{\ 2} u_{1\ }^{\ 2}\\
\bar{u}_1 &= U_1 & \bar{u}_2 &= w^{2\ }_{\ 2} u^{2\ }_{\ 2} + U_2
\end{align*}
and are stable as usual. The single outputs $\bar{u}^{2\ }_{\ 2}$ and $\bar{u}_{2\ }^{\ 12}$ are in the same case like all the units, namely case b) of figure \ref{geometrical_setting} with stable states
\begin{align*}
\bar{u}^{2\ }_{\ 2} = u^2 \qquad \bar{u}_{2\ }^{\ 12} = u_2
\end{align*}
The partial 1-surface and 2-surface equilibria for the competitive connections $u_{1\ }^{\ 1}$ and $u_{1\ }^{\ 2}$ are as in (\ref{mm5_is_k}) but the 3-surface equilibria and any resulting full-space equilibria differ significantly due to the presence of feedback connections and a cycle within the motif. We can see this by considering the full depth of the cycle in order to decouple $u_{1\ }^{\ 2}$ from everything but $u_{1\ }^{\ 1}$.

If we fix $u_{2\ }^{\ 12}$ to $\bar{u}_{2\ }^{\ 12}$ and all its dependencies or simply consider 7-surface stable states we obtain the algebraic polynomial system
\begin{align*}
\begin{dcases}
& \theta(\bar{u}_{1\ }^{\ 1}) \bar{u}_{1\ }^{\ 1} (\bar{u}_{1\ }^{\ 1} + \bar{u}_{1\ }^{\ 2}) = \bar{u}_{1\ }^{\ 1} U_1\\
& (\theta(\bar{u}_{1\ }^{\ 2}) \bar{u}_{1\ }^{\ 2} - w_{2\ }^{\ 12} \theta(\bar{u}_{2\ }^{\ 12}) \bar{u}_{2\ }^{\ 12}) (\bar{u}_{1\ }^{\ 1} + \bar{u}_{1\ }^{\ 2}) = \bar{u}_{1\ }^{\ 2} U_1\\
& \bar{u}_{2\ }^{\ 12} = \bar{u}_2 = w^{2\ }_{\ 2} \bar{u}^{2\ }_{\ 2} + U_2 = w^{2\ }_{\ 2} \bar{u}^2 + U_2 = w^{2\ }_{\ 2} w_{1\ }^{\ 2} \bar{u}_{1\ }^{\ 2} + U_2
\end{dcases}
\end{align*}
which is solved similarly as in (\ref{mm5_is_k}) where the second equation when $\bar{u}_{1\ }^{\ 1} = 0$ is
\begin{align*}
& (\theta(\bar{u}_{1\ }^{\ 2}) \bar{u}_{1\ }^{\ 2} - w_{2\ }^{\ 12} w^{2\ }_{\ 2} w_{1\ }^{\ 2} \bar{u}_{1\ }^{\ 2} - w_{2\ }^{\ 12} U_2)(0 + \bar{u}_{1\ }^{\ 2}) = \bar{u}_{1\ }^{\ 2} U_1\\
& \bar{u}_{1\ }^{\ 2} (\theta(\bar{u}_{1\ }^{\ 2}) \bar{u}_{1\ }^{\ 2} - w_{2\ }^{\ 12} w^{2\ }_{\ 2} w_{1\ }^{\ 2} \bar{u}_{1\ }^{\ 2} - w_{2\ }^{\ 12} U_2 - U_1) = 0\\
& \Rightarrow \bar{u}_{1\ }^{\ 2} = 0 \text{ or } \bar{u}_{1\ }^{\ 2} (1 - w_{2\ }^{\ 12} w^{2\ }_{\ 2} w_{1\ }^{\ 2}) = w_{2\ }^{\ 12} U_2 + U_1
\end{align*}
and the partial equilibria are $(0,0)$ and $(0,(U_1 + w_{2\ }^{\ 12} U_2)/(1-w_{2\ }^{\ 12}w^{2\ }_{\ 2}w_{1\ }^{\ 2})) = (0,\tilde{U}_1)$ (again excluding $(0,0)$ due to $u_{1\ }^{\ 1} + u_{1\ }^{\ 2} \neq 0$). The second equation when $\bar{u}_{1\ }^{\ 1} + \bar{u}_{1\ }^{\ 2} = U_1$ is
\begin{align*}
& (\theta(\bar{u}_{1\ }^{\ 2}) \bar{u}_{1\ }^{\ 2} - w_{2\ }^{\ 12} w^{2\ }_{\ 2} w_{1\ }^{\ 2} \bar{u}_{1\ }^{\ 2} - w_{2\ }^{\ 12} U_2) U_1 = \bar{u}_{1\ }^{\ 2} U_1\\
& \bar{u}_{1\ }^{\ 2} - w_{2\ }^{\ 12} w^{2\ }_{\ 2} w_{1\ }^{\ 2} \bar{u}_{1\ }^{\ 2} - w_{2\ }^{\ 12} U_2 = \bar{u}_{1\ }^{\ 2}\\
& \Rightarrow w_{2\ }^{\ 12} w^{2\ }_{\ 2} w_{1\ }^{\ 2} \bar{u}_{1\ }^{\ 2} = - w_{2\ }^{\ 12} U_2
\end{align*}
with a partial equilibrium $(U_1-s,s), s \in [0,u_1]$ if $w_{2\ }^{\ 12}=0$ or $U_2=w^{2\ }_{\ 2}w_{1\ }^{\ 2}=0$, only a special case thereof $(U_1+U_2/(w^{2\ }_{\ 2} w_{1\ }^{\ 2}), -U_2/(w^{2\ }_{\ 2} w_{1\ }^{\ 2})) = (U_1+\tilde{U}_2, -\tilde{U}_2)$ with $s=-\tilde{U}_2$ if $w_{2\ }^{\ 12} \neq 0$, $w^{2\ }_{\ 2} w_{1\ }^{\ 2} \neq 0$,  and $U_2 \neq 0$ and only a special case thereof $(U_1,0)$ when $w_{2\ }^{\ 12} \neq 0$, $w^{2\ }_{\ 2} w_{1\ }^{\ 2} \neq 0$ and $U_2=0$. If $w_{2\ }^{\ 12}=0$ or $U_2=w^{2\ }_{\ 2}w_{1\ }^{\ 2}=0$ the first equilibrium will reduce to another equilibrium on the line $(0,\tilde{U}_1)=(0,(U_1 + 0)/(1-0))=(0,U_1)$ with $s=U_1$ and we can conclude that generally in this case we have a line attractor in a 7-dimensional space. In the remaining cases we have two equilibrium points $(\bar{u}_{1\ }^{\ 1}, \bar{u}_{1\ }^{\ 2}, \bar{u}^2, \bar{u}_1, \bar{u}_2, \bar{u}^{2\ }_{\ 2}, \bar{u}_{2\ }^{\ 12})$ as
\begin{align*}
& (0, \tilde{U}_1, w_{1\ }^{\ 2} \tilde{U}_1, U_1, w^{2\ }_{\ 2} w_{1\ }^{\ 2} \tilde{U}_1 + U_2, w_{1\ }^{\ 2} \tilde{U}_1, w^{2\ }_{\ 2} w_{1\ }^{\ 2} \tilde{U}_1 + U_2) =\\
& = (0, \tilde{U}_1, w_{1\ }^{\ 2} \tilde{U}_1, U_1, \frac{w^{2\ }_{\ 2} w_{1\ }^{\ 2}  U_1 + U_2}{1-w_{2\ }^{\ 12}w^{2\ }_{\ 2}w_{1\ }^{\ 2}}, w_{1\ }^{\ 2} \tilde{U}_1, \frac{w^{2\ }_{\ 2} w_{1\ }^{\ 2}  U_1 + U_2}{1-w_{2\ }^{\ 12}w^{2\ }_{\ 2}w_{1\ }^{\ 2}})
\end{align*}
and
\begin{align*}
& (U_1+\tilde{U}_2, -\tilde{U}_2, -w_{1\ }^{\ 2} \tilde{U}_2, U_1, -w^{2\ }_{\ 2} w_{1\ }^{\ 2} \tilde{U}_2 + U_2, -w_{1\ }^{\ 2} \tilde{U}_2, -w^{2\ }_{\ 2} w_{1\ }^{\ 2} \tilde{U}_2 + U_2) =\\
& = (U_1+\tilde{U}_2, -\tilde{U}_2, -\frac{U_2}{w^{2\ }_{\ 2}}, U_1, 0, -\frac{U_2}{w^{2\ }_{\ 2}}, 0)
\end{align*}
When we take a cross-section of the 7-space to observe the $(u_{1\ }^{\ 1}, u_{1\ }^{\ 2})$ phase plane we considered earlier, we can associate the first point with the 2-surface equilibrium point $(0,u_1+w)$ and the second with a point on the 2-surface attractor line (since $\bar{u}_{2\ }^{\ 12}=0$).

The full equilibria are then $(\bar{u}_{1\ }^{\ 1}, \bar{u}_{1\ }^{\ 2}, \bar{u}^1, \bar{u}^2, \bar{u}_1, \bar{u}_2, \bar{u}^{2\ }_{\ 2}, \bar{u}_{2\ }^{\ 12})$ with the addition of an extra dimension for $u^1$ which is $0$ for the first attractor point and $w_{1\ }^{\ 1} (U_1+\tilde{U}_2)$ for the second attractor point. In the case of attractor line in the full phase space, the infinitely many equilibrium points are parametrized by $(U_1-s,s,w_{1\ }^{\ 1}(U_1-s),w_{1\ }^{\ 2}s,U_1,w^{2\ }_{\ 2}w_{1\ }^{\ 2}s+U_2,w_{1\ }^{\ 2}s,w^{2\ }_{\ 2}w_{1\ }^{\ 2}s+U_2)$. Excluding the attractor line, the interpretation for the first equilibrium which is a feedback extension of the previous $w_{2\ }^{\ 12} u_{2\ }^{\ 12} \neq 0$ equilibrium of (\ref{mm5_is_k}) demonstrates the role of cycles within the network. If we take $w_{2\ }^{\ 12} w^{2\ }_{\ 2} w_{1\ }^{\ 2} = 1$, the cycle will keep accumulating potential forever and the equilibrium will be located at $\bar{u}_{1\ }^{\ 2}=\infty$. At the same time $\bar{u}_{1\ }^{\ 2}$ can be finite if the weights are less than unity and therefore scale down the accumulated potential over time. Alternatively, if we have inhibitory feedback $w^{2\ }_{\ 2} < 0$ this will scale down $\bar{u}_{1\ }^{\ 2}$ for any choice of positive $w_{2\ }^{\ 12}$ and $w_{1\ }^{\ 2}$. If $w_{2\ }^{\ 12} = 0$, the steady state becomes simply $\bar{u}_{1\ }^{\ 2}=U_1$ and the crucial connection for the previous behavior is removed. If $w^{2\ }_{\ 2} = 0$ or $w_{1\ }^{\ 2} = 0$, the steady state becomes $\bar{u}_{1\ }^{\ 2} = U_1 + w_{2\ }^{\ 12} U_2$ and we only have the effect of the second external input as before (the cycle is interrupted). The second equilibrium is in the nonnegative region and therefore admissible if $U_2/(w^{2\ }_{\ 2} w_{1\ }^{\ 2}) \leq 0$ which implies either a nonpositive weight or nonpositive external input both of which would reduce $\bar{u}_{1\ }^{\ 1}$. The steady state is reached since the feedback connection exactly cancels the external input $U_2$ leaving the two output connections somewhere on a point of the attractor line where this happens. This is possible in particular if we consider inhibitory feedback connection or negative external input $U_2<0$ but not $w_{1\ }^{\ 2}<0$ since the resetting condition will simply set $u^2$ to zero.

\begin{figure}[!htbp]
\centering
\includegraphics[width=0.85\textwidth]{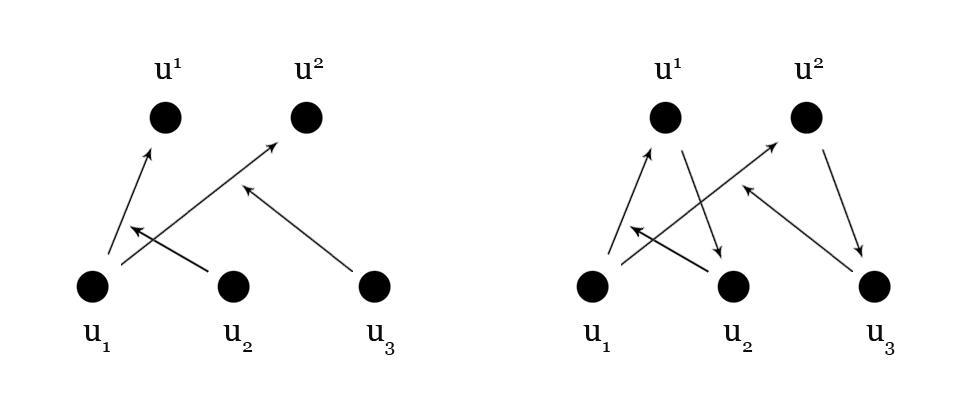}
\caption{Motifs of increasing complexity: d) circuit with competing metaconnections; e) circuit with competing feedback connections;}
\label{circuit3}
\end{figure}

\begin{figure}[!htbp]
\centering
\includegraphics[width=0.8\textwidth]{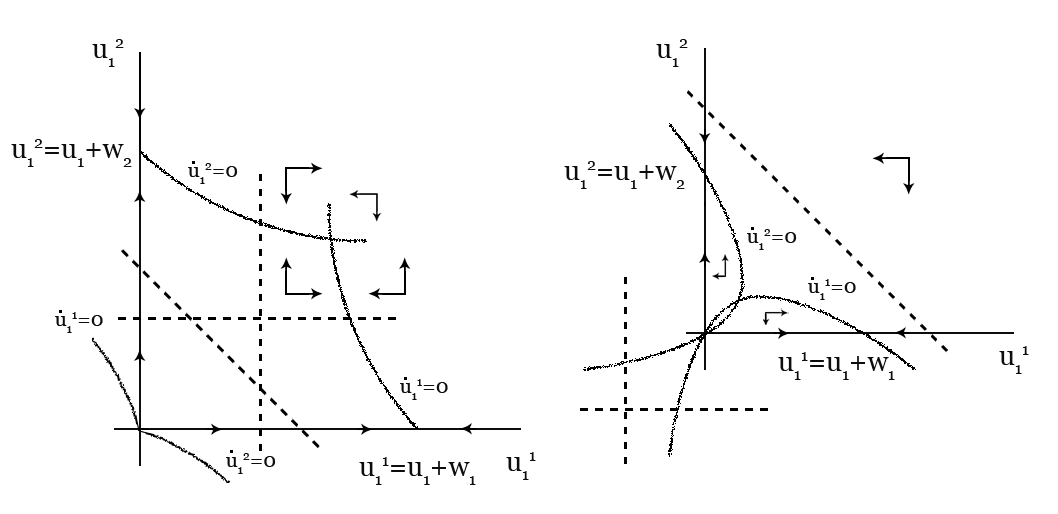}
\caption{Phase portrait of $u_{1\ }^{\ 1}$ and $u_{1\ }^{\ 2}$ from the competitive motif where $u_1 \geq 0$ and: a) $w_{2\ }^{\ 11}>0$ and $w_{3\ }^{\ 12}>0$; b) $w_{2\ }^{\ 11}<0$ and $w_{3\ }^{\ 12}<0$;}
\label{phase_m}
\end{figure}

Extending to cases with competition while still minimizing the added complexity will lead to motifs like the ones shown on figure \ref{circuit3}. A competitive reduced version of (\ref{mm5_is}) adds a second metaconnection to balance the potential among the two outputs from before. The dynamical system definition extends in the intuitive way on top of (\ref{mm5_is_k}) and the 1-surface equilibria of both output connections $u_{1\ }^{\ 1}$ and $u_{1\ }^{\ 2}$ are the same as the ones for $u_{1\ }^{\ 2}$ in (\ref{mm5_is_k}). However the 2-surface equilibria are different and solve the system
\begin{align*}
\begin{dcases}
& (\theta(\bar{u}_{1\ }^{\ 1}) \bar{u}_{1\ }^{\ 1} - w_{2\ }^{\ 11} \theta(u_{2\ }^{\ 11}) u_{2\ }^{\ 11}) (\bar{u}_{1\ }^{\ 1} + \bar{u}_{1\ }^{\ 2}) = \bar{u}_{1\ }^{\ 1} u_1\\
& (\theta(\bar{u}_{1\ }^{\ 2}) \bar{u}_{1\ }^{\ 2} - w_{3\ }^{\ 12} \theta(u_{3\ }^{\ 12}) u_{3\ }^{\ 12}) (\bar{u}_{1\ }^{\ 1} + \bar{u}_{1\ }^{\ 2}) = \bar{u}_{1\ }^{\ 2} u_1
\end{dcases}
\end{align*}
In  the case $w_{2\ }^{\ 11} \theta(u_{2\ }^{\ 11}) u_{2\ }^{\ 11} = 0$ or $w_{3\ }^{\ 12} \theta(u_{3\ }^{\ 12}) u_{3\ }^{\ 12} = 0$ the system and corresponding equilibria are reduced to the previous motifs (\ref{mm5_is_v}, \ref{mm5_is_k}). It remains to solve
\begin{align*}
\begin{dcases}
& (\theta(\bar{u}_{1\ }^{\ 1}) \bar{u}_{1\ }^{\ 1} - w_1) (\bar{u}_{1\ }^{\ 1} + \bar{u}_{1\ }^{\ 2}) = \bar{u}_{1\ }^{\ 1} u_1\\
& (\theta(\bar{u}_{1\ }^{\ 2}) \bar{u}_{1\ }^{\ 2} - w_2) (\bar{u}_{1\ }^{\ 1} + \bar{u}_{1\ }^{\ 2}) = \bar{u}_{1\ }^{\ 2} u_1
\end{dcases}
\end{align*}
for $w_1 = w_{2\ }^{\ 11} \theta(u_{2\ }^{\ 11}) u_{2\ }^{\ 11} \neq 0$ and $w_2 = w_{3\ }^{\ 12} \theta(u_{3\ }^{\ 12}) u_{3\ }^{\ 12} \neq 0$. First we obtain
\begin{align*}
\begin{dcases}
& \bar{u}_{1\ }^{\ 1} + \bar{u}_{1\ }^{\ 2} = \frac{\bar{u}_{1\ }^{\ 1} u_1}{\theta(\bar{u}_{1\ }^{\ 1}) \bar{u}_{1\ }^{\ 1} - w_1}\\
& \frac{(\theta(\bar{u}_{1\ }^{\ 2}) \bar{u}_{1\ }^{\ 2} - w_2) \bar{u}_{1\ }^{\ 1} u_1}{\theta(\bar{u}_{1\ }^{\ 1}) \bar{u}_{1\ }^{\ 1} - w_1} = \bar{u}_{1\ }^{\ 2} u_1
\end{dcases}
\end{align*}
which after canceling any possible terms becomes
\begin{align*}
\begin{dcases}
& \bar{u}_{1\ }^{\ 1} + \frac{w_2 \bar{u}_{1\ }^{\ 1}}{w_1} = \frac{\bar{u}_{1\ }^{\ 1} u_1}{\theta(\bar{u}_{1\ }^{\ 1}) \bar{u}_{1\ }^{\ 1} - w_1}\\
& w_2 \bar{u}_{1\ }^{\ 1} = w_1 \bar{u}_{1\ }^{\ 2}
\end{dcases}
\end{align*}
which has nonnegative solutions $\bar{u}_{1\ }^{\ 1}$ and $\bar{u}_{1\ }^{\ 2}$ and therefore an equilibrium in the nonnegative region only if $w_1$ and $w_2$ have the same sign. Finally, assuming this we get
\begin{align*}
& \theta(\bar{u}_{1\ }^{\ 1}) \bar{u}_{1\ }^{\ 1} - w_1 = \frac{w_1 u_1}{w_1 + w_2} \Rightarrow \bar{u}_{1\ }^{\ 1} = \frac{w_1 (w_1 + w_2)}{w_1 + w_2} + \frac{w_1 u_1}{w_1 + w_2}\\
& \Rightarrow (\bar{u}_{1\ }^{\ 1}, \bar{u}_{1\ }^{\ 2}) = \bigg(\frac{w_1 (u_1 + w_1 + w_2)}{w_1 + w_2}, \frac{w_2 (u_1 + w_1 + w_2)}{w_1 + w_2}\bigg)
\end{align*}
with fully explicit potentials
\begin{align*}
& \Rightarrow \bar{u}_{1\ }^{\ 1} = \frac{w_{2\ }^{\ 11} \theta(u_{2\ }^{\ 11}) u_{2\ }^{\ 11} (u_1 + w_{2\ }^{\ 11} \theta(u_{2\ }^{\ 11}) u_{2\ }^{\ 11} + w_{3\ }^{\ 12} \theta(u_{3\ }^{\ 12}) u_{3\ }^{\ 12})}{w_{2\ }^{\ 11} \theta(u_{2\ }^{\ 11}) u_{2\ }^{\ 11} + w_{3\ }^{\ 12} \theta(u_{3\ }^{\ 12}) u_{3\ }^{\ 12}}\\
& \Rightarrow \bar{u}_{1\ }^{\ 2} = \frac{w_{3\ }^{\ 12} \theta(u_{2\ }^{\ 11}) u_{2\ }^{\ 11} (u_1 + w_{2\ }^{\ 11} \theta(u_{2\ }^{\ 11}) u_{2\ }^{\ 11} + w_{3\ }^{\ 12} \theta(u_{3\ }^{\ 12}) u_{3\ }^{\ 12})}{w_{2\ }^{\ 11} \theta(u_{2\ }^{\ 11}) u_{2\ }^{\ 11} + w_{3\ }^{\ 12} \theta(u_{3\ }^{\ 12}) u_{3\ }^{\ 12}}
\end{align*}
In the case of three outputs the equilibrium would be
\begin{equation*}
(\bar{u}_{1\ }^{\ 1}, \bar{u}_{1\ }^{\ 2}, \bar{u}_{1\ }^{\ 3}) = \bigg(\frac{w_1 (u_1 + \sum_i^3 w_i)}{\sum_i^3 w_i}, \frac{w_2 (u_1 + \sum_i^3 w_i)}{\sum_i^3 w_i}, \frac{w_3 (u_1 + \sum_i^3 w_i)}{\sum_i^3 w_i}\bigg)
\end{equation*}
and analogically for more outputs. The phase plane for positive and negative weights (with one zero weight and two zero weights represented by the previous cases) is shown on figure \ref{phase_m}.

In the case of positive metaconnection weights $w_{2\ }^{\ 11}$ and $w_{3\ }^{\ 12}$ and therefore contributions $w_1$ and $w_2$ (zero cases again reducing to previous motifs) this is a stable 2-surface equilibrium where the two competing output connections $u_{1\ }^{\ 1}$ and $u_{1\ }^{\ 2}$ balance according to the ratio of their contribution and the total contribution of all metaconnections. The positivity of the contributions implies positivity of the horizontal asymptotes for each connection (the slanted asymptotes being the same $u_{1\ }^{\ 1} + u_{1\ }^{\ 2} = u_1$ line) which then suggests that the nullclines will always intersect once resulting in this equilibrium point. In the case of negative metaconnection weights $w_{2\ }^{\ 11}$ and $w_{3\ }^{\ 12}$ it is a unstable 2-surface equilibrium whose perturbation in one of the two unbalancing zones will result in it reaching either $(0,u_1+w_2)$ or $(u_1+w_1,0)$ and getting locked there due to the resetting condition. In this case we have negative horizontal asymptotes and both nullclines will always contain the point $(0,0)$ but for asymptotes that are negative enough the nullclines will no longer intersect in the positive region and the system will converge to $(0,0)$. Finally, in the cases of one positive and one negative weight (not shown on the figure) we have no nontrivial 2-surface equilibria in the nonnegative region since both nullclines that enter it will be on the two sides of the $u_{1\ }^{\ 1} + u_{1\ }^{\ 2} = u_1$ asymptote. In this case, the system will converge to $(u_1+w_1,0)$ if $w_1$ is the positive contribution and to $(0,u_1+w_2)$ if $w_2$ is the positive contribution.

The interpretation of a positive and a negative weight is straightforward - the unit outputs through the excited connection and not through the inhibited connection. The case of negative weights has a possible balance and possible perturbations (still balanced increase or decrease in both connections) under which this balance will be restored but other perturbations (e.g. any increase in one connection's potential and decrease in the other) under which the balance will tip over to a single winner. The case of positive weights is always balanced, never lies on the axis of any of the connections (unless we have zero weights and reduction to previous motifs), and most interestingly provides any connection with the ratio of its contribution to the contributions of all. If one of the two weights approaches zero, the connection with the other weight will approach the total of the input and its own contribution.

A competitive reduced version of (\ref{mm13_is}) adds feedback connections and therefore competing cycles. Once again, assuming nonzero weights $w_{2\ }^{\ 11} \neq 0$ and $w_{3\ }^{\ 12} \neq 0$ (any other case reduces to the previous motifs (\ref{mm5_is_v}, \ref{mm13_is_v})), we get the balanced full equilibrium
\begin{align*}
& \bar{u}_{1\ }^{\ 1} = \frac{w_{2\ }^{\ 11} U_2 (U_1 + w_{2\ }^{\ 11} U_2 + w_{3\ }^{\ 12} U_3)}{(1 - w_{2\ }^{\ 11}w^{1\ }_{\ 2} w_{1\ }^{\ 1}) (w_{2\ }^{\ 11} U_2 + w_{3\ }^{\ 12} U_3)}\\
& \bar{u}_{1\ }^{\ 2} = \frac{w_{3\ }^{\ 12} U_3 (U_1 + w_{2\ }^{\ 11} U_2 + w_{3\ }^{\ 12} U_3)}{(1 - w_{3\ }^{\ 12}w^{2\ }_{\ 3} w_{1\ }^{\ 2}) (w_{2\ }^{\ 11} U_2 + w_{3\ }^{\ 12} U_3)}
\end{align*}
which is stable for positive weights, unstable for negative weights, and exists only if the two weights have the same sign similarly to the previous motif. This equilibrium corresponds to the amplified (first) equilibrium of (\ref{mm13_is_v}) when one of the two weights is set to zero. The inhibitory-only (second) equilibrium of (\ref{mm13_is_v}) is now missing since every output connection is influenced by a metaconnection and canceling the contribution to a connection will tip over to the one that still has a contribution. At the same time, similar amplification rules apply as for (\ref{mm13_is_v}) which can allow otherwise nearly depleted connections due to high competition like for instance $(w_{2\ }^{\ 11} U_2)/(w_{2\ }^{\ 11} U_2 + w_{3\ }^{\ 12} U_3) \ll 1$ to still be highly active through $1/(1 - w_{3\ }^{\ 12}w^{2\ }_{\ 3} w_{1\ }^{\ 2}) \gg 1$.

Looking back at the full version of (\ref{mm5_is}), the partial 1-surface equilibrium states for all units are
\begin{align*}
\bar{u}^i = \sum_{j=1}^N w_{j\ }^{\ i} \theta(u_{j\ }^{\ i}) u_{j\ }^{\ i} = \sum_{j=1}^N w_{j\ }^{\ i} u_{j\ }^{\ i}
\end{align*}
since $\theta(u_{j\ }^{\ i})=0$ if $u_{j\ }^{\ i}=0$ or else $\theta(u_{j\ }^{\ i})=1$. The partial 1-surface equilibrium states for all connections satisfy
\begin{align*}
(\theta(\bar{u}_{j\ }^{\ i}) \bar{u}_{j\ }^{\ i} - \sum_{\substack{k = 1 \\ k \neq j}}^N w_{k\ }^{\ ji} \theta(u_{k\ }^{\ ji}) u_{k\ }^{\ ji}) \sum\limits_{l=1}^M \sum\limits_{\substack{m = 0 \\ m \neq j}}^N u_{j\ }^{\ ml} = \bar{u}_{j\ }^{\ i} u_j
\end{align*}
and for all metaconnections satisfy
\begin{align*}
\theta(\bar{u}_{k\ }^{\ ji}) \bar{u}_{k\ }^{\ ji} \sum\limits_{l=1}^M \sum\limits_{\substack{m = 0 \\ m \neq k}}^N u_{k\ }^{\ ml} = \bar{u}_{k\ }^{\ ji} u_k \Rightarrow \bar{u}_{k\ }^{\ ji} = 0 \text{ or } \sum\limits_{l=1}^M \sum\limits_{\substack{m = 0 \\ m \neq k}}^N u_{k\ }^{\ ml} = u_k
\end{align*}
The conditions (\ref{ns_conds},\ref{nss_conds}) for all connections are
\begin{align*}
\begin{dcases}
& \bigg|\sqrt{r_{j\ }^{\ i}} - \sqrt{u_j}\bigg| \geq \sqrt{- \sum_{\substack{k = 1 \\ k \neq j}}^N w_{k\ }^{\ ji} \theta(u_{k\ }^{\ ji}) u_{k\ }^{\ ji}} \in \mathbb{R} \text{ for } \bar{u}_{j\ }^{\ i}\\
& \text{always true if} \sum_{\substack{k = 1 \\ k \neq j}}^N w_{k\ }^{\ ji} \theta(u_{k\ }^{\ ji}) u_{k\ }^{\ ji} > 0 \text{ for } \bar{u}_{j\ }^{\ i}
\end{dcases}
\end{align*}
while in the case of metaconnections, they reduce further to
\begin{align*}
& \bigg|\sqrt{r_{k\ }^{\ ji}} - \sqrt{u_k}\bigg| \geq 0 \text{ for } \bar{u}_{k\ }^{\ ji}
\end{align*}
Thus, the partial 1-surface equilibria for all connections are
\begin{align*}
\begin{dcases}
& \bar{u}_{j\ }^{\ i} = u_j + \sum_{\substack{k = 1 \\ k \neq j}}^N w_{k\ }^{\ ji} \theta(u_{k\ }^{\ ji}) u_{k\ }^{\ ji} \text{ if } r_{j\ }^{\ i} = 0 \text{ and } \bar{u}_{j\ }^{\ i} \neq 0\\
& \bar{u}_{j\ }^{\ i} = u_{1/2} \text{ if } \big|\sqrt{r_{j\ }^{\ i}} - \sqrt{u_j}\big| > \sqrt{- \sum_{\substack{k = 1 \\ k \neq j}}^N w_{k\ }^{\ ji} \theta(u_{k\ }^{\ ji}) u_{k\ }^{\ ji}}\\
& \bar{u}_{j\ }^{\ i} = -r_{j\ }^{\ i} + \sqrt{u_j r_{j\ }^{\ i}}\\
& \qquad \text{ if } \big|\sqrt{r_{j\ }^{\ i}} - \sqrt{u_j}\big| = \sqrt{- \sum_{\substack{k = 1 \\ k \neq j}}^N w_{k\ }^{\ ji} \theta(u_{k\ }^{\ ji}) u_{k\ }^{\ ji}}
\end{dcases}
\end{align*}
for $\sum_{\substack{k = 1 \\ k \neq j}}^N w_{k\ }^{\ ji} \theta(u_{k\ }^{\ ji}) u_{k\ }^{\ ji} \leq 0$ and
\begin{align*}
\begin{dcases}
& \bar{u}_{j\ }^{\ i} = u_j + \sum_{\substack{k = 1 \\ k \neq j}}^N w_{k\ }^{\ ji} \theta(u_{k\ }^{\ ji}) u_{k\ }^{\ ji} \text{ if } r_{j\ }^{\ i} = 0 \text{ and } \bar{u}_{j\ }^{\ i} \neq 0\\
& \bar{u}_{j\ }^{\ i} = u_{1/2} \text{ otherwise }
\end{dcases}
\end{align*}
for $\sum_{\substack{k = 1 \\ k \neq j}}^N w_{k\ }^{\ ji} \theta(u_{k\ }^{\ ji}) u_{k\ }^{\ ji} > 0$. The partial 1-surface equilibria for all metaconnections are
\begin{align*}
\begin{dcases}
& \bar{u}_{k\ }^{\ ji} = u_k \text{ if } r_{k\ }^{\ ji} = 0 \text{ and } \bar{u}_{k\ }^{\ ji} \neq 0\\
& \bar{u}_{k\ }^{\ ji} = 0 \text{ or } \bar{u}_{k\ }^{\ ji} = u_k - r_{k\ }^{\ ji} \text{ if } \bigg|\sqrt{r_{k\ }^{\ ji}} - \sqrt{u_k}\bigg| > 0\\
& \bar{u}_{k\ }^{\ ji} = 0 \text{ if } r_{k\ }^{\ ji} = u_k
\end{dcases}
\end{align*}

If we fix $\bar{k} \in [1;N]$, i.e. we look at the emitting dynamics of a single input unit, we can utilize the simpler structure of the metaconnections to deduce some higher-dimensional space equilibria. In particular, for all $i \in [1;M]$ we have
\begin{align*}
& (\theta(\bar{u}_{\bar{k}\ }^{\ i}) \bar{u}_{\bar{k}\ }^{\ i} - \sum_{\substack{m = 1 \\ m \neq \bar{k}}}^N w_{m\ }^{\ \bar{k}i} \theta(u_{m\ }^{\ \bar{k}i}) u_{m\ }^{\ \bar{k}i}) u_{\bar{k}} = \bar{u}_{\bar{k}\ }^{\ i} u_{\bar{k}} \text{ if } \sum\limits_{l=1}^M \sum\limits_{\substack{m = 0 \\ m \neq \bar{k}}}^N \bar{u}_{\bar{k}\ }^{\ ml} = u_{\bar{k}}\\
& (\theta(\bar{u}_{\bar{k}\ }^{\ i}) \bar{u}_{\bar{k}\ }^{\ i} - \sum_{\substack{m = 1 \\ m \neq \bar{k}}}^N w_{m\ }^{\ \bar{k}i} \theta(u_{m\ }^{\ \bar{k}i}) u_{m\ }^{\ \bar{k}i}) \sum\limits_{l=1}^M \bar{u}_{\bar{k}\ }^{\ l} = \bar{u}_{\bar{k}\ }^{\ i} u_{\bar{k}} \text{ if } \bar{u}_{\bar{k}\ }^{\ ml} = 0, m \neq 0
\end{align*}
where the first case yields an attractor $(MN-1)$-plane of equilibria
\begin{equation*}
\sum\limits_{l=1}^M \sum\limits_{\substack{m = 0 \\ m \neq \bar{k}}}^N \bar{u}_{\bar{k}\ }^{\ ml} = u_{\bar{k}} \quad\text{ if }\quad \sum_{\substack{m = 1 \\ m \neq \bar{k}}}^N w_{m\ }^{\ \bar{k}i} \theta(u_{m\ }^{\ \bar{k}i}) u_{m\ }^{\ \bar{k}i} = 0
\end{equation*}
and the second case contains a system of connection equations which is solved in the same way as before and yields a balanced equilibrium
\begin{equation*}
\bar{u}_{\bar{k}\ }^{\ i} = \frac{\sum\limits_{\substack{m = 1 \\ m \neq \bar{k}}}^N w_{m\ }^{\ \bar{k}i} \theta(u_{m\ }^{\ \bar{k}i}) u_{m\ }^{\ \bar{k}i} (u_{\bar{k}} + \sum\limits_{l = 1}^{M} \sum\limits_{\substack{m = 1 \\ m \neq \bar{k}}}^N w_{m\ }^{\ \bar{k}l} \theta(u_{m\ }^{\ \bar{k}l}) u_{m\ }^{\ \bar{k}l})}{\sum\limits_{l = 1}^{M} \sum\limits_{\substack{m = 1 \\ m \neq \bar{k}}}^N w_{m\ }^{\ \bar{k}l} \theta(u_{m\ }^{\ \bar{k}l}) u_{m\ }^{\ \bar{k}l}}
\end{equation*}
If we assume the input balance condition $\sum_{m \neq \bar{k}} w_{m\ }^{\ \bar{k}i} \theta(u_{m\ }^{\ \bar{k}i}) u_{m\ }^{\ \bar{k}i} = 0$ for all connections or simply for all $i$, the second case will rather become a subcase of the first
\begin{equation*}
\sum\limits_{l=1}^M \bar{u}_{\bar{k}\ }^{\ l} = u_{\bar{k}} \quad\text{ and }\quad \bar{u}_{\bar{k}\ }^{\ ml} = 0, m \neq 0 \quad\text{ if }\quad \sum_{\substack{m = 1 \\ m \neq \bar{k}}}^N w_{m\ }^{\ \bar{k}i} \theta(\bar{u}_{m\ }^{\ \bar{k}i}) \bar{u}_{m\ }^{\ \bar{k}i} = 0
\end{equation*}
Instead, for our derivation of the second case we need the balancing condition for at least one output connection to be violated or its total contribution to be nontrivial. If the balance condition holds for a connection $\bar{u}_{\bar{k}\ }^{\ i}$ in general, from replacements above it can be seen that $\bar{u}_{\bar{k}\ }^{\ i}=0$ and every other competitive connection lacks the contributions of the metaconnections of $\bar{u}_{\bar{k}\ }^{\ i}$. In the case of negative total contribution for a connection and positive for another, by previous conclusions we have no stable state, the first connection will decay to zero, and the balance equilibrium will be restored among the other output connections. This means that we can disqualify all connections with negative total contribution in the consideration of a steady state. There could also be other cancellations in the above fraction but if one total contribution is guaranteed to be positive and we disqualify all negative cases, the fraction remains well-defined.

The partial equilibria for a fixed $\bar{k}$ are equilibria on the $MN$-dimensional manifold only. If we consider two input units $k=1,2$ we can obtain a higher-dimensional surface equilibrium with both of them if we satisfy the attractor $(MN-1)$-plane conditions
\begin{equation*}
\sum\limits_{l=1}^M \sum\limits_{\substack{m = 0 \\ m \neq k}}^N \bar{u}_{k\ }^{\ ml} = u_k \quad\text{ and }\quad \sum_{\substack{l = 1 \\ l \neq k}}^N w_{l\ }^{\ ki} \theta(\bar{u}_{l\ }^{\ ki}) \bar{u}_{l\ }^{\ ki} = 0
\end{equation*}
which are easy to interpret: all metaconnections settle once the total potential of a unit is distributed among its output connections or even are permanently deactivated (once deactivated, a connection needs input from metaconnections to get more potential from the unit) and all connections settle to a steady state when their input metaconnections balance and settle. However, having a $(MN-1)$-plane attractor in the output units is not always preferred and it is not possible to instead find balanced equilibria between two units since each one requires deactivated output metaconnections and active input metaconnections at the same time (due to the nontrivial total contribution requirement). In lieu of this, a balanced unit equilibrium must be accompanied by an attractor plane of another
\begin{align} \label{mm5_is2}
& \sum\limits_{l=1}^M \sum\limits_{\substack{m = 0 \\ m \neq 1}}^N \bar{u}_{1\ }^{\ ml} = u_1 \quad\text{ and }\quad \sum_{\substack{l = 1 \\ l \neq 1}}^N w_{l\ }^{\ 1i} \theta(\bar{u}_{l\ }^{\ 1i}) \bar{u}_{l\ }^{\ 1i} = 0\\
& \bar{u}_{2\ }^{\ i} = \frac{\sum\limits_{\substack{m = 1 \\ m \neq 2}}^N w_{m\ }^{\ 2i} \theta(u_{m\ }^{\ 2i}) u_{m\ }^{\ 2i} (u_2 + \sum\limits_{l = 1}^{M} \sum\limits_{\substack{m = 1 \\ m \neq 2}}^N w_{m\ }^{\ 2l} \theta(u_{m\ }^{\ 2l}) u_{m\ }^{\ 2l})}{\sum\limits_{l = 1}^{M} \sum\limits_{\substack{m = 1 \\ m \neq 2}}^N w_{m\ }^{\ 2l} \theta(u_{m\ }^{\ 2l}) u_{m\ }^{\ 2l}}
\end{align}
The interpretation of all this is important - we cannot have properly explained and therefore concentrated inputs unless we also have some context-providing inputs or overall input features. Notice however that all of this must hold asymptotically and there is no such restriction imposed for any transient behavior, i.e. both units might behave as both contextualized and feature units before settling. This and a more involved two-unit version of (\ref{mm13_is}) will be used to shed first light on the possibility of learning in section \ref{pattern_classification}.

We should complete the section with the full version (\ref{mm13_is}) which also contains connection cycles. In short, using the simpler structure of all metaconnections but also the lateral and feedback connections (which also don't have metaconnection inputs), we can obtain the $(MN-1)$-plane attractors as (with notation abuse like $u=\bar{u}$)
\begin{align*}
\begin{dcases}
& u^i = \sum_{j=1}^N w_{j\ }^{\ i} \theta(u_{j\ }^{\ i}) u_{j\ }^{\ i} + \sum_{l=1}^M w^{li} \theta(u^{li}) u^{li}\\
& \theta(u_{j\ }^{\ i}) u_{j\ }^{\ i} - \sum_{\substack{k = 1 \\ k \neq j}}^N w_{k\ }^{\ ji} \theta(u_{k\ }^{\ ji}) u_{k\ }^{\ ji} - \sum_{l=1}^M w^{l\ }_{\ ji} \theta(u^{l\ }_{\ ji}) u^{l\ }_{\ ji} = u_{j\ }^{\ i}\\
& \sum\limits_{m=1}^M [u^{lm} + \sum\limits_{\substack{n = 0 \\ n \neq l}}^N u^{l\ }_{\ nm}] = u^l = \sum_{j=1}^N w_{j\ }^{\ l} \theta(u_{j\ }^{\ l}) u_{j\ }^{\ l} + \sum_{i=1}^M w^{il} \theta(u^{il}) u^{il}\\
& \sum\limits_{m=1}^M [u^{im} + \sum\limits_{\substack{n = 0 \\ n \neq i}}^N u^{i\ }_{\ nm}] = u^i = \sum_{j=1}^N w_{j\ }^{\ i} \theta(u_{j\ }^{\ i}) u_{j\ }^{\ i} + \sum_{l=1}^M w^{li} \theta(u^{li}) u^{li}\\
& \sum\limits_{m=1}^M \sum\limits_{\substack{n = 0 \\ n \neq k}}^N u_{k\ }^{\ nm} = u_k = \sum_{i=1}^M w^{i\ }_{\ k} \theta(u^{i\ }_{\ k}) u^{i\ }_{\ k} + U_k\\
& \sum\limits_{m=1}^M [u^{lm} + \sum\limits_{\substack{n = 0 \\ n \neq l}}^N u^{l\ }_{\ nm}] = u^l = \sum_{j=1}^N w_{j\ }^{\ l} \theta(u_{j\ }^{\ l}) u_{j\ }^{\ l} + \sum_{i=1}^M w^{il} \theta(u^{il}) u^{il}\\
& u_j = \sum_{i=1}^M w^{i\ }_{\ j} \theta(u^{i\ }_{\ j}) u^{i\ }_{\ j} + U_j
\end{dcases}
\end{align*}
which reduces to
\begin{align*}
\begin{dcases}
& \sum_{\substack{k = 1 \\ k \neq j}}^N w_{k\ }^{\ ji} \theta(u_{k\ }^{\ ji}) u_{k\ }^{\ ji} + \sum_{l=1}^M w^{l\ }_{\ ji} \theta(u^{l\ }_{\ ji}) u^{l\ }_{\ ji} = 0, i \in [1;M], j \in [1;N]\\
& \sum\limits_{m=1}^M [u^{im} + \sum\limits_{\substack{n = 0 \\ n \neq i}}^N u^{i\ }_{\ nm}] = \sum_{k=1}^N w_{k\ }^{\ i} \theta(u_{k\ }^{\ i}) u_{k\ }^{\ i} + \sum_{l=1}^M w^{li} \theta(u^{li}) u^{li} = u^i\\
& u_j = \sum\limits_{m=1}^M \sum\limits_{\substack{n = 0 \\ n \neq j}}^N u_{j\ }^{\ nm} = \sum_{l=1}^M w^{l\ }_{\ j} \theta(u^{l\ }_{\ j}) u^{l\ }_{\ j} + U_j
\end{dcases}
\end{align*}
The previous balancing condition for each feedforward connection is now extended to include feedback metaconnections as well. There is an extra output settling condition this time from the output to the input units and both such conditions now also relate the inputs with the outputs of a unit. Before we used one of the $N$ equations of the third type and then of one of the $M$ equations of the second type to close the system (\ref{mm13_is_v}).

As before, we can also obtain some  balancing equilibria but this time no unit needs its output connections to satisfy the attractor $(NM-1)$-plane conditions since all feedback connections don't have input metaconnections and no feedback connection will go to zero guaranteeing nontrivial contributions. This can be done analogically to before assuming simple enough connection cycles like the case of the previous two motifs of (\ref{mm13_is}). Otherwise, obtaining analytic expression is still possible tracing the various dependencies of an unknown on itself or using symbolic solver but not practical to include here and more useful to analyze on the numerical side.

\subsection{Numerical implementations}

Even though the emphasis of this work is on the mathematical formulation and analysis of metanetwork models, we will briefly outline some of their numerical realizations. In particular, we will formalize the setting for each field of application and share observations from any relevant simulation. None of these developed prototypes were planned to compete with specialized state-of-the-art solutions and they serve rather an illustrative purpose for the family of models and the motivation behind them.

\subsubsection{Feature detector} \label{feature_detection}

\begin{figure}[!htbp]
\centering
\includegraphics[width=0.7\textwidth]{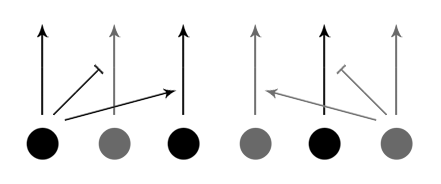}
\caption{Connectivity structure used for the feature detector - the drawn connections are replicated for each input unit of the given type.}
\label{filter}
\end{figure}

The first numerical implementation we will present is the generation of input selectivity in an approximation of (\ref{mm12}) which can be considered as contrast on a lower level in the processing chain, i.e. detection of corner and edges features. Using a simple forward Euler method or a first order approximation for the derivative with $\delta=\tau_r$ as $\tau_r du/dt \approx \tau_r (u(t+\delta)-u(t))/\delta = u(t+\tau_r) - u(t)$ and normalizing the time scale, we can produce the simulation algorithm
\begin{align} \label{mm12_approx}
\begin{dcases}
& u^i(t+1) = u^i(t)  - a(u^i(t)) + \sum_{j=1}^N \bar{a}(u_{j\ }^{\ i}(t), u_j(t))\\
& u_{j\ }^{\ i}(t+1) = u_{j\ }^{\ i}(t)  - \bar{a}(u_{j\ }^{\ i}(t), u_j(t)) + \sum_{\substack{k = 1 \\ k \neq j}}^N \bar{a}(u_{k\ }^{\ ji}(t), u_k(t)) +\\
& \qquad - \sum_{\substack{k = 1 \\ k \neq j}}^N \bar{a}(v_{k\ }^{\ ji}(t), v_k(t)) + e(u_{j\ }^{\ i}(t), u_j(t), u_{j\ }^{\ 0,1}(t), \dots, u_{j\ }^{\ NM}(t))\\
& u_{k\ }^{\ ji}(t+1) = u_{k\ }^{\ ji}(t) - \bar{a}(u_{k\ }^{\ ji}(t), u_k(t)) + e(u_{k\ }^{\ ji}(t), u_k(t), ...),\ k \neq j
\end{dcases}
\end{align}
Each pixel of an input grayscale image is processed by two types of units. The first type must process the amount of white $\omega \in [0,1]$ in the pixel and the second the amount of black $1-\omega$. The two-layer structure of (\ref{mm12_approx}) then filters this image by providing an output of the same size which contains higher values for more important parts of the image or parts drawing the attention of the network. The connectivity between the input and output layer is described in figure \ref{filter} with the addition that the network is implemented as a cellular neural network (also abbreviated as CNN). What this means is that full interlayer connectivity is replaced with local one where a metaconnection from a unit will excite/inhibit only the connections of its neighbors of the same/other type within a certain range. In this way we obtain sparse connectivity and an output unit reacts only to a region of the input image or biologically speaking its visual field. To avoid effects of the layer boundaries due to the sparse connectivity, the network is implemented on a torus, i.e. the connectivity follows a periodic surface.

We have studied many variations of the above filter where we also use reverse sign interactions or even only excitations (using weak inhibition which was defined earlier as reducing the potential of a connection by diverting the potential of its input unit), where we output the potential only if it is within certain range (to make the output sparse) or replace the output potential with a measurement of the uncertainty of each unit (which was defined earlier as broadcasting small amounts of potential through each output connection) as an indicator of importance and greater attention of the network. Many of them give interesting insights into possible interpretations of features but such a discussion is too long for this presentation. The numerical results we will report here will concern the self-excitation and mutual inhibition of the two types of units and the uncertainty measurement of attended input.

We ran simulations with visual inputs that range from a single contrasting line to small resolution videos, some of which are included in the appendix on figure \ref{static_features}. A sanity check with uniform white or grey (less but equal activity in both types of units) input produced no output at all. When looking at a contour line separating a black and a white uniform regions, the network immediately increased its attention at the line and if the line separated a white and a middle grey regions the same happened but it took longer. Further decreasing contrast resulted in further decreasing speed which eventually took too long to observe. When looking at two simultaneous contour lines of high and low contrast, the network first noticed the first line and if provided with sufficient time also the second. This is desired for an optimal visual processing when there could be too much input to attend to and the computational resources are limited for all of it. A gradient was also reduced to simpler contrast which implies a possibility to process blurry input (one of the most challenging type of features with relatively high error rates for both convolutional neural networks and major supervision algorithms like SIFT). A contour line or gradient with higher overall luminosity was registered by the first type and vice versa which implies that units are more sensitive to disbalance if the competing type is silent.

When looking at a triangle, the network first noticed the corners and gradually expanded its attention to the edges and ultimately the internal area. This implies higher feature importance of corners for the determination of shapes and resulting objects which is a thing to expect. When challenged with a mixture of corners with varying contrast on a cube, the network started with the high contrast angle and moved towards the low contrast angle with preference of contrast over corners. It did this more easily with the first unit type and remained more reserved with the second unit type, mostly due to the dominance of white in the input. When looking at a star filled with noise, the network was able to first ignore part of the noise and concentrate on the shape where it preferred the sharper angles in the outer part and then directed its attention to the inner obtuse angles and the edges. The same was observed for a square with different levels of noise but with parts of the square being harder to sustain and in favor of the noise. In both these last cases, the first unit type exhibited the described resilience to noise while the lack of sufficient activation in the second unit type could be the possible (and even of a necessity) reason it was more susceptible to noise.

The feature detector from (\ref{mm12_approx}) evolved considerably even over static input, shifting its attention to novelty or part of the input which wasn't processed before due to limited attention span at all times. When looking at a rotating triangle, the network shifted its attention in accordance with the moving corners, tracing the most changing parts of the input image. While it generally has time to pay attention to less important features of a static triangle, it only has time to perceive the main features of the moving triangle before the input changes again. When looking at static, blinking, and moving dots, the network noticed all three although we would hope to perceive them in a fixed order depending on complexity. Such order could be different depending on the knowledge of the network and which input would be too simple or too complex for it to notice first - for a sufficiently experienced network one should expect that changes in the input larger than the previously learned ones should get more attention. Nevertheless, we found increase of attention also around the blinking point and along the line tracing the trajectory of the moving point which hints towards temporal shifts of attention, i.e. shifts of attention influenced by past locations of a visual stimulus. We have provided visuals about this and some tested videos in the appendix figure \ref{dynamic_features}.

All the results above were obtained for fixed size images of 100 by 100 pixels with the exception of the videos which were 176x144 pixels. No scaling or preprocessing was performed on any image or sequence frame although most of the images were created especially for these tests. The same set of parameters was used for all cases in order to maximize reproducibility, namely a connection range of one (metaconnections to nearest neighbors only) and a potential range of $[0.05, 3.0]$ for output connections which if more than one would qualify the units as broadcasting due to uncertainty. The increase or decrease of the connectivity range changes the level of complexity of the detected features, could account for larger (scaled and blurry) angles, and changes the sensitivity of the overall detector by increasing the visual field of each output unit. The increase or decrease of the potential range has significant effect on the final number of detected features and is generally used to obtain sparse features of fixed size from a multitude of features of various sizes which are harder to analyze.

The sparse connectivity implementation and the symmetrical arrangement of single or multiple cell types is an intrinsic part of many network models of the center-surround receptive field. These models involve lateral inhibition either as lateral connections on the output layer or as feedforward connections from the inputs to the output units \cite{Mallot2013, Grossberg1976A} and also produce forms of contrast at the output layer. There are also infinitesimal formulations of simple enough inhibitory interactions that use convolution of a point image with the input but the current network has additional dynamics which violate and prevent the superposition principle and therefore at least the conceptually straightforward possibility of using convolution.

Generally, there is no established way to compare quality of feature detection unless it is accompanied by other layers of processing involving feature descriptors and feature matching. What is done sometimes in computational vision is to reproduce visual illusions like the ones described by \cite{LevineGrossberg1976, MacknikMartinez2004}. In this way we can extract particularities of the visual system which can reveal more about the way it works and provide concrete ground for biomimic engineering approaches and comparison. There are numerous illusions related to contrast and explained through lateral inhibition, most known of which are the Mach bands illusion, the Cornsweet illusion, Hermann grid illusion, the contrast effect, etc. Testing the feature detector on these did not provide any results, possibly due to small connection range or even some of the choices about the network connectivity.

\subsubsection{Object tracker} \label{object_tracking}

The previous feature detector can be extended to track the most interesting features with higher resolution (density of units) near the point of interest and lower resolution far from it. These differences in resolution are inspired by the retinotopic mapping demonstrated experimentaly in \cite{TootellSilvermanSwitkesDevalois1982} which relates proximity in areas of the visual image to proximity on the retina. In fact, their paper showed that the mapping is distorted in favor of the most central area with the most peripheral area taking a rather insignificant portion of the retina. Using a simpler version of this with just two levels of resolution, the present object tracker can be used for moving an eye or camera in the direction of the most interesting feature therefore optimizing the information inflow and reducing the processing of unnecessary details.

Another significant role of allowing the network to change its perspective is explained by \cite{MartinezMacknikHubel2004} in their study of the fixational eye movement. As they show, even when the eye fixates on an object, there is always a microscopic movement which counteracts fading of the input due to adaptation. Connecting the output of the network to a camera motor, virtual selection of a subimage in a larger image, or any other control that would allow it to make decisions about its orientation and resolution of the environment would prevent it from overadapting to a constant input which in the case of our feature detector could be represented by an overwhelming increase of broadcasting and therefore spread of attention and decrease of the potential on the output layer.

We have included one last figure in the visual appendix namely figure \ref{tracker_resolutions} where the two resolutions together with the currently observed region of the input are shown. The information perceived in the lower resolution environment will direct the higher resolution region towards the feature of highest interest. In the case of a sequence of a rotating triangles, the attention and therefore the "gaze" of the tracker is focused on the moving angles. As a result, it follows the rotation of the triangle. In the case of a sequence of static, moving and blinking points, the network switches view from the moving point to the static point until it settles for the static point. It looks only briefly at the blinking point. In the previous case where the network couldn't direct its attention, even though this input was dynamic it led to overadaptation where the activity spread around all points and their trajectories. Of course the same happened also in the peripheral vision of the tracker but for significantly long simulation its direction remained fixated on the static point. We would still expect it to move towards the more dynamical points which might be achievable with a better choice of parameters, longer simulation or with the addition of learning which should help drive attention away from overadapted input.

A thorough comparison of the state of the art object tracking algorithms on various challenging datasets was performed by \cite{WangChenXuYang2011}. The two criteria for comparison they use are tracking success rate and location error with respect to the object center. In order to compare the object center, the present tracker needs at least multiple layers of abstraction in order to build objects as composite features and track those. While a tracking success rate could be extracted, the one used by the paper requires boxed objects to overlap and we are currently only boxing around a single feature. As all present implementations are merely illustrative of the fields of application and the way the family of models can be shaped into such applications, we will not investigate this further but suggest multiple layers for the curious in this direction.

\subsubsection{Pattern classifier} \label{pattern_classification}

Even though the prospect of creating effective learning rules to form the right attractors with the right basins of attraction seems far off given the scope of the current work, we could at least consider a possible synthesis procedure, similarly to (\ref{syn1}, \ref{syn2}) for the additive network models. Such procedures are the perfect middle step between the better understood study of network activity (potential phase space) and the full scale study of network connectivity (weight phase space). First, they are a bridge from the structurally predetermined feature detection into the unsupervised classification by memory recall and pattern recognition but a bridge which remains structurally predetermined. Second, they can be considered as instantaneous learning where one doesn't need to worry about slower weight dynamics and any resulting structural stability as well as bifurcations of the potential dynamics (for instance memory consolidation can take the form of differentiation of recallable states which could be realized through pitchfork bifurcation or appearance of two equilibrium points). All memories are usually stored in one shot when the weights are determined so that the network has these as attractors and the weights are fixed and static afterwards with no learning taking place from the network activity.

The formal requirements for a synthesis procedure are two:
\begin{itemize}
	\item all stored memories must be equilibria
	\item all stored memories must be asymptotically stable
\end{itemize}
and some known synthesis approaches satisfy one or both conditions with additional computational drawbacks \cite{MichelFarrell1990}. The goal then is to find weights for all connections so that the requirements are satisfied for a desired and externally preselected set of memories. To illustrate synthesis following these guidelines, we will consider again the two-unit model (\ref{mm5_is2}) of a feedforward network (\ref{mm5_is}) and to extend on the entire idea conceptually and numerically, we will consider a two-unit model of a reduced recurrent network (\ref{mm13_is}). For both of them we will also assume all practical conditions on nonnegativity, continuous activation, etc. used for their derivation. In addition to these conditions, we will require that the dynamics of the output connections of the feature unit $u_1$ be restricted to the set
\begin{equation} \label{unbiased_features}
\{u_{1\ }^{\ ml}=u_{1\ }^{\ m'l'}, \forall m', m \in [0;N]\backslash\{1\}, \forall l', l \in [1;M]\}
\end{equation}
Since we are guaranteed this to hold on broadcasting with each $u_{1\ }^{\ ml}=u_1/MN$, the only other thing we have to do is prevent nontrivial contributions to the output connections $\bar{u}_{1\ }^{\ ml}$ so that it holds for all times. We can do this by setting the weights of any metaconnections to these output connections to zero and disallowing nonzero external inputs for them. The effect of this extra assumption is threefold:
\begin{itemize}
	\item We avoid transient behavior where a unit could switch between the roles of feature and contextualized unit. This behavior is a lot more interesting but is currently avoided for the sake of simplicity.
	\item In this way the output connections of the unit can only converge to the middle point of the attractor $(MN-1)$-plane and we won't violate the second condition of asymptotic stability although we have a degenerate equilibrium point.
	\item We will reduce the degrees of freedom of the algebraic system for the equilibria, having fully determined values for all potentials so that we can formulate a system for the weights.
\end{itemize}
We will call $u_1$ an unbiased feature unit (a unit that at no times has metaconnection contributions to its output connections, i.e. is contextualized or explained by others) and $u_2$ a contextualized unit.

\begin{figure}[!htbp]
\centering
\includegraphics[width=0.8\textwidth]{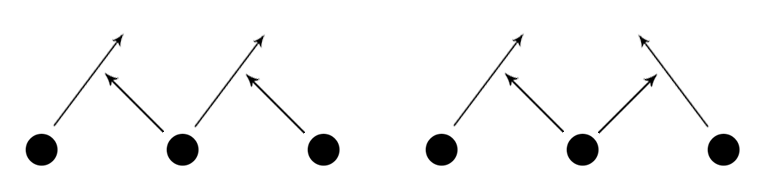}
\caption{Assumption on unbiased features forbids the first case with allows only the second case.}
\end{figure}

The current synthesis approach given the above requirements and assumptions is to construct an algebraic system in the weight space after first determining all equilibrium unknowns in the potential space, all of this for a fixed input $\tilde{u}_k$ and a fixed (desired for recall) output $\tilde{u}^i$. Taking a 2-output version ($i=1,2$) of (\ref{mm5_is2}) with the extra assumption (\ref{unbiased_features}) on unbiased features
\begin{equation*}
\tilde{u}_1 = u_{1\ }^{\ 1} + u_{1\ }^{\ 2} + u_{1\ }^{\ 21} + u_{1\ }^{\ 22} \Rightarrow 1/4 \tilde{u}_1 = u_{1\ }^{\ 1} = u_{1\ }^{\ 2} = u_{1\ }^{\ 21} = u_{1\ }^{\ 22}
\end{equation*}
then simplifying the remaining equations of the two input units
\begin{align*}
& 0 = w_{2\ }^{\ 11} u_{2\ }^{\ 11} = w_{2\ }^{\ 12} u_{2\ }^{\ 12} \text{ and } 0 = u_{2\ }^{\ 11} = u_{2\ }^{\ 12} \Rightarrow \forall w_{2\ }^{\ 11}, \forall w_{2\ }^{\ 12}\\
& \bar{u}_{2\ }^{\ i} = \frac{w_{1\ }^{\ 2i} u_{1\ }^{\ 2i} (u_2 + w_{1\ }^{\ 21} u_{1\ }^{\ 21} + w_{1\ }^{\ 22} u_{1\ }^{\ 22})}{w_{1\ }^{\ 21} u_{1\ }^{\ 21} + w_{1\ }^{\ 22} u_{1\ }^{\ 22}} = \frac{w_{1\ }^{\ 2i} (u_2 + 1/4 \tilde{u}_1 (w_{1\ }^{\ 21} + w_{1\ }^{\ 22}))}{w_{1\ }^{\ 21} + w_{1\ }^{\ 22}}
\end{align*}
and adding restrictions from the predetermined output units
\begin{align*}
&  \tilde{u}^1 = w_{1\ }^{\ 1} u_{1\ }^{\ 1} + w_{2\ }^{\ 1} u_{2\ }^{\ 1} = w_{1\ }^{\ 1} \frac{\tilde{u_1}}{4} + w_{2\ }^{\ 1} u_{2\ }^{\ 1}\\
&  \tilde{u}^2 = w_{1\ }^{\ 2} u_{1\ }^{\ 2} + w_{2\ }^{\ 2} u_{2\ }^{\ 2} = w_{1\ }^{\ 2} \frac{\tilde{u_1}}{4} + w_{2\ }^{\ 2} u_{2\ }^{\ 2}
\end{align*}
would lead to the underdetermined weight system
\begin{align*}
\begin{dcases}
& 0 w_{2\ }^{\ 1i} = 0\\
& 4 (\tilde{u}^i - w_{1\ }^{\ i} \tilde{u}_1) (w_{1\ }^{\ 21} + w_{1\ }^{\ 22}) = w_{1\ }^{\ i} w_{1\ }^{\ 2i} (4 \tilde{u}_2 + \tilde{u}_1 (w_{1\ }^{\ 21} + w_{1\ }^{\ 22}))\\
&  4 \tilde{u}^i - w_{1\ }^{\ i} \tilde{u}_1 = 4 w_{2\ }^{\ i} u_{2\ }^{\ i}
\end{dcases}
\end{align*}
Simulations with different choices of weights solving this system and initial conditions (always satisfying $u_{1\ }^{\ ml}=u_{1\ }^{\ m'l'}$ since other behaviors of the system are unreachable with the restriction (\ref{unbiased_features})) were performed where a desired output vector $[\tilde{u}^1, \tilde{u}^2]^T$ was repeatedly obtained from the constant input vector $[\tilde{u}_1, \tilde{u}_2]^T$. Analogously, we can consider the original algebraic system as a polynomial system with both the potentials and the weights as unknowns. Because of the unbiased features condition (\ref{unbiased_features}), the attractor $(MN-1)$-plane is replaced by an equilibrium point and the original system is fully determined in all potentials. We can can therefore express these potentials entirely in terms of the fixed input and output potentials, obtain equations only for the weights, and use these to place the partial (output only) equilibrium in a desired location within the subspace spanned by the output units. Because the full equilibrium points satisfying the system are stable, the network will converge to the desired configuration. Because the network is feedforward only and because of (\ref{unbiased_features}) there is just one such equilibrium point and it is the $\omega$-limit set of all system trajectories. In this way we can guarantee the reachability of the steady state in addition to its existence.

The lack of uniqueness of the solutions of this system can be used in order to match multiple pairs of inputs and equilibria. For instance, in order to obtain the equilibrium $[\tilde{u}^1, \tilde{u}^2]^T=[2, 2]^T$ from the input $[\tilde{u}_1, \tilde{u}_2]^T=[1, 1]^T$ and the equilibrium $[\tilde{u}^1, \tilde{u}^2]^T=[1, 2]^T$ from the input $[\tilde{u}_1, \tilde{u}_2]^T=[1, 2]^T$ , we have to form a new system from the solutions of the two systems to obtain a more restricted solution $w_{1\ }^{\ 1}=12+r$, $w_{1\ }^{\ 2}=8$, $w_{1\ }^{\ 22}=0$, $w_{2\ }^{\ 1}=-1$, $w_{1\ }^{\ 21}=r$, $w_{2\ }^{\ 2}=s$, $w_{2\ }^{\ 11} = p$, , $w_{2\ }^{\ 12} = q$. This can still be done manually but for sufficiently many systems we need to use at least a symbolic solver. It is important to note that the extra degrees of freedom cannot be useful in all possible cases of input-output associations and finding weights for multiple such might result in trivial or no solution. The reason that it is possible to find weights for some cases at all possibly has to do with the fact that the equations are polynomials of degree larger than one. It can be seen in the stability analysis of the motifs of (\ref{mm13_is}) that their balancing equilibria are even higher degree polynomials in the weights and look more promising for such endeavor. At the same time, such endeavor is necessary as we will show next.

The original cognitive interpretation of the attractor networks usually considers the initial conditions in the phase space as the actual clue that is used for the recall. The memories are then synthesized as asymptotically stable equilibria so that their basins of attraction represent their volumes of robustness with respect to these initial conditions. Such robustness with respect to previous states of the network makes a lot less sense than robustness with respect to the external input to the system which is usually kept constant or zero in attractor network studies. The external input which is nonzero only for input units is a lot more natural way to represent external information passed to the network than its initial condition. It also makes more sense as a conceptualization of cognitive tasks like recognition and prediction. Therefore, we will ultimately require that we have robustness with respect to the external input rather than the initial conditions.

We started considering the robustness with respect to external input for the 2-unit feedforward network (\ref{mm5_is2}) by obtaining weights that match multiple inputs with multiple desired output equilibria. The 2-output version of (\ref{mm13_is}) can be fully stated as
\begin{align*}
\begin{dcases}
& 0 = u_{1\ }^{\ 21} = u_{1\ }^{\ 22} = u_{2\ }^{\ 11} = u_{2\ }^{\ 12}\\
& \tilde{u}^i/6 = u^{i\ }_{\ 1} = u^{i\ }_{\ 2} = u^{i\ }_{\ 11} = u^{i\ }_{\ 12} = u^{i\ }_{\ 21} = u^{i\ }_{\ 22}\\
& \tilde{u}_i = U_i + w^{1\ }_{\ i} u^{1\ }_{\ i} + w^{2\ }_{\ i} u^{2\ }_{\ i} = U_i + (w^{1\ }_{\ i} \tilde{u}^1 + w^{2\ }_{\ i} \tilde{u}^2)/6\\
& \tilde{u}^i = w_{1\ }^{\ i} u_{1\ }^{\ i} + w_{2\ }^{\ i} u_{2\ }^{\ i}\\
& (6 u_{1\ }^{\ i} - w^{1\ }_{\ 1i} \tilde{u}^1 - w^{2\ }_{\ 1i} \tilde{u}^2) (u_{1\ }^{\ 1} + u_{1\ }^{\ 2}) = 6 u_{1\ }^{\ i} \tilde{u}_1\\
& (6 u_{2\ }^{\ i} - w^{1\ }_{\ 2i} \tilde{u}^1 - w^{2\ }_{\ 2i} \tilde{u}^2) (u_{2\ }^{\ 1} + u_{2\ }^{\ 2}) = 6 u_{2\ }^{\ i} \tilde{u}_2
\end{dcases}
\end{align*}
While the feedforward case is easy enough to be done by a symbolic solver, here this becomes computationally expensive and we have to rely on numeric solvers only. Using iterating method, we had to specify initial guess for the search for which we used unity vector and obtained a root for the system which was then reproduced via simulation. For the simulation we use a specialized algorithm which plays the role of network motif calculator, i.e. we can quickly construct and simulate any network motif of sufficiently small size (this includes all the above mentioned networks). Because the current network involves cycles, approximation errors for the weights tend to accumulate and we had to increase the precision of the system root to at least $\num{1e-16}$ to obtain good estimation for the equilibrium state. Furthermore, we had to restrict the potentials to the nonnegative region which was done by replacing the unknowns in the objective function by their squares and then recalculating the answer as square of the found root. Since the solution we can obtain can also be within the nonnegative region without this restriction and yields more accurate approximation in some cases, we took the liberty to use this as well. Conversely, we sometimes required nonnegativeness for the weights of the connections to restrict our focus even though there is no immediate reason for such requirement.

Some additional preliminary results include storing multiple input-output pairs, multiple outputs with the same input, and multiple inputs with the same output. This was done by generating additional equations and replacing the input and output unknowns with the fixed values for each pair. Here are conclusions we draw for each of the three cases respectively:
\begin{itemize}
	\item While numerical solutions were found for all attempted cases of input-output pairs, including ones with overlapping, close or distant (in terms of simple Euclidean norm) inputs, the accuracy of the recall deteriorated with the number of stored pairs. This could be due to increased sensitivity to the exact solution when it is a solution of many possibilities for equilibria. There were also spurious intermediary equilibria if the network is provided with noisy input outside of the learned pairs. This is usually expected in the sense that we change the external input which changes the dynamics and therefore the resulting equilibrium point. However, we are interested in precisely mapping multiple inputs to the same equilibrium which is the third case here.
	\item The observations in the case of multiple equilibria from a single external input match the theoretical predictions drawn from previous sections. In particular, there can either be one isolated stable equilibrium point or an attractor $(NM-1)$-plane and only the second option is possible if we require multiple equilibria. We considered for instance a case of the equilibrium $[\tilde{u}^1, \tilde{u}^2]^T=[1, 1]^T$ from the input $[\tilde{u}_1, \tilde{u}_2]^T=[1, 1]^T$ and $[\tilde{u}^1, \tilde{u}^2]^T=[2, 2]^T$ from the input $[\tilde{u}_1, \tilde{u}_2]^T=[1, 1]^T$ and obtained forward connection weights $w_{1\ }^{\ 1}=w_{2\ }^{\ 1}=\num{1.5297}$, $w_{1\ }^{\ 2}=w_{2\ }^{\ 2}=\num{1.4714}$, near-zero backward connection weights $w^{1\ }_{\ 1}=w^{2\ }_{\ 1}=\num{1.0336e-15}$, $w^{1\ }_{\ 2}=w^{2\ }_{\ 2}=\num{-1.2584e-15}$, and near-zero backward metaconnection weights $w^{1\ }_{\ 11}=w^{2\ }_{\ 11}=\num{-1.7451e-14}$, $w^{1\ }_{\ 12}=w^{2\ }_{\ 12}=\num{1.9641e-14}$, $w^{1\ }_{\ 21}=w^{2\ }_{\ 21}=\num{-1.3985e-14}$, $w^{1\ }_{\ 22}=\num{0.013908}$, and $w^{2\ }_{\ 22} = \num{-0.013908}$. This implies that the numerical result has excluded any influence on the forward connections by setting the metaconnection weights to zero. The result from excluding all backward connections' influence is that the network reduces to a feedforward network where we already know that we should have at least one feature unit and therefore attractor $(NM-1)$-plane. Simulations for these weights and external input $[\tilde{u}_1, \tilde{u}_2]^T=[1, 1]^T$ converge to an intermediate value on the attractor $[\tilde{u}^1, \tilde{u}^2]^T=[\num{1.5297}, \num{1.4714}]^T$ or nearby values using different initial conditions.
	\item The most important case is of course the one about robustness to external input, i.e. obtaining the same equilibrium from multiple external inputs. Numerical solutions for such weights like obtaining $[\tilde{u}^1, \tilde{u}^2]^T=[1, 1]^T$ from both $[\tilde{u}_1, \tilde{u}_2]^T=[1, 1]^T$ and $[\tilde{u}_1, \tilde{u}_2]^T=[2, 2]^T$ resulted in either overshoot $[\tilde{u}^1, \tilde{u}^2]^T=[1.33, 1.33]^T$ or undershoot $[\tilde{u}^1, \tilde{u}^2]^T=[0.66, 0.66]^T$ of the desired equilibrium depending on the choice of desired input. Single attractor point is therefore only the answer for obtaining robustness with respect to initial conditions and not with respect to external input. To obtain the same attractor for a set of external inputs, we need at least a degenerate equilibrium or more specifically - an attractor coinciding with the locus of $[\tilde{u}^1, \tilde{u}^2]^T$ for all external inputs in the set. However, to do this we have to consider cases where such locus could be a line and have to remove the simplifying assumption of unbiased features (\ref{unbiased_features}) which goes beyond a simple review of possibilities for metanetwork synthesis.
\end{itemize}

Finally, to demonstrate a more usable (mostly larger in size) implementation of pattern classification, we need networks of more than two input units. Larger networks introduce a rapid increase in the number of free parameters and need numerical solvers to find roots for high-dimensional systems. Such roots represent states where the system is at equilibrium in both the potential and weight spaces, i.e. for particular connection strengths and potentials. The unit potentials are kept fixed to a desired value corresponding to a preselected memory as usual: the input units to a training data sample and the output units to a training label. Each training pair generates a system for the weights and we can calculate a composite system or average over all of the solutions of the simpler systems. Measuring the accuracy of classification on the test set then happens as each test sample is presented to the network until it converges and upon convergence the equilibrium potentials of the output units are extracted and compared with the right answer.

Preliminary results on the MNIST handwritten digits as one of the classical machine learning datasets used for image classification required the solution of systems with 9910 equations and 118810 unknowns, 28x28 of which input units fixed to a particular image and 10 of which output units fixed to a binary vector with 1 for the correct digit class and 0 for the rest. Since solving a system for all cases of input-output pairs would require the same number of equations for each of a total of 50000 training samples and it doesn't guarantee robust solution as observed from simpler experiments, a more preferred way for training was to average over multiple cases. The numerical observations coincided with the predictions from the 2-output version of (\ref{mm13_is}) for the case of multiple input-output pairs but from a larger implementation.

\section{Discussion}

\subsection{Additional considerations}

Here we will list some additional considerations that we couldn't place anywhere in the text but are still important for a complete understanding of the various topics we had to cover. First we will counter some oversimplified criticisms of the attractor networks as associative memories, then we will briefly explore some possible extensions of these in terms of external inputs and classical stability, and finally we will relate our choice of mathematical theory to both the artificial and biological neural networks.

A major motivation for using attractors in applications of object recognition is in the fact that the network could handle a continuous input stream (or equivalently act as an online algorithm). While proponents of the attractor recall idea could argue that this is more biologically plausible in general, opponents could argue that brain activity would never become steady following such mathematical description. However, processing a stream of information would never lead to complete convergence since the input is practically never steady and such convergence could be optimal at the very best. An objective in learning would then be to minimize the processing time or maximize the convergence rate so that the next time the same input is encountered it is handled faster and contextualized better.

Another oversimplifying idea we have to disqualify is the need to reset the initial conditions when changing input suggested by \cite{Hirsch1989}. The main argument of Hirsch is that we cannot define a map from the external input to the equilibrium state $H(U)=u^*$. If we switch the input $U$ with another $\bar{U}$ while converging to $u^*$, it will change the dynamics and the convergence to a possibly different equilibrium $\bar{u}^*$. Switching back to $U$ will no longer guarantee convergence to $u^*$ since the initial condition at switching can now be within the basin of attraction of a different point $\hat{u}^* \neq u^*$. Therefore, one cannot map an input to an attractor and needs to reset the initial condition on change of the input. While these conclusions are true, not being able to associate equilibria with constant external inputs might actually be a desired effect. In particular, we can still map a sequence of constant external inputs to a final equilibrium state and can thus encode sequential information and have a network that recalls different memories depending on the sequence of perceived inputs.

Our conclusions on robustness with respect to external input, i.e. recall of the same attractor from multiple external inputs, can be extended to sustaining the same attractor for a complete external input sequence. The simplest possible sequence is of course a single constant input and is the expected first step in the increase of complexity of any stored information. A sequence of constant inputs is equivalent to a piecewise constant external input and can be generalized to any time dependent external input. The reversed form of sustaining the same attractor for a time dependent input then is generating time dependent output of this attractor. A cognitive interpretation for the first case was already given - recalling the same concept from time evolving stimulus like recognizing a known melody or predicting a known trajectory. A cognitive interpretation for the second case could be stimulus generation like producing handwriting or telling a verbal story. In all cases and exciting as it is, the mathematical and engineering realization of these suggestions deviates away from our current scope.

Another extension in the way we interpret stability for neural networks that we had to introduce here is the concept of partial stability. A recalled pattern is generally expected to be stored in a part of the network rather than the entire network. The general pattern sparsity is usually estimated to be somewhere in between of all cells and a single "grandmother" cell but neither of the two extremes. Storing in a subregion in a network also reduces the required processing and coupling needed for recall. Thus, we may be interested in convergence on a particular manifold as opposed to the full phase space which is also easier to handle mathematically. The behavior of the system outside of the manifold could oscillate, be chaotic or even unstable as long as such instability is unreachable in practice due to some preventive conditions. We are satisfied with it as long as the set of possible computational models we consider contains a set of models achieving the high cognitive tasks we aim at.

This brings us to a final point about the recurring topic of biological plausibility. We prefer performing classification with an attractor network, investigating any principles of its memory consolidation or synthesis shortcuts, instead of backpropagating error from a prescribed correct label. Our preference for feature detection involves two types of units instead of simpler lateral inhibition alternatives. One might argue that we need validation from reality and therefore are striving for optimal biological realism and successful numerical or empirical predictions of the model would affirm the approaches it was build on and in turn reveal new information about the biological reality itself. In the case of the feature detector, this would be a prediction of the extra unit type for the perception of black which could be a subtype of a known cell in the retina inhibited and inhibiting the first type, a new type of interaction, or something else. However, our interest is specifically in mathematical modelling in engineering, i.e. we are not concerned with whether or not our approach matches the biological reality as long as it achieves its purpose. The modelling performed here is merely a provider for engineering directions that we need in order to search for a common solution of a wide spectrum of problems in artificial intelligence.

None of our models are centered around engineering alone either. Our main criticism of the computationally efficient and specifically accurate yet very artificial neural networks is exactly their excessively large deviation from simpler and more natural solutions. At the same time these solutions are not available from the still difficult to generalize but truly biological neural networks. Engineering freedom in solving a problem is unquestionable but how far this freedom can go will remain a grey area. Nevertheless, the possibility of existence of a universal solution and its large attractiveness is one of the main inspirations for using neural networks for many computational tasks instead of support vector machines or other powerful and simpler machine learning methods. Even though such multifunctional solution could be a lot harder to identify, it must be chased ardently and developments in instances solving particular problems should try to be as close to it as possible.

\subsection{Current limitations}

As this is a newly introduced model, it has many limitations. Here, we will briefly outline some of them:
\begin{itemize}
    \item Since the current metanetwork is a rate-based model, it suffers from all the limitations that the firing rate approximation carries. Most important of these is the fact that it cannot account for sequential encoding and time related phenomena as do pulsed models which are also trying to be as minimal as possible in terms of biological detail.
	\item Existence of connections and usage of zero weights is not the same for the current model. While to define the existence of a connection by a nontrivial weight makes sense and ties very well mathematical generality with specific implementation, the context term (\ref{e_ext}) in the equations is not compatible with this concept. Adding weights to the term is also not possible since the denominator could become zero if all weights in it are zero.
	\item Many parts of the model can be modified later on and are currently experimental. This is one of the main reasons it was introduced as a family of models with greater modularity of the added dynamics.
	\item The added structure increases significantly the computational requirements for simulation because it adds equations for all connections which are all pairs of units and equations for all metaconnections which are all pairs of units and connections.
	\item As mentioned before, robustness with respect to external input and input sequences is a lot more intuitive for the concept of memory recall and is currently lacking.
\end{itemize}

\subsection{Future developments}

Besides finding better ways to deal with the limitations the current model is facing, two major extensions which are still not well-developed and well-studied are
\begin{itemize}
	\item plasticity - specifically learning which can bring to adaptation and generalization of input
	\item abstraction - a multilayered network is capable of hierarchical representations where lower-level features are combined in higher level ones
\end{itemize}

Regarding the first direction, finding a way to make the network rearrange itself properly is the only way of storing the information it needs to know to perform even the most basic cognitive tasks. This poses further challenges in terms of supervision from how to obtain feedback about the quality or success of storing to could we form known network structures like feedforward and feedback distinction, connection cycles, cell specializations and specialized cell arrangements.

Regarding the second direction, multiple layers can be helpful as multiple levels of abstraction for tasks like object detection and classification similarly to the case of DNNs. They could allow the network to perform classification of images on more challenging datasets like CIFAR-10, CIFAR-100, STL-10, SVHN, and ILSVRC2012 task 1. In the most realistic case a multilayer learning capable version can be tested on video data sets which are the closest to a real environment data allowing for cognitive development.

\newpage
\appendix
\appendixpage
\addappheadtotoc

\section*{Visuals}

\begin{figure}[!htbp]
\centering
\includegraphics[width=0.75\textwidth]{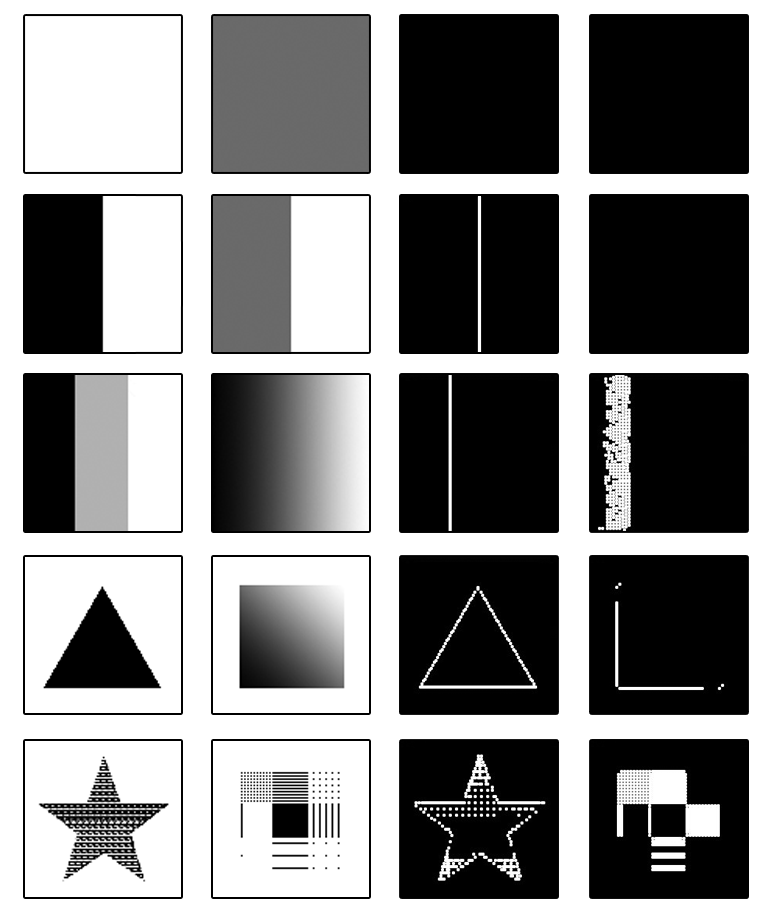}
\caption{Visual experiments with the feature detector - each row is a separate case with two stages for the first two boxes and results in the second two boxes: a) white and grey uniform input - outputs for both unit types for both stages; b) contour line of high and low contrast - output after sufficiently long time (both unit types for the high contrast and the first unit type for the low contrast) and short-term output of the low contrast (for first unit type, remains long-term for second unit type); c) two simultaneous contour lines of high and low contrast and a gradient - output of second unit type throughout for the two lines and at a later stage for the gradient; d) shapes introducing corners - output for both unit types for the triangle and for the first unit type for the square (fixed stage); e) shapes hidden by noise - output for the first unit type for both the star and the square (fixed stage);}
\label{static_features}
\end{figure}

\begin{figure}[!htbp]
\centering
$\begin{array}{cc}
	\includegraphics[scale=0.4]{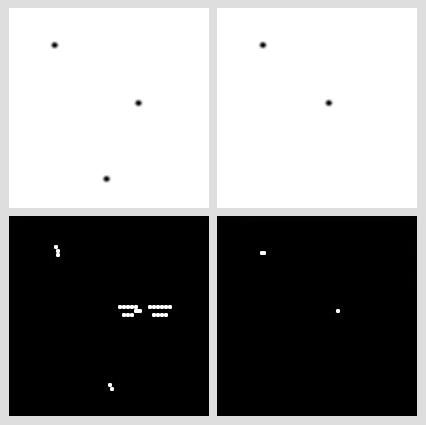} &
	\includegraphics[scale=0.4]{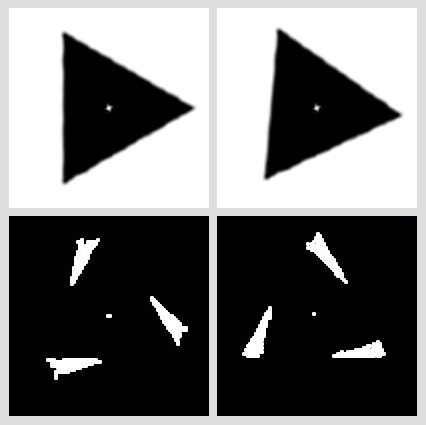}\\
	\includegraphics[scale=0.36]{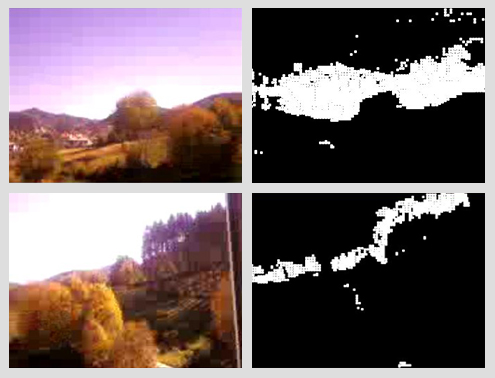} &
	\includegraphics[scale=0.27]{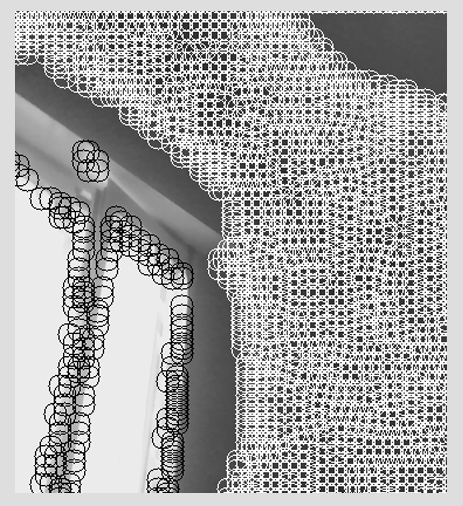}
\end{array}$
\caption{Motion detection as temporal features: a) static, moving, and blinking dots (top-left) with two snapshots in white and outputs of the first and second unit type in black; b) rotating triangle (top-right) with two snapshots in white and outputs of the first and second unit type in black; c) video (bottom-left) with two snapshots to the left and the first unit type response on the right; d) camera (bottom-right) with overadaptation due to long exposure;}
\label{dynamic_features}
\end{figure}

\begin{figure}[!htbp]
  \centering
	$\begin{array}{cc}
		\includegraphics[scale=0.8]{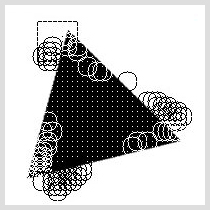} &
		\includegraphics[scale=0.8]{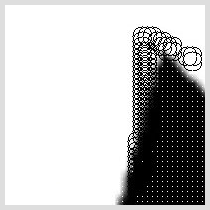}
	\end{array}$
  \caption{Object tracker resolutions - peripheral vision (left) and central vision (right) corresponding to the subregion of directed attention.}
\label{tracker_resolutions}
\end{figure}

\section*{Technologies used}

Our main language of choice is python (2.7) together with numpy and OpenCV packages for numerical and respectivelly computer vision processing. We also made use of one of the main MPI bindings packages in the open source world namely mpi4py \cite{Dalcin2016}. The porting and development of any prototype code specifically for the MPI (message passing interface) was necessary in order to make the best use of the supercomputing resources provided by Caliban - the high performance cluster in the University of L'Aquila - mainly for parameter exploration regarding the feature detector and investigation of multiple types of cellular neural network connectivity. Parallelizing the code was possible because of its design which involves step synchronized virtual time for the simulation, locally to loosely connected regions making use of a halo swapping technique, and memory mapped and region restricted data file loading.

We also made additional use of software like octave and maxima for the network synthesis part and wrote and tested some small code sections on theano, torch, and tensorflow.


\bibliographystyle{amsplain}
\bibliography{sources}
\addcontentsline{toc}{section}{Bibliography}

\end{document}